\documentclass[12pt]{article}

\usepackage{newtxtext,newtxmath}
\usepackage{graphicx}

\usepackage[letterpaper,margin=1in]{geometry}
\usepackage{nameref}

\renewenvironment{abstract}
	{\quotation}
	{\endquotation}

\date{}

\makeatletter
\renewcommand{\fnum@figure}{\textbf{Figure \thefigure}}
\renewcommand{\fnum@table}{\textbf{Table \thetable}}
\makeatother

\usepackage{scicite}

\usepackage{url}

\usepackage{gensymb}
\usepackage[normalem]{ulem}

\def\scititle{
          Advancing Minimally Invasive Precision Surgery in Large Open Cavities with Robotic Flexible Endoscopy
}
\title{\bfseries \boldmath \scititle}

\author{
	Michelle Mattille$^{1}$,
	Alexandre Mesot$^{1,6}$, 
    Miriam Weisskopf$^{2}$,
    \and
	Nicole Ochsenbein-Kölble$^{3,4}$,
    Ueli Moehrlen$^{4,5}$,
    Bradley J. Nelson$^{1}$,
    Quentin Boehler$^{6}$$^\ast$
    \and
	\small$^{1}$ Multi-Scale Robotics Lab, ETH, Zurich, Zurich, Switzerland.\and
        \small$^{2}$ Center for Preclinical Development, University Hospital Zurich, University of Zurich, Zurich, Switzerland.\and
        \small$^{3}$ Department of Obstetrics, University Hospital of Zurich, Zurich, Switzerland. \and
        \small$^{4}$ The Zurich Center for Fetal Diagnosis and Therapy, University of Zurich, Zurich, Switzerland.\and
        \small$^{5}$ Department of Pediatric Surgery, University Children's Hospital Zurich, Zurich, Switzerland. \and
        \small$^{6}$ Medical Robotics Lab, ETH, Zurich, Zurich, Switzerland.\and
	\small$^\ast$Corresponding author. Email: qboehler@ethz.ch
}

\begin{document} 

%TC:ignore
\maketitle

\begin{abstract} \bfseries \boldmath
Flexible robots hold great promise for enhancing minimally invasive surgery (MIS) by providing superior dexterity, precise control, and safe tissue interaction. Yet, translating these advantages into endoscopic interventions within large open cavities remains challenging. The lack of anatomical constraints and the inherent flexibility of such devices complicate their control, and the limited field of view of endoscopes restricts situational awareness. We present a robotic platform designed to overcome these challenges and demonstrate its potential in fetoscopic laser coagulation—a complex MIS procedure typically performed only by highly experienced surgeons. Our system combines a magnetically actuated flexible endoscope with teleoperated and semi-autonomous navigation capabilities for performing targeted laser ablations. To enhance surgical awareness, the platform reconstructs real-time mosaics of the endoscopic scene, providing an extended and continuous visual context. The ability of this system to address the key limitations of MIS in open spaces is validated in vivo in an ovine model.
\end{abstract}

% The first paragraph of any Science paper does NOT have a heading
% Nor is it indented
\noindent

\paragraph{One-Sentence Summary:} Robotic flexible endoscopy improves minimally invasive surgery in open cavities through assisted control and navigation.
%TC:endignore
\section*{Introduction}

Flexible endoscopy robots have the potential to transform minimally invasive surgery (MIS) by offering enhanced dexterity, deformability, and intrinsic safety compared to conventional rigid instruments, and are currently being used for a wide range of procedures~\cite{dupont_decade_2021, da_veiga_challenges_2020,zhu_intelligent_2021} (see Figure~\ref{fig:overview}A). However, their inherent compliance and high-dimensional configuration space pose substantial challenges for modeling, controlling, and sensing~\cite{russo_continuum_2023,nazari_visual_2022, da_veiga_challenges_2020}. To date, most advances have concentrated on luminal environments, where flexible devices navigate in narrow and tortuous environments and are constrained by the lumen's wall~\cite{prado_robotic-assisted_2024, brumfiel_variable_2025, greenidge_harnessing_2025, dreyfus_dexterous_2024, martin_enabling_2020}. 
Although flexible endoscopy robotic platforms targeting large open cavities in the pregnant uterus, the stomach, or the urinary bladder are promising, they have so far remained in the early research stages with few preclinical results focused on device dexterity and shape morphing rather than demonstrating entire surgeries~\cite{mao_magnetic_2024, ahmad_development_2023, da_veiga_challenges_2020}. The nature of these environments makes navigation and trajectory planning challenging due to the absence of anatomical constraints to naturally guide and stabilize flexible devices. Additional challenges of these single-port procedures include small instrument diameters that limit the amount of integrated sensing and a lack of overview of the surgical scene due to the limited field of view of endoscopic cameras, which are navigated close to the tissue surface~\cite{bano_chapter_2024,phan_optical_2020}. The latter poses a substantial mental load on surgeons when the extent of the surface to be explored is much larger than the field of view of the camera, as is the case in these organs. The clinicians must memorize the environment during the procedure for localization and navigation, as well as identification and characterization of areas of interest. To address this issue, the generation of 3D reconstructions or panorama images, also called mosaics, has been proposed. Recent advances in computer vision and machine learning have led to promising results with clinical endoscopic camera images of those challenging low-texture environments ~\cite{bano_chapter_2024,li_robust_2023,phan_optical_2020,widya_stomach_2021}, where traditional methods fail. However, most of these methods focus on offline processing, and none have been translated to the clinic to date.

A challenging endoscopic MIS in an open cavity that would benefit from flexible surgical robots and the generation of endoscopic image mosaics is fetoscopic laser coagulation (FLC). It is the gold standard treatment for twin-to-twin transfusion syndrome (TTTS), a life-threatening disease of twin fetuses in the womb. 

TTTS is characterized by an imbalanced blood flow between twins who share the same placenta and affects 10-15\% of such twin pregnancies~\cite{VanDerVeeken2019,bamberg_twin--twin_2022}. Even when performed by highly trained and experienced surgeons in high-volume centers, double-twin survival rates of only 70\% have been reported, and 10\% of the surviving children exhibit long-term neurodevelopmental impairments~\cite{bamberg_update_2019}.

The purpose of FLC is to stop the blood flow between the twins. The surgeon inserts a fetoscope, an endoscope for fetal surgery, through a trocar into the uterus, scans the placental surface to identify all vascular anastomoses that connect the twins' circulations, and ablates these vessels using a surgical laser~\cite{VanDerVeeken2019,Pandya2020a}. Operating on anterior placentas located towards the abdomen of the mother is particularly challenging due to their limited accessibility with the current curved rigid fetoscope (11508AAK, Karl Storz)~\cite{bouchghoul_management_2025, bamberg_update_2019}. Limited manipulability results in a tangential view of the placental surface, making visualization, identification, and ablation of anastomoses difficult or even impossible~\cite{cruz-martinez_flexible_2023}. As a consequence, incomplete separation of the circulations may occur, increasing the risk of recurrence of the condition after surgery~\cite{VanDerVeeken2019}. It is also a major risk factor for neurological impairments in surviving children~\cite{stirnemann_fetal_2018,Pandya2020a, bouchghoul_management_2025}.

In a clinical study conducted by Cruz-Martinez et al., the authors reported the benefits of using a flexible video endoscope in addition to conventional instruments for FLC~\cite{cruz-martinez_flexible_2023}. Although this approach led to better visibility of the placental surface and enabled coagulation of anastomoses that would not be possible otherwise, the device was challenging to operate due to its additional degrees of freedom and required two trained operators, rather than the single fetal surgeon needed for the conventional treatment.

Ahmad et al. developed a hand-held flexible fetoscope with a pneumatically actuated distal tip deflection mechanism for FLC~\cite{ahmad_development_2023}. They demonstrated improved access, visualization, and ablation performance over conventional fetoscopes during a user study. They remain the only group to date to have evaluated their active fetoscope in vivo, with an emphasis on showing enhanced access for ablation in an ovine model. Although their work highlights the advantages of having a fetoscope with a flexible distal tip, the saturation of the endoscopic camera sensor during lasering prevented them from demonstrating a reliable vessel ablation. Their system also did not leverage the active pneumatic actuation to provide automated robotic assistance or closed-loop control, which has the potential to substantially reduce the complexity of device manipulation. Hernansanz et al. demonstrated robotic assistance capabilities in vitro~\cite{hernansanz_robot_2024}. Their platform allows the user to save and navigate back to points of interest, but is limited to easy-to-access posterior placentas as their platform consists of a straight fetoscope mounted on a robot arm.

Our group previously demonstrated the potential of a magnetically guided flexible fetoscope for FLC~\cite{lussi_magnetically_2022}. This preliminary work demonstrated in vitro and ex vivo navigation capabilities and ablation of vessels in ex vivo human placentas. However, controlling the direction of the external magnetic field with a haptic controller rather than the direction of motion of the endoscopic images proved too counterintuitive for future clinical use. Autonomous navigation capabilities were shown for short distances toward a target visible in the endoscopic view and limited to ex vivo settings. More recently, we demonstrated in vitro and ex vivo how endoscopic image mosaics enable autonomous navigation in open cavities, which has the potential to substantially improve the outcome and reduce the complexity of procedures such as FLC~\cite{mattille_autonomous_2024}. However, the presented method was restricted to a single fetoscope orientation.

This work introduces a flexible surgical platform that addresses key challenges of MIS in large open cavities including the pregnant uterus, stomach and urinary bladder through its combined navigation and mapping capabilities that are capable of handling in vivo environments. Its benefits and efficacy are demonstrated in realistic preclinical conditions for the treatment of complex cases of TTTS (see Figure~\ref{fig:overview}B). Our platform is able to access and ablate vessels precisely in any placental location and can generate mosaics of endoscopic images in real time. Our flexible robotic fetoscope incorporates all functionalities required for FLC and is small enough to pass through a 10~Fr trocar. This is important, as larger trocars have been associated with less favorable clinical outcomes~\cite{bouchghoul_management_2025, van_der_schot_impact_2024}. We introduce an intuitive manual steering method and provide supervised automated navigation over both short and long ranges, enabling robotic assistance relevant to the different phases of the FLC procedure (see Figure~\ref{fig:overview}C-F). The steering is purely image-based, eliminating the need to integrate additional localization sensors into the small instrument tip, and its model-free approach supports a broad range of device stiffnesses. Our approach overcomes the negative effects of hand tremors on lasering performance (see Figure~\ref{fig:overview}G), and significantly improves the accuracy and precision to reach and ablate targets compared to a state-of-the-art conventional instrument, which could lower complication rates by preserving healthy placental tissue~\cite{akkermans_what_2017}. Our robotic platform also significantly reduces the physical strain and frustration imposed on its users compared to conventional FLC. These findings are supported by user studies conducted with both an experienced clinician and novices, and by in vitro, and ex vivo experiments, as well as in vivo experiments at human scale in a living ovine model. This work presents an in vivo demonstration of a robotic fetal surgery platform featuring assisted navigation capabilities and visualization during ablation in the context of FLC, marking an important step toward the clinical translation of the robotic treatment of TTTS. In vivo experiments in large mammals offer the most realistic and complex environment before clinical translation~\cite{nelson_magnetically_2022}. Performance often degrades when transitioning from the highly controlled in vitro and ex vivo settings~\cite{nazari_visual_2022}. Compared to our prior work~\cite{lussi_magnetically_2022, mattille_autonomous_2024}, the hardware and software components were redesigned or enhanced to achieve the required level of readiness.

The proposed navigation and mapping methods show potential for other MIS in large open cavities. Gastroscopies in the stomach and cystoscopies in the urinary bladder pose similar challenges to clinicians to those encountered in FLC. These procedures serve as the gold standard for diagnosing, monitoring, and treating gastric and bladder cancer, respectively~\cite{phan_optical_2020,kurniawan_flexible_2017, januszewicz_quality_2020, emura_principles_2020,rindorf_market_2021,pena_usefulness_2022}. Gastroscopies are performed hundreds of millions of times annually~\cite{kurniawan_flexible_2017, januszewicz_quality_2020, emura_principles_2020} and cystoscopies must be conducted at close intervals for bladder cancer patients due to the high recurrence rate of the disease, making it one of the most expensive malignancies~\cite{lee_urine_2022}. Endoscopes (called gastroscopes or cystoscopes) are used in both procedures to examine the inner walls of the stomach or bladder, searching for lesions such as ulcers or cancerous tissue, taking biopsy samples, or performing various procedures such as the removal of small tumors. Similarly to FLC, clinicians rely on live endoscopic image views as the primary source of feedback. The camera is held close to the surface for optimal image quality, and clinicians rely on memory to ensure a complete scan of the organ and reduce the mental load of clinicians~\cite{phan_optical_2020}. Mosaicking and 3D reconstructions of endoscopic images have been proposed to ensure a complete scan. They enable the visualization of entire lesions with high image resolution, and facilitate device localization within the organs~\cite{phan_optical_2020, widya_stomach_2021}. These techniques have shown promising results with clinical data but are not yet applied clinically~\cite{li_bladder_2024,phan_optical_2020,lurie_3d_2017}. We demonstrate the feasibility of the proposed navigation and mosaicking methods for cystoscopies and gastroscopies through in vitro and ex vivo experiments under various conditions and endoscopes. The methods can also handle tissue contact, as the flexible proximal sections of the magnetic cystoscope and gastroscope remain in constant contact with the esophagus or the urethra, and biopsy samples can be taken.

\section*{Results}
\subsection*{Robotic platform design}

Our robotic platform comprised a flexible fetoscope with a magnetically actuated distal tip (see Figure~\ref{fig:workspace}A-C), which was steered by external magnetic fields generated by a preclinical electromagnetic navigation system (eMNS)~\cite{gervasoni_human-scale_2024}, and the insertion depth was controlled by a mechanical advancer unit (see Figure~\ref{fig:workspace}D). The eMNS enabled control of the fetoscope’s orientation (see Figure~\ref{fig:overview}B). Endoscopic images could be stitched in real time to generate image mosaics (see Figure~\ref{fig:overview}D), which further enhanced visualization and exploration capabilities of the anatomical environment based on automated scanning motion as proposed in~\cite{mattille_autonomous_2024}. Navigation could be performed using three strategies that cover the range of tasks encountered during FLC: manual navigation, in which the user controlled the motion of the endoscopic image via a joystick; short-range automated navigation, which steered the tip to targets visible within the endoscopic view; and long-range automated navigation to targets outside the current view, selected on a previously generated image mosaic (see Figure~\ref{fig:overview}C-F).

\subsection*{Robotic fetoscope design}

We developed a robotic fetoscope with an outer diameter (OD) of 3.2~mm (see Figure~\ref{fig:workspace}A and Figure~\ref{sfig:fetoscope}). It was compatible with 10~French (Fr) trocars commonly used in FLC.

The device featured a transparent tip that integrated optical fibers for illumination, a camera with an optical filter placed in front of it, and a working channel for a laser fiber (OD of 500~$\mu$m or less) as illustrated in Figure~\ref{fig:workspace}C. The distal section comprised a series of stacked magnets and precision-engineered steel ball joints (see Figure~\ref{fig:workspace}B), designed using the open-source software toolbox introduced in~\cite{mesot_parametric_2025}, to allow large bending angles and precise magnetic navigation. The detailed design is shown in Figure~\ref{sfig:fetoscope}.

\subsection*{Workspace characterization}

The robotic fetoscope workspace was characterized by subjecting the device to a magnetic field of 25~mT generated by the eMNS, with a 200~$\mu m$ core laser fiber inserted into the device. By sweeping over the magnetic field directions, the fetoscope's tip bent in various directions. Figure~\ref{fig:workspace}E shows the tracked fetoscope's tip positions in space during this experiment. A maximum bending angle of 173$\degree$ was achieved for the fetoscope in the upward configuration.

\subsection*{Accessible uterine locations characterization}

To characterize the ability of our robotic fetoscope to ablate vessels located on an anteriorly located placenta during FLC, we determined the best angle~$\alpha$ between the fetoscope tip axis and the uterine surface for all accessible locations on a uterus model at 26~weeks of gestation for an insertion point on the lateral uterine wall (see Figure~\ref{fig:workspace}F). An angle of~$\alpha = 90\degree$ was considered optimal for visualization and vessel ablation~\cite{Akkermans2017}. The results were compared to a conventional, precurved, rigid fetoscope (11508AAK, Karl Strotz, Germany) designed for anteriorly located placentas. 

Results showed that the robotic fetoscope could access more locations on the uterus surface (see Figure~\ref{fig:workspace}F), and with better ablation angles across the anterior uterine wall compared to the conventional device. The robotic fetoscope achieved angles~$\alpha$ greater than 45$\degree$ at 91\% of the locations on the anterior uterine surface, compared to 73\% for the conventional fetoscope. For angles exceeding 70$\degree$, the robotic and conventional fetoscope covered 89\% and 33\% of the anterior surface, respectively. On the left lateral uterine wall, the conventional device exhibited progressively smaller angles as the target points approached the insertion site until the area became unreachable. In contrast, the robotic fetoscope maintained high ablation angles across the surface, apart from the immediate vicinity of the insertion point.

\subsection*{Image-based magnetic field control}

The control of the robotic fetoscope's motion was solely based on the fetoscopic image feedback. It consisted of changing the external magnetic field to bend its tip, and using image-based visual servoing to achieve a desired velocity in the fetoscopic image, or reach a 2D target point~$\mathbf{s}$ selected in that image or in a previously generated image mosaic. The proposed navigation methods did not require any additional embedded sensors or tracking devices. The 2D velocity~$\dot{\mathbf{s}}$ of a point~$\mathbf{s}$ defined in the endoscopic image was linearly mapped to the instantaneous rotation of the magnetic field vector $\dot{\mathbf{q}}= [\dot{\alpha}~ \dot{\beta}]^T$ (see Figure~\ref{sfig:navion}) using the image Jacobian matrix~$\mathbf{J} \in \mathbb{R}^{2 \times 2}$. The matrix $\mathbf{J}$ was initially estimated with a calibration process and was then updated over time using the Broyden method~\cite{wu_model-free_2015} to guarantee the validity of the mapping as the fetoscope moved. 

The choice of a point~$\mathbf{s}$ and of its desired velocity~$\dot{\mathbf{s}}_d$ depended on the selected navigation strategy. For manual navigation, $\dot{\mathbf{s}}_d$ was defined as the desired motion of the image center. Its direction coincided with the direction of the joystick operated by the user, and its magnitude $\dot{s}$ was set to a fixed value. For short-range automated navigation,~$\mathbf{s}$ was defined as a target location selected in the endoscopic image, and its desired velocity was updated using a PID control law so that the target moved to a reference position~$\mathbf{r}$, which was typically in the center of the image. For long-range automated navigation, the target location was chosen in a previously generated image mosaic outside of the current endoscopic image. First, the robotic fetoscope was moved with a magnetic field estimated to correspond to the target location. Once the target was in proximity of the reference position in the endoscopic image, short-range automated navigation was used for the remaining distance.

\subsection*{In vitro characterization of the imaged-based control}

This image-based control was first evaluated in vitro by navigating the robotic fetoscope for desired image motions of the image center with stepwise (see Figure~\ref{fig:pov_eval}A) and continuous direction changes (see Figure~\ref{fig:pov_eval}B) at a constant speed. The directional component of the motion was more important to the user than the magnitude to make navigation feel intuitive. The closer the desired and actual directions were, the more intuitive the navigation strategy felt. The fetoscope was inserted horizontally from the side with its tip pointing forward and upward (see Figure~\ref{fig:pov_eval}C-D), which was the most challenging configuration to control given the destabilization effect of gravity. This simulated interventions on lateral and anterior placentas during FLC.

The trajectories of the image center over time could be estimated by integrating its speed (see Figure~\ref{fig:pov_eval}A-B), and they approximated a square and a circle, respectively, as expected from the given inputs. The stepwise direction change was the most demanding as the change in motion direction was instantaneously applied (see Figure~\ref{fig:pov_eval}E). In this case, we observed an overshoot with a median peak error of $24\degree \pm 0.69\degree$ and an overall median error of 2.2$\degree \pm 1.3\degree$. Continuous direction changes showed smoother motion and only minor deviations from the desired direction (see Figure~\ref{fig:pov_eval}F), with a median error of $1.8\degree \pm 1.3 \degree$.

The importance of updating the image Jacobian matrix~$\mathbf{J}$ is illustrated in Figure~\ref{sfig:detailed_nav_eval}, which shows a substantially lower performance in following the desired motion when the Jacobian was kept constant after an initial calibration. This resulted in a notable deviation from the square and circular trajectories obtained with the update. This update required a robust estimate of the motion of the image center over time. The estimated local image motion and the ground truth are shown in Figure~\ref{sfig:detailed_nav_eval_est_vs_truth} with a mean and standard deviation of $0.86\pm 1.63$~pixel/s of the absolute error over time.

\subsection*{In vivo navigation in an ovine model}

We demonstrated the feasibility of our navigation strategies under preclinical conditions in vivo in an ovine model (see Figure~\ref{sfig:in_vivo_setup}). Conducting such experiments was critical for validating the system's performance, as this was the closest available model we have to mimic the complex physiological environment encountered in humans.

An experienced fetal surgeon first steered the robotic fetoscope with a joystick using the manual navigation method. They successfully navigated along a vessel on the uterine wall, simulating a task performed in FLC (see Movie~S1). The results showed a close match between the desired motion from the joystick input and the estimated motion over time (see Figure~\ref{fig:in_vivo_nav_manual}A-D), with a median and median absolute deviation of the angle error of $8.6\pm 4.8\degree$ throughout the trajectory. The successive endoscopic images were stitched into an image mosaic (see Figure~\ref{fig:in_vivo_nav_manual}E), which shows the surgeon could follow the desired vessel with only a limited deviation between the center line of the vessel and the center of the image.

Using short-range automated navigation, six targets selected within the endoscopic view were reached with the center of the image as the reference position and within a threshold of two pixels (see Figure~\ref{sfig:in_vivo_vs_in_img}). Figure~\ref{fig:in_vivo_nav_automated}A-D illustrates the target trajectory at different keyframes, and the evolution of the error over time for the second target (see Figure~\ref{sfig:in_vivo_vs_in_img}B, Movie~S2). The six targets were reached within $3.1\pm 1.7$~s following an initial calibration. Travel distances were $128\pm 17.3$~pixels and fall times were $0.79\pm 0.43$~s. The delay time of the system response time was $0.41\pm 0.041$~s.

The long-range automated navigation to targets selected outside the current endoscopic image and within an image mosaic was also validated under in vivo conditions (see Figure~\ref{fig:in_vivo_nav_automated}E-I, Movie~S3, Figure~\ref{sfig:mosaic_w_all_targets}, and Figure~\ref{sfig:in_vivo_jumping}). The mosaic was first built by automatically navigating the robotic fetoscope along a spiral trajectory within the mosaic reference frame. The images were stitched together to form a mosaic image in real time (see Figure~\ref{sfig:in_vivo_automated_exploration}). The completed mosaic served as an interactive user interface, allowing for the successful automated navigation to five manually selected target locations throughout the mosaic. Each target was reached with an error threshold of two pixels with a mean duration of $6.31 \pm 1.3~s$ (see Movie~S3, Figure~\ref{sfig:in_vivo_jumping}), resulting in a higher accuracy compared to the ex vivo experiment in our previous work~\cite{mattille_autonomous_2024}. Targets could be reached with the same error threshold, irrespective of the distance between the starting and target position.

\subsection*{Ex vivo vessel ablation study on human placenta}

The influence of hand tremors on the energy required for vessel ablation was investigated on an ex vivo human placenta. The ablation performance of a rigid, hand-held replica of a conventional fetoscope was compared to that of the same laser fiber fixed in place (see Figure~\ref{sfig:ex_vivo_placenta_setups}). The latter simulated the ablation with our robotic fetoscope, which remained stationary within a constant magnetic field and was therefore not affected by hand tremors. 

The so-called “replica” was a research-prototype reproduction (see Figure~\ref{sfig:replica}) of the rigid, hand-held, conventional fetoscope 11508AAK commercialized by Karl Storz, which is the clinical state of the art for FLC in cases with anterior placentas. During ablations with the replica, a trained user attempted to maintain the same distance between the laser fiber and the vessel as in the fixed-fiber setup. For vessels of comparable diameters, the same laser power and duration were used in both setups. To evaluate whether the ablated vessels were closed, a colored dye was injected through the umbilical cord after ablation using the protocol introduced in~\cite{lopriore_accurate_2011}. Figure~\ref{fig:vessel_ablation}A and Figure~\ref{sfig:laser_eval_supp_results_1} show the results of the experiment. The ablated vessels were classified into three different types: ``Closed" if the ablation was complete (with no penetration of the color dye passed the ablation point), ``semi-closed" if no color was visible at the ablation site but the dye still penetrated further down along the vessel, and ``open" if the dye was still visible at the ablation site. 

At the selected laser duration and laser power, none of the vessels could be fully ablated with the hand-held replica, and out of the 7 vessels targeted with the fixed fiber, 3 could be closed and 2 semi-closed. These results indicated that the robotic fetoscope could successfully ablate vessels with reduced laser energy compared to a conventional, rigid, hand-held fetoscope reproduced with our replica.

Fixing the laser fiber in place was comparable to stabilizing the robotic fetoscope in a constant magnetic field. This was validated by measuring the displacement of the robotic fetoscope in a magnetic field of 25~mT at a distance of 135~mm from the eMNS and of a stationary camera of the same model as used in the robotic fetoscope. The standard deviation of the camera motion was comparable, with 0.024~mm for the fixed camera and 0.029~mm for the robotic fetoscope over a time period of 60~s.

\subsection*{In vivo vessel ablation in an ovine model}

\begin{figure}[p]
	\centering
	\includegraphics[width=0.8\textwidth]{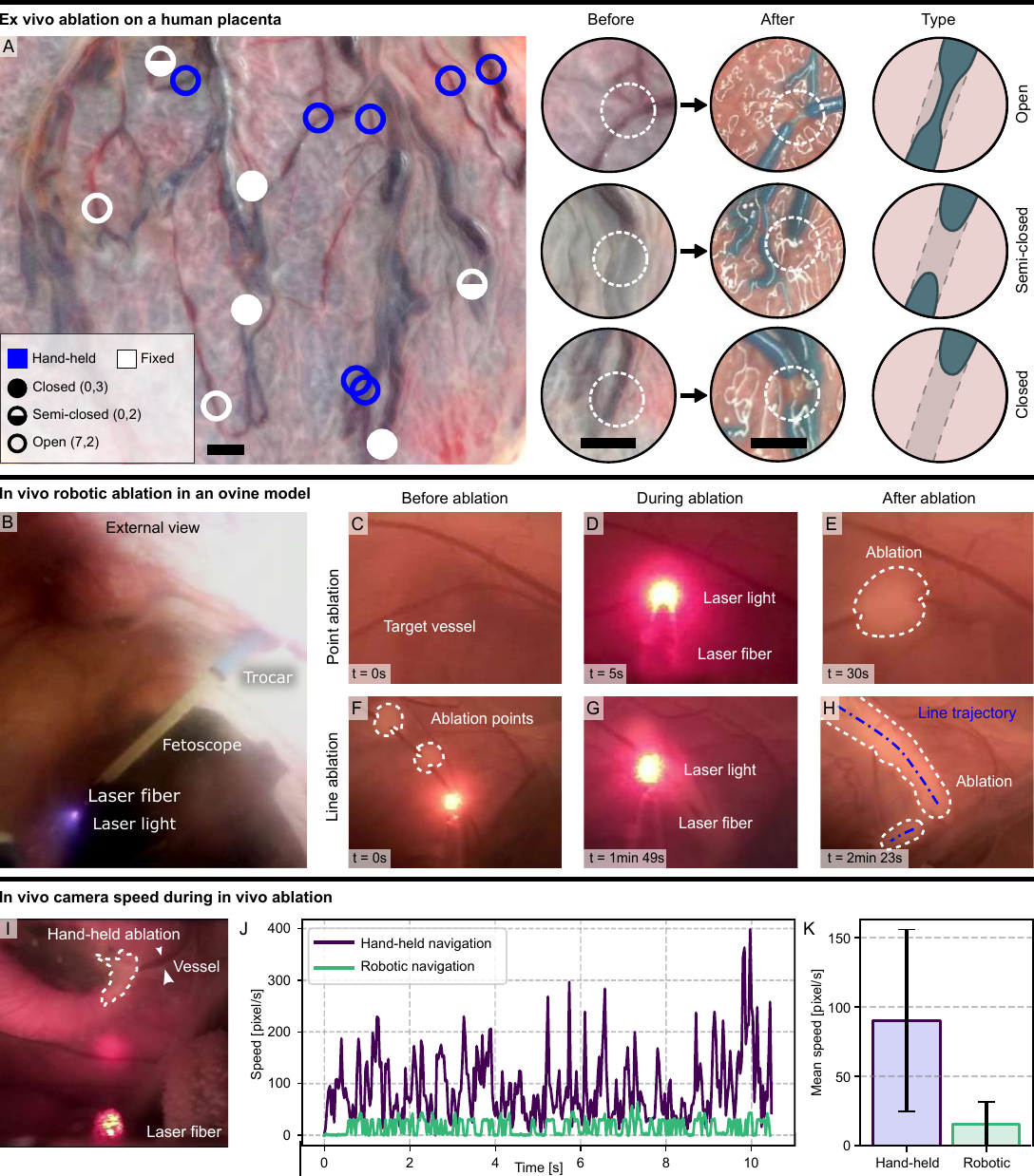} 
	\caption{\textbf{Ex vivo and in vivo vessel ablation.}
    (\textbf{A}) Locations of vessel ablations on ex vivo human placenta and their levels of completeness for a laser fiber fixed in place and for replica of a rigid, hand-held fetoscope (11508AAK, Karl Storz). The level of completeness is assessed by comparing the vessels before and after ablation and dye injection. The scale bar corresponds to 10~mm. (\textbf{B}) External view of the robotic fetoscope during the in vivo ablations. Fetoscopic views for (\textbf{C-E}) point and (\textbf{F-H}) line ablations. (\textbf{I}) In vivo hand-held ablation and (\textbf{J}-{K}) camera speed comparison for hand-held and robotic ablation with the robotic fetoscope including (\textbf{J}) the evolution of the speed over time, and (\textbf{K}) the mean (box) and standard deviation (whiskers) over the measurements.}
	\label{fig:vessel_ablation}
\end{figure}

To demonstrate the feasibility of vessel ablation in clinical settings, a fetal surgeon ablated vessels in vivo with the robotic fetoscope in an ovine model. They performed both point and line ablations on the placental vessels (see Figure~\ref{fig:vessel_ablation}B-H and Movies~S4-5). This simulated the Solomon technique, which is a state-of-the-art maneuver in FLC procedures consisting of connecting previously coagulated anastomoses with an ablation line~\cite{VanDerVeeken2019}. The results also showed that vessels were still visible during ablation (see Figure~\ref{fig:vessel_ablation}D and G), which was important for precise and safe ablations. Here, this was made possible by the integration of a filter on top of the camera sensor (see Figure~\ref{fig:workspace}C) to avoid the saturation of the camera sensor due to reflected laser light during lasering.

To investigate the influence of a hand tremor in vivo, the robotic fetoscope was held manually instead of being fixed in the advancer, and the magnetic field was kept constant to stabilize the flexible distal tip. The surgeon held the robotic fetoscope as steady as possible while ablating another vessel (see Figure~\ref{fig:vessel_ablation}I-J, Movie~S6). Figure~\ref{fig:vessel_ablation}J shows the speed of the center of the image over time for both point ablations. The mean speed and standard deviation were lower for the robotic navigation with $16\pm 16$~pixel/s compared to $90\pm 66$~pixel/s for the hand-held approach. This accounted for the noticeable difference between the spread of the ablation energy, which was qualitatively illustrated in Figure~\ref{fig:vessel_ablation}F and I, showing a more localized lasering mark on vessels ablated under magnetic guidance.

\subsection*{Usability study}

To demonstrate the usability of the robotic platform, we conducted a user study to assess and compare the performance of assisted manual control of the robotic fetoscope with the replica of the hand-held conventional fetoscope (see Figure~\ref{sfig:replica}) to perform two tasks commonly performed during FLC: ablation of targets at different locations in the uterus to simulate coagulation of individual anastomoses, and ablation along trajectories simulating the Solomon technique~\cite{VanDerVeeken2019}. The study was carried out using an augmented reality setup (see Figure~\ref{sfig:study_setup}), where participants received visual feedback consisting of endoscopic images augmented with a game environment to overlay virtual targets. Eleven participants completed each task three times with both instruments, and completed a NASA-TLX questionnaire for each run to score perceived workloads in terms of mental, physical, and temporal demand, as well as performance, effort, and frustration. For both tasks, robotic fetoscope navigation was performed with image-based manual navigation, and the replica was operated by hand without additional assistance.

In Task 1, the participants were asked to ablate 12 virtual targets on a simulated anterior placenta as quickly and accurately as possible (see Figure~\ref{fig:user_study}A-B and Movie~S7). This task alternated between navigation phases to go from the current target to the next, and ablation phases where the participant used a foot pedal to simulate the laser trigger (see Figure~\ref{fig:user_study}B). The accuracy and precision were significantly better for our robotic fetoscope compared to the replica (see Figure~\ref{fig:user_study}C), with median errors in all runs, targets, and participants of 140$\pm$22~$\mu m$ compared to 621$\pm$361~$\mu m$ for the accuracy during the ablation of targets (p = 0.001, W= 0.0, paired Wilcoxon signed-rank test). The time to navigate between targets was significantly longer for the robotic fetoscope, with a median of 20~s seconds, compared to 8.5~s for the commercial replica (p = 0.001, W= 0.0). No significant differences were observed between the devices for the duration of the ablation phase. 

In Task 2, the participants were asked to follow a continuous ablation trajectory in the endoscopic image (see Figure~\ref{fig:user_study}E and Movie~S8). Only the sections of the trajectory where the participant pressed the foot pedal for ablation were considered for evaluation (see Figure~\ref{fig:user_study}E). The accuracy and precision were significantly better for our robotic fetoscope compared to the hand-held replica (see Figure~\ref{fig:user_study}F), with weighted median error of 302$\pm$187~$\mu m$ for the robotic fetoscope compared to 451$\pm$268~$\mu m$ for the replica. The completion time was significantly shorter for the replica with a median value over runs and participants of 104~s compared to 162~s. 

NASA-TLX results analysis showed that the use of the robotic fetoscope led to a significant reduction in physical demand, effort, and frustration for both tasks compared to the hand-held replica according to a paired Wilcoxon signed-rank test (see Figure~\ref{fig:user_study}D and G). In addition, it also led to a significant reduction in mental demand in Task 2.

\subsection*{Other applications in large open cavities}
The versatility of the proposed navigation and mapping approaches beyond FLC was demonstrated using magnetic endoscopes of different sizes and stiffness developed for gastroscopy and cystoscopy (see Figure~\ref{fig:other_app}). The same parameters were used for all three applications.

The bending angles of the devices used are illustrated in Figure~\ref{fig:other_app}A, when they were exposed to the same magnetic field of 25~mT with a magnetic field orientation that resulted in a stable upward orientation for all tools. 

Figure~\ref{fig:other_app}B-C illustrates a magnetic cystoscope controlled in a porcine ex vivo bladder filled with urine using the image-based manual control (Figure~\ref{sfig:cysto_pov} and Movie~S10). Figure~\ref{sfig:cysto_pov} illustrates the directional components of the desired and estimated image center velocities during a 30~s navigation sequence, achieving a mean angle error of $11 \pm 6.9\degree$. A first mosaic was generated in real-time during this navigation, with a mean error of~$12.06\pm14.50$~pixels over a duration of~164~s (Figure~\ref{fig:other_app}C and Movie~S11). A second mosaic was generated with an automated exploration motion with a mean error of $17.46\pm6.11$~pixels and was then used for long-range navigation to successfully reach six targets selected in the mosaic (Figure~\ref{sfig:cysto_jump} and Movie~S12). The cystoscope also navigated successfully to six targets with the short-range navigation with an error margin of~2~pixels (see Figure~\ref{fig:other_app}D and Figure~\ref{sfig:cysto_vs_in_img}, and Movie~S13). 

Figure~\ref{fig:other_app}E-G illustrates a magnetic gastroscope controlled in vitro in a wet human stomach model. Image-based manual navigation was demonstrated both with and without an inserted biopsy forceps (Figure~\ref{sfig:tne_pov} and Figure~\ref{sfig:tne_pov_biopsy}, and Movie~S14 and Movie~S15) with mean angular errors of $16.46 \pm 22.25\degree$ and $7.18 \pm 7.52\degree$ respectively over durations of 30~s. Mosaics could be generated in real-time under an automated exploration motion with a mean error of~$28.90\pm2.24$~pixels (Figure~\ref{sfig:tne_jump}A, and Movie~S16). The mosaic was then used to reach five targets selected within the mosaic using long-range navigation with an error margin of 1~pixel (Figure~\ref{fig:other_app}F, Figure~\ref{sfig:tne_jump}B-F, and Movie~S17). The gastroscope was also controlled under short-range navigation with various illumination settings, and both with and without inserted biopsy forceps and reached targets with an error margin of 2~pixels (Figure~\ref{sfig:tne_vs_in_img}, Movie~S18).

\section*{Discussion}

Our endoscopic robotic platform addresses key challenges of MIS in open organs with flexible robotic devices. We propose image-based navigation methods in open space with various levels of autonomy which do not rely on modeling the complex robot kinematics or on the integration of additional sensors for precise localization. Our robotic fetoscope is able to create an overview of the surgical scene through real-time mosaicking of endoscopic images, and show the potential of surgical robots for precise and safe vessel ablation by eliminating the negative influence of hand tremors. The efficacy of our approach was demonstrated in clinically realistic conditions in vivo on a human scale in an ovine model. In vivo experiments in large mammals offer the most realistic and complex environment before clinical translation~\cite{nelson_magnetically_2022}. Performance often degrades when transitioning from the highly controlled in vitro and ex vivo settings~\cite{nazari_visual_2022}. Compared to our prior work~\cite{lussi_magnetically_2022, mattille_autonomous_2024}, the hardware and software components were redesigned or enhanced to achieve the required level of readiness.

This work focuses on addressing the current limitations encountered during FLC, a challenging minimally invasive fetal surgery in the uterus to treat TTTS. It should only be performed by highly trained and experienced surgeons, as twin survival rates have been strongly correlated with the annual case load of hospitals~\cite{Akkermans2015}. These considerations make this procedure a relevant target for our platform, with expected benefits from its use in clinical settings. Facilitating surgery through robotic assistance has the potential to improve outcomes in low-volume centers, as surgeons with less training can perform it at a higher level of skill, thus reducing geographical discrepancies~\cite{dupont_grand_2025}.

We show that the access to anteriorly located placentas during fetal surgery can be improved with our platform by using a flexible robotic fetoscope instead of a conventional rigid instrument (see Figure~\ref{fig:workspace}). Flexible devices can provide a more dexterous and potentially safer navigation during MIS~\cite{dupont_decade_2021}, including for the treatment of TTTS~\cite{rodrigue_soft_2024,bamberg_update_2019,spruijt_twin-twin_2020}. In this case, they allow for improved visualization of the placental vasculature and better ablation angles, which is a main factor in the efficacy of vessel ablation~\cite{Akkermans2017}. Access to previously inaccessible vessels leads to a more complete separation of the twins' circulatory systems, thereby avoiding recurrence of the disease~\cite{cruz-martinez_flexible_2023, bouchghoul_management_2025} and in turn the risk of neurological damage~\cite{stirnemann_fetal_2018}.

Real-time mosaicking of endoscopic images during MIS in open organs provides maps of the anatomical environment~\cite{bano_chapter_2024}. In this study, we generate these mosaics in real time during autonomous exploration motions and demonstrate how mosaics can serve as a user interface to efficiently access locations outside of the current camera image under supervised autonomy (see Figure~\ref{fig:in_vivo_nav_automated}E-I and Figure~\ref{sfig:mosaic_w_all_targets}). This contribution is relevant to FLC where the identification and ablation of all shared vessels between the twins are key~\cite{peeters_identification_2015}. The surgeon must currently memorize the vasculature as the fetoscope must be held close to the placental surface for good visualization~\cite{peeters_identification_2015,Pandya2020a}. In this case, mosaicking and supervised autonomous navigation can lower the mental load of the surgeon and facilitate navigation back to previously inspected vessels.

\begin{figure}[p]
	\centering
	\includegraphics[width=1\textwidth]{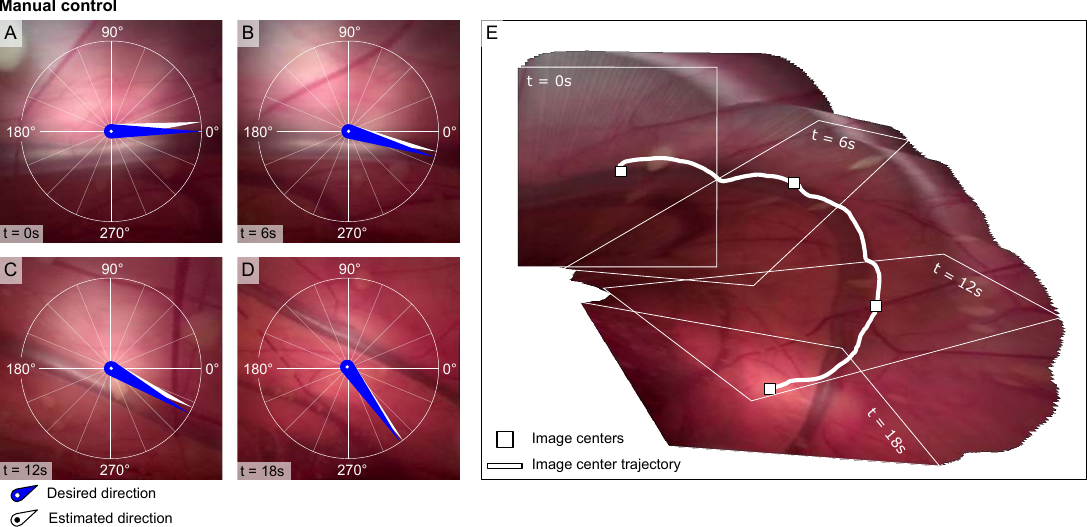} 
	\caption{\textbf{In vivo manual navigation.}
    (\textbf{A-D}) Endoscopic images at different points in time during the in vivo manual navigation in an ovine model with the desired and the estimated image center velocity overlaid. (\textbf{E}) Mosaic generated from the endoscopic images during manual navigation. 
        }
	\label{fig:in_vivo_nav_manual}
\end{figure}

Our platform offers intuitive navigation strategies with various levels of robotic assistance, which have the potential to reduce inter- and intra-surgeon variability by standardizing surgical techniques, minimizing human factors such as stress or fatigue, and allowing less experienced surgeons to handle more complex cases by facilitating the procedure~\cite{han_systematic_2022}. This in turn would allow trainees to focus less on steering the device and more on the critical factors of the procedure~\cite{fagogenis_autonomous_2019}. Autonomy can provide more consistency and precision in surgeries, improving efficiency, safety, and potentially reducing complication rates~\cite{schmidgall_will_2025}. Although task autonomy provides many benefits, it does not eliminate the need for manual navigation. When autonomous behavior is used in clinics, it is under the supervision or in collaboration with surgeons~\cite{lee_levels_2024,dupont_grand_2025}, who must be able to take over at any time in the event of unforeseen conditions or emergencies. In these situations, having an intuitive control over the tool is crucial. Therefore, we also introduce a manual navigation strategy that allows the surgeon to freely follow and ablate individual vessels (see Figure~\ref{fig:in_vivo_nav_manual} and Figure~\ref{fig:vessel_ablation}B-H). Together, the four navigation strategies implemented in this work cover the main surgeons' needs during an FLC procedure. These include the possibility of manually following and ablating individual vessels, autonomously exploring an area, and navigating back to specific targets within or outside of the current endoscopic view for point ablations (see Figure~\ref{fig:in_vivo_nav_manual} and Figure~\ref{fig:in_vivo_nav_automated}). Beyond their relevance to FLC, these strategies lend themselves to advancing navigation capabilities in MIS using flexible instruments. 

Although the limited size of the mosaic achieved in vivo limited the distance of the long-range navigation motions that we could demonstrate (see Figure~\ref{sfig:mosaic_w_all_targets}), we did not observe any influence of the distance between the starting and target positions on the final pixel error (see Figure~\ref{sfig:in_vivo_jumping}). The larger long-range motions achieved in vitro in a wet stomach model and ex vivo in a porcine bladder using larger mosaics (Figure~\ref{sfig:cysto_jump} and Figure~\ref{sfig:tne_jump}, and Movies~S11-12 and Movies~S16-17) are in line with this observation. Future work should focus on improving mosaicking algorithms to enable the generation of larger mosaics, improving their robustness to occlusions such as a moving fetus, and providing reliable localization of the device within the mosaic to handle large patient movements. In addition, the development and integration of a robust, real-time 3D reconstruction algorithm would further enhance the versatility of the long-range navigation.

Our experiments show that robotic control significantly improves accuracy and precision during endoscopic laser ablation procedures (see Figure~\ref{fig:vessel_ablation} and Figure~\ref{fig:user_study}C and F) as it eliminates hand tremors, leading to a reduced laser exposure time and energy. Although physiological movements are also expected to hinder ablation quality and lead to the ablation of healthy tissue, they remained limited during in vivo experiments and could be manually corrected. In the case of FLC, the preservation of healthy placental tissue is critical, as placental damage has been associated with serious complications, including preterm premature rupture of membranes (PPROM)~\cite{akkermans_what_2017} and preterm birth before 32~weeks of gestation~\cite{dezoysa_membrane_2020}. Reducing laser energy and enhancing its efficacy may, therefore, lead to better perinatal outcomes~\cite{spruijt_twin-twin_2020}. Furthermore, reducing invasiveness by having the robotic fetoscope fixed at the insertion site while moving its tip can further contribute to decreasing the rate of PPROM after FLC by limiting the friction and shear forces of the instrument port with the fetal membranes~\cite{amberg_why_2021,beck_preterm_2012}. 

Our user study showed an improvement in both usability and performance using a robotic approach over a manual approach. Users manipulating the robotic fetoscope reported a significant reduction in physical demand, effort, and frustration compared to the conventional tool (see Figure~\ref{fig:user_study}D and G). Statistically significant reduction in mental load was only obtained for the trajectory ablation (see Figure~\ref{fig:user_study}G). This is attributed to difficulties in estimating the depth required for the target during the first task, which was a challenge for many participants. Reduction of the mental and physical load on physicians and improvement of ergonomics are among the motivations and expected benefits of incorporating surgical robots into clinical practice~\cite{han_systematic_2022, wee_systematic_2020}. Our results indicate that even with limited training, the robotic platform is intuitive and easy to handle. Future work includes the development of distance detection and the integration of depth information into automation to further improve usability. Although this study was simulated for tasks encountered in FLC specifically, its findings emphasize the promise of our platform to facilitate MIS and reduce the mental and physical burden on physicians in endoscopic procedures in general.

Magnetic actuation represents one of several viable approaches for designing dexterous tools capable of navigating in open space~\cite{ahmad_development_2023, cruz-martinez_flexible_2023, dwyer_continuum_2017}. The navigation strategies we propose have the potential to be readily adapted to various actuation mechanisms beyond magnetics because they are exclusively image-based. Visual servoing and recent progress in surgical computer vision have shown promise for the control of rigid and flexible continuum robots in challenging surgical scenes~\cite{nazari_visual_2022,azizian_visual_2014,azizian_visual_2015}. Our methods are model-free, so they can be readily applied to various devices, including flexible instruments, which are generally more challenging to control. They also do not require the integration of additional sensors to infer the position and shape of the device, which contributes to keeping the dimensions of the instruments as small as possible. In addition, image-based processing is robust to different surgical environments, as the approaches were designed for low-texture environments rather than focusing on the placental vasculature.

The feasibility and safety of our platform was demonstrated in vivo. By advancing the technical readiness of this approach for FLC, this work takes a substantial step toward its clinical translation for fetal surgery. The assistive functions provided by the platform were designed to address critical and mentally demanding aspects of the procedure while maintaining the engagement and oversight of the surgeon. The combination of endoscopic image mosaics with autonomous navigation capabilities exhibits potential benefits for MIS in other anatomical regions such as the stomach and bladder, where endoscopic image mosaicking and 3D reconstructions have already been proposed to improve visualization and localization of instruments during cancer screenings~\cite{lurie_3d_2017, phan_optical_2020,widya_stomach_2021}. The feasibility of the proposed manual and automated navigation methods for these procedures is demonstrated in vitro and ex vivo.

\section*{Materials and Methods}

\subsection*{Fetoscope workspace characterization}

The robotic fetoscope was mounted at 150~mm in front of the eMNS coils, which generated a magnetic field of 25~mT (see Figure~\ref{sfig:navion}A). The magnetic field was rotated in a sweeping motion in different planes, while the tip of the robotic fetoscope was tracked with an optical tracking system (OTS, Vicon Motion Systems Ltd). Unstable configurations of the fetoscope under this rotating field described in~\cite{tunay_modeling_2004,peyron_kinematic_2018} were removed from the data in post-processing. Transitions from stable to unstable configurations were detected by comparing the measured tip acceleration to a pre-defined threshold. The poses depicted in Figure~\ref{fig:workspace}E were filtered with a Savitzky-Golay filter.

\subsection*{Accessible uterus location simulation}

The accessible locations of the uterus and the ablation angles shown in Figure~\ref{fig:workspace}F were determined in two steps. First, all reachable poses in free space were determined for each device for a given insertion pose. For the robotic fetoscope the poses were interpolated based on the measurements shown in Figure~\ref{fig:workspace}E and for the conventional, rigid fetoscope (11508AAK, Karl Storz) they were calculated over a constant curvature model with a radius of 210~mm. The different insertion poses were then calculated by sweeping over the insertion angles~$\psi \in [0,90\degree]$ and $\varphi \in [0,90\degree]$ and the insertion depth~$l$. Secondly, the poses were converted to rays simulating the laser light during ablation. The intersections with the triangles of a mesh uterus model at 26~weeks gestation were found over ray casting. Rays were discarded if the distance~$d$, from their source to their intersection with the mesh was outside the range~$ [3,10]$~mm. The angle~$\alpha$ between the remaining rays and their tangents at the intersection point were then calculated. Figure~\ref{fig:workspace}F shows the highest~$\alpha$ per triangle, which was filtered with a low-pass filter. In clinical practice, the range of motion~$(\psi, \varphi)$ may be limited by anatomical constraints, such as maternal hip bones or the pelvic floor, which were neglected in our simulations. 

\subsection*{Electromagnetic navigation}

When the robotic endoscope is placed in an external magnetic field $\mathbf{b}$ a magnetic torque $\boldsymbol{\tau}$ is exerted on each magnet comprised in the tool

\begin{equation}
    \boldsymbol{\tau}=\mathbf{m} \times \mathbf{b}
\end{equation}

\noindent
where $\mathbf{m}$ is the magnetic dipole moment of each magnet that causes its distal section to bend. To control the endoscope tip with a magnetic field $\mathbf{b}$ at a position $\mathbf{p}$ in space, we used the Navion eMNS~\cite{gervasoni_human-scale_2024}. It was composed of three electromagnets which currents $\mathbf{i}\in \mathbb{R}^3$ were calculated over the linear mapping:
\begin{equation}
    \mathbf{i}=\mathbf{A}^{\dagger}(\mathbf{p}) \mathbf{b}(\mathbf{p})
\end{equation}

\noindent where $\mathbf{A}^{\dagger}$ is the Moore-Penrose pseudoinverse of the actuation matrix, which was found by calibration~\cite{petruska_model-based_2017}.

\subsection*{Image-based control}
\label{s:image_based_control}

The proposed navigation strategies control the 2D motion~$\dot{\mathbf{s}}$ of a point~${\mathbf{s}}$ in the endoscopic image through eye-in-hand image-based visual servoing. To generate this motion, the orientation of magnetic field vector~$\mathbf{b}$ was updated at each time step with an intrinsic rotation $\dot{\mathbf{q}}= [\dot{\alpha}~ \dot{\beta}]^T$ around the reference frame attached to $\mathbf{b}$ such that ${}_B\mathbf{b} = [b~0~0]^T$. It was first rotated around its $y^B$-axis with $\dot{\beta}$ and then around its $z^B$-axis with $\dot{\alpha}$ (see Figure~\ref{sfig:navion}B). The desired 2D motion~$\dot{\mathbf{s}}$ of a point~${\mathbf{s}}$ in the endoscopic image was then mapped to the instantaneous rotation of the magnetic field vector as follows

\begin{equation}
    \dot{\mathbf{s}}=\mathbf{J} \dot{\mathbf{q}}
\label{eq:manual_nav}
\end{equation}

\noindent
The image Jacobian matrix~$\mathbf{J}$ was initially estimated with a calibration process. It was then updated at each time step~$k$ using the Broyden method~\cite{wu_model-free_2015} with
\begin{equation}
    \hat{\mathbf{J}}_{k}= \hat{\mathbf{J}}_{k-1} + \gamma \frac{\left(\Delta \hat{\mathbf{s}}-\hat{\mathbf{J}}_{k-1} \Delta \mathbf{q}\right) \Delta \mathbf{q}^{\mathrm{T}}}{\left\|\Delta \mathbf{q}^{\mathrm{T}} \Delta \mathbf{q}\right\|^{2}}
\end{equation}

\noindent where $\hat{\mathbf{J}}_{k-1}$ and $\hat{\mathbf{J}}_{k}$ are the estimates of $\mathbf{J}$ at the time steps~$k-1$ and $k$, $\gamma$ is the update rate, $\Delta{\mathbf{q}}$ is the magnetic field rotation between the time steps $k-1$ and $k$, and $\Delta \hat{\mathbf{s}}$ is the estimated motion of the point~$\mathbf{s}$ per time step $\Delta t$ and must not be noisy~\cite{bonkovi_new_2006}. This estimated motion was defined as
\begin{equation}
\Delta \hat{\mathbf{s}} = {}_k\mathbf{s}_{k} -  {}_{k}\mathbf{s}
\label{eq:delta_s}
_{k-1}\end{equation}

\noindent describing the change in the image coordinates of $\mathbf{s}$ between the steps $k-1$ and $k$ (right subscript) in the frame of the current image $I_k$ (left subscript). At each new step, the desired magnetic field change $\Delta{\mathbf{q}}_d$ was computed to achieve the desired motion $\dot{\mathbf{s}}_d$ as 

\begin{equation}
\Delta{\mathbf{q}}_d=\hat{\mathbf{J}}_k^{-1} \Delta {\mathbf{s}}_d
\end{equation}

\noindent where $\Delta {\mathbf{s}}_d = \dot{\mathbf{s}}_d  \Delta t$ and assuming that the estimated Jacobian has full rank.

The estimated motion~$\Delta \hat{\mathbf{s}}$ in the camera image originated either from the motion of the robotic endoscope or from physiological motion. The latter was negligible compared to the endoscope's movements provided that the endoscope motion is sufficiently fast. To ensure that $\mathbf{J}$ was only updated when the endoscope was moving, the estimated speed~$\Delta \hat{{s}} = \left\| \Delta \hat{\mathbf{s}} \right\|$ had to exceed a predefined threshold.

The Jacobian update method presented in our previous work~\cite{lussi_magnetically_2022} was restricted to updating only the directional component, was less robust, and relied on tracking specific targets in the endoscopic images.

\subsection*{Manual control}\label{s:manual_control}

We introduced a manual navigation strategy that allowed the user to steer the robotic endoscope image using a joystick. The user controlled the motion of the image center of coordinates~$[x_c ~ y_c]^T$ in the frame of the current image. The desired motion~$\Delta \mathbf{s}_d$ had a fixed speed and its direction matched the direction of the joystick controlled by the user. The estimated motion~$\Delta \hat{\mathbf{s}}$ was defined as the difference between~${}_{k} \mathbf{s}_{k} =[x_c ~ y_c]^T$, and the feature at the center of the previous image projected onto the current frame ${}_{k} \mathbf{s}_{k-1}$ (see equation~\ref{eq:delta_s}) that was obtained here as
\begin{equation}
    \lambda
     \begin{bmatrix}
    {}_k \mathbf{s}_{k-1} \\
    1
    \end{bmatrix}
    =\mathbf{H}_{k,k-1}
    \begin{bmatrix}
     x_c \\
     y_c \\
    1
    \end{bmatrix}
    \label{eq:homography_for_manual_control}
\end{equation}
\noindent where $\lambda$ is a scale factor and $\mathbf{H}_{k,k-1} \in \mathbb{R}^{2 \times 2}$ is the homography projecting $I_{k-1}$ onto $I_{k}$ and calculated based on the optical flow of the pixels according to~\cite{Alabi2022}. To avoid inaccurate matches between pixels in the images due to reflections or insufficient illumination, the images were thresholded and masked before estimating $\mathbf{H}_{k,k-1}$. The images were converted to grayscale and pixel values above a predefined threshold were masked to remove reflections, the influence of the guide light or laser light, and below a second threshold in case the illumination was insufficient in some areas.

\subsection*{Short-range automated control}

For the short-range automated navigation the user selected a target location ${}_{k}\mathbf{s} = [x_k ~ y_k]^T$ 
in the current endoscopic image $I_k$ and the robotic endoscope automatically moved it to the reference position ${}_k \mathbf{r}$ in the image (typically located in the center of the image). The desired motion $\Delta\mathbf{s}_d$ was defined as the output of a proportional–integral–derivative (PID) controller over the error between the reference position and the target position $\mathbf{e}= {_{k}\mathbf{r} - {}_{k}\mathbf{s}}$ as follows

\begin{equation}
   \Delta\mathbf{s}_d = 
    (
    K_p\mathbf{e}
    +
    K_i \int \mathbf{e} \mathrm{d}t
    +
    K_d 
    \dot{\mathbf{e}}
    )
\end{equation}

\noindent where the control parameters $K_p$, $K_i$ and $K_d$ were found by empirical tuning. The estimated actual motion~$\Delta \hat{\mathbf{s}}$ was defined as the difference in target locations in images $I_{k}$ and $I_{k-1}$. The PID parameters of the controller were initially tuned in vitro to achieve a fast response and minimal overshoot, and the same parameters were applied during in vivo experiments without further adjustment.

\subsection*{Long-range automated control}

The long-range automated control guided the robotic endoscope to targets selected in a previously generated image mosaic, which were typically outside of the endoscopic view. Mosaics were generated using automated spiral or grid trajectory in the mosaic frame (see Figure~\ref{sfig:in_vivo_automated_exploration}).

First, the magnetic field~$\hat{\mathbf{b}}$ at the target location was estimated by interpolating previous field orientations, visual servoing was then used to navigate the remaining distance. If the target was not visible in the endoscopic image after interpolation, waypoints were computed as long as $I_k$ overlaps with the image from which the target originates. The endoscope was then moved through these waypoints until the target lied near the reference position. If the patient had substantially moved, such that no image overlap remained, the failure was detected, the user was notified, and the mosaic could be recreated. A detailed description of the approach is available in Supplementary Methods.

\subsection*{In vitro navigation characterization}

To characterize the performance of the image Jacobian update and the visual servoing in Figure~\ref{fig:pov_eval} and~Figure~\ref{sfig:detailed_nav_eval}, the robotic fetoscope was mounted horizontally at 145~mm from the eMNS, which generated a magnetic field of 15~mT. The inputs consisted of stepwise direction changes with 90$\degree$ steps of 18~s in duration, or of a continuous direction change over time with a rate of change of 6.3~deg/s. The magnetic field magnitude was kept constant. For each input setting, 5 runs were performed from the same starting position for visual servoing while updating the image Jacobian matrix~$\mathbf{J}$, and another 5 runs keeping it constant. Before each run, the matrix~$\mathbf{J}$ was calibrated at the starting position.

\subsection*{Ex vivo ablations}

The placenta used for ex vivo ablation (Figure~\ref{fig:vessel_ablation}A and Figure~\ref{sfig:laser_eval_supp_results_1}) was placed in a saline solution to mimic amniotic fluid (see Figure~\ref{sfig:ex_vivo_placenta_setups}). Vessel ablations were performed by a trained user with a clinical Nd:Yag laser (My 60, KLS Martin, Austria) and a 300~$\mu m$ core laser fiber (KLS Martin, Austria). Two setups (see Figure~\ref{sfig:ex_vivo_placenta_setups} and Supplementary Methods) were used to assess vessel ablations on an ex vivo human placenta. The first used a hand-held, rigid fetoscope to simulate a manual ablation, and the second used a fixed fiber to simulate robotic ablations.

 First, the minimal required laser power and duration for a complete ablation were determined empirically for each vessel diameter using the fixed laser fiber setup. The same settings were subsequently used for all ablations of vessels with the same diameters in both setups. Ablation completeness was analyzed according to the protocol proposed in~\cite{lopriore_accurate_2011}, where colored dye was injected through the umbilical cord into the placental vessels (see Figure~\ref{sfig:ex_vivo_placenta_setups}C). If a vessel was fully ablated, the colored dye was not able to penetrate the vessel further than the location of the ablation. This method is routinely performed for placenta analysis after birth of TTTS patients~\cite{bamberg_update_2019, spruijt_twin-twin_2020}.

The placenta was donated after delivery, and written informed consent was obtained. The experiment was approved by the Ethics Committee of the District of Zürich, Switzerland under the number BASEC-Nr: 2023-00110.

\subsection*{In vivo experiment}

The in vivo experiment was conducted in an ovine model. Ovine models are well-established models in fetal and reproductive surgery due to their physiological similarities in neonatal development and placenta with humans~\cite{lussier_sheep_2025, kabagambe_lessons_2018}. The experimental setup is depicted in Figure~\ref{sfig:in_vivo_setup}. One pregnant ewe (black mountain sheep, 50~kg, 90-100~days of gestation, singleton pregnancy) was placed under general anesthesia and placed on a surgical table with the eMNS placed lateral to the operating table as close to the abdomen as possible. Throughout the procedure, a veterinarian team closely monitored the sheep, and the surgery itself was performed by two experienced fetal surgeons. A detailed description of the approach is available in Supplementary Methods.

The robotic fetoscope was pushed in the amniotic sac through an anteriorly placed trocar (Cook Medical, US) and navigation and ablation experiments were performed (shown in Figure~\ref{fig:in_vivo_nav_manual}-\ref{fig:vessel_ablation} and Movies~S1-6). The fetoscope motion was monitored through a second endoscopic camera inserted on the side of the uterus, providing an external view of the fetoscope during navigation and ablation (see Figures~\ref{fig:vessel_ablation}B, Figure~\ref{sfig:fetoscope_dexterity_in_vivo} and Figure~\ref{sfig:in_vivo_setup}D-E, and Movie~S9).

No complications arose during the experiments. Subsequently, the fetus and the ewe were euthanized and the fetus was examined by an experienced veterinarian and it was confirmed that it was healthy and did not suffer any injuries due to the procedure. The animal study was approved by the local Committee for Animal Experimental Research (Cantonal Veterinary Office Zurich, Switzerland) under the license number ZH117/2023.

\subsection*{Usability study}

The augmented reality setup for the usability study is depicted in Figure~\ref{sfig:study_setup}. Of the 11 participants in the usability study, no one had prior experience steering the robotic fetoscope, five people had 0.5-5~years of experience in magnetic catheter and endoscope navigation with other devices and steering algorithms, and one fetal surgeon had with 25~years of experience in endoscopy. Each participant completed the user study in a single session using the same setup. Informed written consent was given after confirming eligibility.

Each task consisted of a training period, which was followed by three test runs, separated by short breaks and the completion of a NASA-TLX questionnaire. For the first task, the training and testing periods were 4~min and 11~min and for the second task they were 2~min and 5~min, respectively. The protocol was repeated for the second device. The order of the devices was alternated between the participants. The protocol is shown in Figure~\ref{sfig:study_protocol}, and details for the statistical analysis of the results are provided in Supplementary Methods. 

\subsection*{Statistical analysis}

The statistical significance in the results of the user study was obtained over a paired Wilcoxon signed-rank test. The degree of statistical significance is illustrated in the figures with star annotations, where one star corresponds to $p<0.05$, two stars represent $p<0.01$ and three stars $p<0.001$.

%%%%%%%%%%%%%%%% REFERENCES %%%%%%%%%%%%%%%

\bibliography{ref_mattille_scirob} 

@article{widya_stomach_2021,
	title = {Stomach {3D} {Reconstruction} {Using} {Virtual} {Chromoendoscopic} {Images}},
	volume = {9},
	issn = {21682372},
	url = {http://dx.doi.org/10.1109/JTEHM.2021.3062226},
	doi = {10.1109/JTEHM.2021.3062226},
	urldate = {2023-08-09},
	journal = {IEEE Journal of Translational Engineering in Health and Medicine},
	publisher = {Institute of Electrical and Electronics Engineers Inc.},
	author = {Widya, Aji Resindra and Monno, Yusuke and Okutomi, Masatoshi and Suzuki, Sho and Gotoda, Takuji and Miki, Kenji},
	year = {2021},
	pages = {1},
}

@article{li_bladder_2024,
	title = {Bladder image stitching algorithm for navigation and referencing using a standard cystoscope},
	volume = {14},
	copyright = {2024 This is a U.S. Government work and not under copyright protection in the US; foreign copyright protection may apply},
	issn = {2045-2322},
	url = {https://www.nature.com/articles/s41598-024-80284-7},
	doi = {10.1038/s41598-024-80284-7},
	language = {en},
	number = {1},
	urldate = {2025-12-16},
	journal = {Scientific Reports},
	publisher = {Nature Publishing Group},
	author = {Li, Ming and Varble, Nicole A. and Gurram, Sandeep and Long, Dilara and Valera, Vladimir and Gopal, Nikhil and Bakhutashvili, Ivane and Reed, Sheridan and Pritchard, William F. and Karanian, John W. and Xu, Sheng and Wood, Bradford J.},
	month = nov,
	year = {2024},
	pages = {29168},
}

@article{lee_urine_2022,
	title = {Urine exosomes as biomarkers in bladder cancer diagnosis and prognosis: {From} functional roles to clinical significance},
	volume = {12},
	issn = {2234-943X},
	shorttitle = {Urine exosomes as biomarkers in bladder cancer diagnosis and prognosis},
	url = {https://www.frontiersin.org/journals/oncology/articles/10.3389/fonc.2022.1019391/full},
	doi = {10.3389/fonc.2022.1019391},
	language = {English},
	urldate = {2026-02-02},
	journal = {Frontiers in Oncology},
	publisher = {Frontiers},
	author = {Lee, Nicholas and Canagasingham, Ashan and Bajaj, Mohit and Shanmugasundaram, Ramesh and Hutton, Anthony and Bucci, Joseph and Graham, Peter and Thompson, James and Ni, Jie},
	month = sep,
	year = {2022},
}

@article{pena_usefulness_2022,
	title = {Usefulness of the {Urine} {Methylation} {Test} ({Bladder} {EpiCheck}®) in {Follow}-{Up} {Patients} with {Non}-{Muscle} {Invasive} {Bladder} {Cancer} and {Cytological} {Diagnosis} of {Atypical} {Urothelial} {Cells}—{An} {Institutional} {Study}},
	volume = {11},
	copyright = {http://creativecommons.org/licenses/by/3.0/},
	issn = {2077-0383},
	url = {https://www.mdpi.com/2077-0383/11/13/3855},
	doi = {10.3390/jcm11133855},
	language = {en},
	number = {13},
	urldate = {2026-02-02},
	journal = {Journal of Clinical Medicine},
	publisher = {publisher},
	author = {Peña, Karla B. and Riu, Francesc and Hernandez, Anna and Guilarte, Carmen and Badia, Joan and Parada, David},
	month = jul,
	year = {2022},
}

@article{rindorf_market_2021,
	title = {Market {Readiness} for {Single}-{Use} {Cystoscopes} {According} to {Urologists} and {Procurement} {Managers} {Worldwide}},
	volume = {13},
	issn = {null},
	url = {https://www.tandfonline.com/doi/abs/10.2147/RRU.S309602},
	doi = {10.2147/RRU.S309602},
	urldate = {2026-02-02},
	journal = {Research and Reports in Urology},
	publisher = {Dove Medical Press},
	author = {Rindorf, Dinah and Larsen, Sara and Ockert, Lotte and Jung, Helene and Dahl, Claus},
	month = may,
	year = {2021},
	note = {\_eprint: https://www.tandfonline.com/doi/pdf/10.2147/RRU.S309602},
	pages = {221--226},
}

@article{emura_principles_2020,
	title = {Principles and practice to facilitate complete photodocumentation of the upper gastrointestinal tract: {World} {Endoscopy} {Organization} position statement},
	volume = {32},
	copyright = {© 2019 Japan Gastroenterological Endoscopy Society},
	issn = {1443-1661},
	shorttitle = {Principles and practice to facilitate complete photodocumentation of the upper gastrointestinal tract},
	url = {https://onlinelibrary.wiley.com/doi/abs/10.1111/den.13530},
	doi = {10.1111/den.13530},
	language = {en},
	number = {2},
	urldate = {2026-02-01},
	journal = {Digestive Endoscopy},
	author = {Emura, Fabian and Sharma, Prateek and Arantes, Vitor and Cerisoli, Cecilio and Parra-Blanco, Adolfo and Sumiyama, Kazuki and Araya, Raul and Sobrino, Sergio and Chiu, Philip and Matsuda, Koji and Gonzalez, Robinson and Fujishiro, Mitsuhiro and Tajiri, Hisao},
	year = {2020},
	note = {\_eprint: https://onlinelibrary.wiley.com/doi/pdf/10.1111/den.13530},
	pages = {168--179},
}

@article{januszewicz_quality_2020,
	title = {Quality indicators in diagnostic upper gastrointestinal endoscopy},
	volume = {13},
	issn = {1756-283X},
	url = {https://pmc.ncbi.nlm.nih.gov/articles/PMC7232050/},
	doi = {10.1177/1756284820916693},
	urldate = {2026-02-01},
	journal = {Therapeutic Advances in Gastroenterology},
	author = {Januszewicz, Wladyslaw and Kaminski, Michal F.},
	month = may,
	year = {2020},
	pages = {1756284820916693},
}

@article{kurniawan_flexible_2017,
	title = {Flexible {Gastro}-intestinal {Endoscopy} — {Clinical} {Challenges} and {Technical} {Achievements}},
	volume = {15},
	issn = {2001-0370},
	url = {https://www.sciencedirect.com/science/article/pii/S2001037016300617},
	doi = {10.1016/j.csbj.2017.01.004},
	urldate = {2026-02-02},
	journal = {Computational and Structural Biotechnology Journal},
	author = {Kurniawan, Niehls and Keuchel, Martin},
	month = jan,
	year = {2017},
	pages = {168--179},
}

@article{dwyer_continuum_2017,
	title = {A {Continuum} {Robot} and {Control} {Interface} for {Surgical} {Assist} in {Fetoscopic} {Interventions}},
	volume = {2},
	issn = {2377-3766},
	url = {https://ieeexplore.ieee.org/document/7874155/},
	doi = {10.1109/LRA.2017.2679902},
	number = {3},
	urldate = {2025-11-25},
	journal = {IEEE Robotics and Automation Letters},
	author = {Dwyer, George and Chadebecq, Francois and Amo, Marcel Tella and Bergeles, Christos and Maneas, Efthymios and Pawar, Vijay and Poorten, Emanuel Vander and Deprest, Jan and Ourselin, Sebastien and De Coppi, Paolo and Vercauteren, Tom and Stoyanov, Danail},
	month = jul,
	year = {2017},
	pages = {1656--1663},
}

@article{zhu_intelligent_2021,
	title = {Intelligent {Soft} {Surgical} {Robots} for {Next}-{Generation} {Minimally} {Invasive} {Surgery}},
	volume = {3},
	copyright = {© 2021 The Authors. Advanced Intelligent Systems published by Wiley-VCH GmbH},
	issn = {2640-4567},
	url = {https://onlinelibrary.wiley.com/doi/abs/10.1002/aisy.202100011},
	doi = {10.1002/aisy.202100011},
	language = {en},
	number = {5},
	urldate = {2025-11-18},
	journal = {Advanced Intelligent Systems},
	author = {Zhu, Jiaqi and Lyu, Liangxiong and Xu, Yi and Liang, Huageng and Zhang, Xiaoping and Ding, Han and Wu, Zhigang},
	year = {2021},
	note = {\_eprint: https://advanced.onlinelibrary.wiley.com/doi/pdf/10.1002/aisy.202100011},
	pages = {2100011},
}

@article{nelson_magnetically_2022,
	title = {Magnetically {Actuated} {Medical} {Robots}: {An} in vivo {Perspective}},
	volume = {110},
	issn = {1558-2256},
	shorttitle = {Magnetically {Actuated} {Medical} {Robots}},
	url = {https://ieeexplore.ieee.org/document/9764710/},
	doi = {10.1109/JPROC.2022.3165713},
	number = {7},
	urldate = {2026-01-04},
	journal = {Proceedings of the IEEE},
	author = {Nelson, Bradley J. and Gervasoni, Simone and Chiu, Philip W. Y. and Zhang, Li and Zemmar, Ajmal},
	month = jul,
	year = {2022},
	pages = {1028--1037},
}

@article{mesot_teleoperated_2025,
	title = {Teleoperated {Magnetic} {Endoscopy}: {A} {Case} {Study} and {Perspective}},
	volume = {7},
	copyright = {© 2024 The Author(s). Advanced Intelligent Systems published by Wiley-VCH GmbH},
	issn = {2640-4567},
	shorttitle = {Teleoperated {Magnetic} {Endoscopy}},
	url = {https://onlinelibrary.wiley.com/doi/abs/10.1002/aisy.202400522},
	doi = {10.1002/aisy.202400522},
	language = {en},
	number = {10},
	urldate = {2026-02-01},
	journal = {Advanced Intelligent Systems},
	author = {Mesot, Alexandre and Mattille, Michelle and Boehler, Quentin and Schmid, Nina and Lyttle, Sean and Heemeyer, Florian and Chan, Shannon Melissa and Chiu, Philip Wai Yan and Nelson, Bradley James},
	year = {2025},
	note = {\_eprint: https://advanced.onlinelibrary.wiley.com/doi/pdf/10.1002/aisy.202400522},
	pages = {2400522},
}

@article{VanDerVeeken2019,
	title = {Laser for twin-to-twin transfusion syndrome: a guide for endoscopic surgeons.},
	volume = {11},
	issn = {2032-0418},
	url = {http://www.ncbi.nlm.nih.gov/pubmed/32082525},
	number = {3},
	urldate = {2020-06-03},
	journal = {Facts, Views \& Vision in ObGyn},
	publisher = {Vlaamse Vereniging voor Obstetrie en Gynaecologie},
	author = {Van Der Veeken, L and Couck, I and Van Der Merwe, J and De Catte, L and Devlieger, R and Deprest, J and Lewi, L},
	month = sep,
	year = {2019},
	pages = {197--205},
}

@article{hernansanz_robot_2024,
	title = {Robot assisted {Fetoscopic} {Laser} {Coagulation}: {Improvements} in navigation, re-location and coagulation},
	volume = {147},
	issn = {0933-3657},
	shorttitle = {Robot assisted {Fetoscopic} {Laser} {Coagulation}},
	url = {https://www.sciencedirect.com/science/article/pii/S0933365723002397},
	doi = {10.1016/j.artmed.2023.102725},
	urldate = {2026-02-11},
	journal = {Artificial Intelligence in Medicine},
	author = {Hernansanz, Albert and Parra, Johanna and Sayols, Narcís and Eixarch, Elisenda and Gratacós, Eduard and Casals, Alícia},
	month = jan,
	year = {2024},
	pages = {102725},
}

@article{bamberg_update_2019,
	title = {Update on twin-to-twin transfusion syndrome},
	volume = {58},
	issn = {1521-6934},
	url = {https://www.sciencedirect.com/science/article/pii/S1521693418302426},
	doi = {10.1016/j.bpobgyn.2018.12.011},
	urldate = {2025-08-28},
	journal = {Best Practice \& Research Clinical Obstetrics \& Gynaecology},
	author = {Bamberg, Christian and Hecher, Kurt},
	month = jul,
	year = {2019},
	pages = {55--65},
}

@article{phan_optical_2020,
	title = {Optical flow-based structure-from-motion for the reconstruction of epithelial surfaces},
	volume = {105},
	issn = {0031-3203},
	url = {https://www.sciencedirect.com/science/article/pii/S0031320320301941},
	doi = {10.1016/j.patcog.2020.107391},
	urldate = {2025-12-16},
	journal = {Pattern Recognition},
	author = {Phan, Tan-Binh and Trinh, Dinh-Hoan and Wolf, Didier and Daul, Christian},
	month = sep,
	year = {2020},
	pages = {107391},
}

@article{mesot_parametric_2025,
	title = {Parametric {Design} of {Continuum} {Robots} {Using} {Interlocking} {Ball} {Joints}},
	volume = {7},
	copyright = {© 2025 The Author(s). Advanced Intelligent Systems published by Wiley-VCH GmbH},
	issn = {2640-4567},
	url = {https://onlinelibrary.wiley.com/doi/abs/10.1002/aisy.202500180},
	doi = {10.1002/aisy.202500180},
	language = {en},
	number = {11},
	urldate = {2025-07-23},
	journal = {Advanced Intelligent Systems},
	author = {Mesot, Alexandre and Boehler, Quentin and Heemeyer, Florian and Lyttle, Sean and Aktaş, Buse and Nelson, Bradley J.},
	year = {2025},
	note = {\_eprint: https://advanced.onlinelibrary.wiley.com/doi/pdf/10.1002/aisy.202500180},
	pages = {2500180},
}

@article{Alabi2022,
	title = {Robust fetoscopic mosaicking from deep learned flow fields},
	issn = {1861-6429},
	url = {https://link.springer.com/article/10.1007/s11548-022-02623-1},
	doi = {10.1007/S11548-022-02623-1},
	urldate = {2022-05-17},
	journal = {International Journal of Computer Assisted Radiology and Surgery 2022},
	publisher = {Springer},
	author = {Alabi, Oluwatosin and Bano, Sophia and Vasconcelos, Francisco and David, Anna L. and Deprest, Jan and Stoyanov, Danail},
	month = may,
	year = {2022},
	pages = {1125--1134},
}

@article{beck_preterm_2012,
	title = {Preterm {Prelabor} {Rupture} of {Membranes} and {Fetal} {Survival} after {Minimally} {Invasive} {Fetal} {Surgery}: {A} {Systematic} {Review} of the {Literature}},
	volume = {31},
	issn = {1015-3837},
	shorttitle = {Preterm {Prelabor} {Rupture} of {Membranes} and {Fetal} {Survival} after {Minimally} {Invasive} {Fetal} {Surgery}},
	url = {https://doi.org/10.1159/000331165},
	doi = {10.1159/000331165},
	number = {1},
	urldate = {2025-08-25},
	journal = {Fetal Diagnosis and Therapy},
	author = {Beck, Veronika and Lewi, Paul and Gucciardo, Leonardo and Devlieger, Roland},
	year = {2012},
	pages = {1--9},
}

@article{dezoysa_membrane_2020,
	title = {Membrane {Separation} and {Perinatal} {Outcomes} after {Laser} {Treatment} for {Twin}-{Twin} {Transfusion} {Syndrome}},
	volume = {47},
	issn = {1015-3837},
	url = {https://doi.org/10.1159/000504361},
	doi = {10.1159/000504361},
	number = {4},
	urldate = {2025-08-28},
	journal = {Fetal Diagnosis and Therapy},
	author = {De Zoysa, Madushka Y. and Chon, Andrew H. and Korst, Lisa M. and Llanes, Arlyn and Chmait, Ramen H.},
	year = {2020},
	pages = {307--314},
}

@article{lopriore_accurate_2011,
	title = {Accurate and {Simple} {Evaluation} of {Vascular} {Anastomoses} in {Monochorionic} {Placenta} using {Colored} {Dye}},
	volume = {55},
	issn = {1940-087X},
	url = {https://www.ncbi.nlm.nih.gov/pmc/articles/PMC3230184/},
	doi = {10.3791/3208},
	urldate = {2025-04-11},
	journal = {Journal of Visualized Experiments : JoVE},
	author = {Lopriore, Enrico and Slaghekke, Femke and Middeldorp, Johanna M. and Klumper, Frans J. and van Lith, Jan M. and Walther, Frans J. and Oepkes, Dick},
	month = sep,
	year = {2011},
	pages = {3208},
}

@article{nazari_visual_2022,
	title = {Visual servoing of continuum robots: {Methods}, challenges, and prospects},
	volume = {18},
	issn = {1478-596X},
	shorttitle = {Visual servoing of continuum robots},
	url = {https://onlinelibrary.wiley.com/doi/abs/10.1002/rcs.2384},
	doi = {10.1002/rcs.2384},
	language = {en},
	number = {3},
	urldate = {2025-08-29},
	journal = {The International Journal of Medical Robotics and Computer Assisted Surgery},
	author = {Nazari, Ali A. and Zareinia, Kourosh and Janabi-Sharifi, Farrokh},
	year = {2022},
	note = {\_eprint: https://onlinelibrary.wiley.com/doi/pdf/10.1002/rcs.2384},
	pages = {e2384},
}

@article{bouchghoul_management_2025,
	title = {Management of twin‐to‐twin transfusion syndrome: update and current challenges},
	volume = {7},
	issn = {2589-9333},
	shorttitle = {Management of twin‐to‐twin transfusion syndrome},
	url = {https://www.sciencedirect.com/science/article/pii/S2589933325001144},
	doi = {10.1016/j.ajogmf.2025.101714},
	number = {8},
	urldate = {2025-10-08},
	journal = {American Journal of Obstetrics \& Gynecology MFM},
	author = {Bouchghoul, Hanane and Madar, Hugo and Mattuizzi, Aurélien and Coatleven, Frédéric and Vincienne, Marie and Prier, Perrine and Froeliger, Alizée and Sentilhes, Loïc},
	month = aug,
	year = {2025},
	pages = {101714},
}

@inproceedings{bonkovi_new_2006,
	title = {A new method for uncalibrated visual servoing},
	issn = {1943-6580},
	url = {https://ieeexplore.ieee.org/abstract/document/1631732},
	doi = {10.1109/AMC.2006.1631732},
	urldate = {2025-10-07},
	booktitle = {9th {IEEE} {International} {Workshop} on {Advanced} {Motion} {Control}, 2006.},
	author = {Bonkovi, M. and Hace, A. and Jezernik, K.},
	month = mar,
	year = {2006},
	pages = {624--629},
}

@article{martin_enabling_2020,
	title = {Enabling the future of colonoscopy with intelligent and autonomous magnetic manipulation},
	volume = {2},
	issn = {25225839},
	url = {http://dx.doi.org/10.1038/s42256-020-00231-9},
	doi = {10.1038/s42256-020-00231-9},
	number = {10},
	urldate = {2023-08-04},
	journal = {Nature Machine Intelligence},
	publisher = {Nature Publishing Group},
	author = {Martin, James W. and Scaglioni, Bruno and Norton, Joseph C. and Subramanian, Venkataraman and Arezzo, Alberto and Obstein, Keith L. and Valdastri, Pietro},
	month = oct,
	year = {2020},
	pages = {595--606},
}

@incollection{bano_chapter_2024,
	series = {The {MICCAI} {Society} book {Series}},
	title = {Chapter 15 - {Image} mosaicking},
	isbn = {978-0-12-813657-7},
	url = {https://www.sciencedirect.com/science/article/pii/B9780128136577000303},
	doi = {10.1016/B978-0-12-813657-7.00030-3},
	urldate = {2025-10-03},
	booktitle = {Medical {Image} {Analysis}},
	publisher = {Academic Press},
	author = {Bano, Sophia and Stoyanov, Danail},
	editor = {Frangi, Alejandro F. and Prince, Jerry L. and Sonka, Milan},
	month = jan,
	year = {2024},
	pages = {387--411},
}

@article{russo_continuum_2023,
	title = {Continuum {Robots}: {An} {Overview}},
	volume = {5},
	copyright = {© 2023 The Authors. Advanced Intelligent Systems published by Wiley-VCH GmbH},
	issn = {2640-4567},
	shorttitle = {Continuum {Robots}},
	url = {https://onlinelibrary.wiley.com/doi/abs/10.1002/aisy.202200367},
	doi = {10.1002/aisy.202200367},
	language = {en},
	number = {5},
	urldate = {2025-10-03},
	journal = {Advanced Intelligent Systems},
	author = {Russo, Matteo and Sadati, Seyed Mohammad Hadi and Dong, Xin and Mohammad, Abdelkhalick and Walker, Ian D. and Bergeles, Christos and Xu, Kai and Axinte, Dragos A.},
	year = {2023},
	note = {\_eprint: https://advanced.onlinelibrary.wiley.com/doi/pdf/10.1002/aisy.202200367},
	pages = {2200367},
}

@article{mao_magnetic_2024,
	title = {Magnetic steering continuum robot for transluminal procedures with programmable shape and functionalities},
	volume = {15},
	copyright = {2024 The Author(s)},
	issn = {2041-1723},
	url = {https://www.nature.com/articles/s41467-024-48058-x},
	doi = {10.1038/s41467-024-48058-x},
	language = {en},
	number = {1},
	urldate = {2025-10-03},
	journal = {Nature Communications},
	publisher = {Nature Publishing Group},
	author = {Mao, Liyang and Yang, Peng and Tian, Chenyao and Shen, Xingjian and Wang, Feihao and Zhang, Hao and Meng, Xianghe and Xie, Hui},
	month = may,
	year = {2024},
	pages = {3759},
}

@article{greenidge_harnessing_2025,
	title = {Harnessing the oloid shape in magnetically driven robots to enable high-resolution ultrasound imaging},
	volume = {10},
	url = {https://www.science.org/doi/full/10.1126/scirobotics.adq4198},
	doi = {10.1126/scirobotics.adq4198},
	number = {100},
	urldate = {2025-10-03},
	journal = {Science Robotics},
	publisher = {American Association for the Advancement of Science},
	author = {Greenidge, Nikita J. and Calmé, Benjamin and Moldovan, Alexandru C. and Abaravicius, Bartas and Martin, James W. and Marahrens, Nils and Woolfrey, Jon and Scaglioni, Bruno and Chathuranga, Damith S. and Mitra, Srinjoy and Cochran, Sandy and Valdastri, Pietro},
	month = mar,
	year = {2025},
	pages = {eadq4198},
}

@article{brumfiel_variable_2025,
	title = {A variable stiffness robotically steerable guidewire for endovascular interventions},
	volume = {3},
	copyright = {2025 The Author(s)},
	issn = {2731-4278},
	url = {https://www.nature.com/articles/s44182-025-00029-0},
	doi = {10.1038/s44182-025-00029-0},
	language = {en},
	number = {1},
	urldate = {2025-10-03},
	journal = {npj Robotics},
	publisher = {Nature Publishing Group},
	author = {Brumfiel, Timothy A. and Konda, Revanth and Malhotra, Nidhi and Desai, Jaydev P.},
	month = jul,
	year = {2025},
	pages = {21},
}

@article{prado_robotic-assisted_2024,
	title = {Robotic-{Assisted} {Bronchoscopy}: {A} {Comprehensive} {Review} of {System} {Functions} and {Analysis} of {Outcome} {Data}},
	volume = {14},
	copyright = {http://creativecommons.org/licenses/by/3.0/},
	issn = {2075-4418},
	shorttitle = {Robotic-{Assisted} {Bronchoscopy}},
	url = {https://www.mdpi.com/2075-4418/14/4/399},
	doi = {10.3390/diagnostics14040399},
	language = {en},
	number = {4},
	urldate = {2025-10-03},
	journal = {Diagnostics},
	publisher = {Multidisciplinary Digital Publishing Institute},
	author = {Prado, Renan Martins Gomes and Cicenia, Joseph and Almeida, Francisco Aécio},
	month = jan,
	year = {2024},
	pages = {399},
}

@article{da_veiga_challenges_2020,
	title = {Challenges of continuum robots in clinical context: a review},
	volume = {2},
	issn = {2516-1091},
	shorttitle = {Challenges of continuum robots in clinical context},
	url = {https://doi.org/10.1088/2516-1091/ab9f41},
	doi = {10.1088/2516-1091/ab9f41},
	language = {en},
	number = {3},
	urldate = {2025-10-03},
	journal = {Progress in Biomedical Engineering},
	publisher = {IOP Publishing},
	author = {da Veiga, Tomas and Chandler, James H and Lloyd, Peter and Pittiglio, Giovanni and Wilkinson, Nathan J and Hoshiar, Ali K and Harris, Russell A and Valdastri, Pietro},
	month = jul,
	year = {2020},
	pages = {032003},
}

@article{van_der_schot_impact_2024,
	title = {Impact of cannula diameter on pregnancy outcomes after minimally invasive fetal laser surgery in the treatment of twin-to-twin transfusion syndrome: {A} systematic review and meta-analysis},
	volume = {103},
	copyright = {© 2024 The Authors. Acta Obstetricia et Gynecologica Scandinavica published by John Wiley \& Sons Ltd on behalf of Nordic Federation of Societies of Obstetrics and Gynecology (NFOG).},
	issn = {1600-0412},
	shorttitle = {Impact of cannula diameter on pregnancy outcomes after minimally invasive fetal laser surgery in the treatment of twin-to-twin transfusion syndrome},
	url = {https://onlinelibrary.wiley.com/doi/abs/10.1111/aogs.14761},
	doi = {10.1111/aogs.14761},
	language = {en},
	number = {7},
	urldate = {2025-09-23},
	journal = {Acta Obstetricia et Gynecologica Scandinavica},
	author = {van der Schot, Anouk M. and van Steenis, Josee L. and Sikkel, Esther and Spaanderman, Marc E. A. and van Drongelen, Joris},
	year = {2024},
	note = {\_eprint: https://obgyn.onlinelibrary.wiley.com/doi/pdf/10.1111/aogs.14761},
	pages = {1242--1253},
}

@article{peeters_identification_2015,
	title = {Identification of essential steps in laser procedure for twin–twin transfusion syndrome using the {Delphi} methodology: {SILICONE} study},
	volume = {45},
	copyright = {Copyright © 2014 ISUOG. Published by John Wiley \& Sons Ltd},
	issn = {1469-0705},
	shorttitle = {Identification of essential steps in laser procedure for twin–twin transfusion syndrome using the {Delphi} methodology},
	url = {https://onlinelibrary.wiley.com/doi/abs/10.1002/uog.14761},
	doi = {10.1002/uog.14761},
	language = {en},
	number = {4},
	urldate = {2025-08-29},
	journal = {Ultrasound in Obstetrics \& Gynecology},
	author = {Peeters, S. H. P. and Akkermans, J. and Westra, M. and Lopriore, E. and Middeldorp, J. M. and Klumper, F. J. and Lewi, L. and Devlieger, R. and Deprest, J. and Kontopoulos, E. V. and Quintero, R. and Chmait, R. H. and Smoleniec, J. S. and Otaño, L. and Oepkes, D.},
	year = {2015},
	note = {\_eprint: https://obgyn.onlinelibrary.wiley.com/doi/pdf/10.1002/uog.14761},
	pages = {439--446},
}

@article{akkermans_what_2017,
	title = {What is the impact of placental tissue damage after laser surgery for twin-twin transfusion syndrome? {A} secondary analysis of the {Solomon} trial},
	volume = {52},
	issn = {0143-4004},
	shorttitle = {What is the impact of placental tissue damage after laser surgery for twin-twin transfusion syndrome?},
	url = {https://www.sciencedirect.com/science/article/pii/S0143400417301753},
	doi = {10.1016/j.placenta.2017.02.023},
	urldate = {2025-08-29},
	journal = {Placenta},
	author = {Akkermans, Joost and de Vries, Saskia M. and Zhao, Depeng and Peeters, Suzanne H. P. and Klumper, Frans J. and Middeldorp, Johanna M. and Oepkes, Dick and Slaghekke, Femke and Lopriore, Enrico},
	month = apr,
	year = {2017},
	pages = {71--76},
}

@article{dupont_decade_2021,
	title = {A decade retrospective of medical robotics research from 2010 to 2020},
	volume = {6},
	url = {https://www.science.org/doi/full/10.1126/scirobotics.abi8017},
	doi = {10.1126/scirobotics.abi8017},
	number = {60},
	urldate = {2025-08-29},
	journal = {Science Robotics},
	publisher = {American Association for the Advancement of Science},
	author = {Dupont, Pierre E. and Nelson, Bradley J. and Goldfarb, Michael and Hannaford, Blake and Menciassi, Arianna and O’Malley, Marcia K. and Simaan, Nabil and Valdastri, Pietro and Yang, Guang-Zhong},
	month = nov,
	year = {2021},
	pages = {eabi8017},
}

@article{azizian_visual_2014,
	title = {Visual servoing in medical robotics: a survey. {Part} {I}: endoscopic and direct vision imaging – techniques and applications},
	volume = {10},
	copyright = {Copyright © 2013 John Wiley \& Sons, Ltd.},
	issn = {1478-596X},
	shorttitle = {Visual servoing in medical robotics},
	url = {https://onlinelibrary.wiley.com/doi/abs/10.1002/rcs.1531},
	doi = {10.1002/rcs.1531},
	language = {en},
	number = {3},
	urldate = {2025-08-29},
	journal = {The International Journal of Medical Robotics and Computer Assisted Surgery},
	author = {Azizian, Mahdi and Khoshnam, Mahta and Najmaei, Nima and Patel, Rajni V.},
	year = {2014},
	note = {\_eprint: https://onlinelibrary.wiley.com/doi/pdf/10.1002/rcs.1531},
	pages = {263--274},
}

@article{azizian_visual_2015,
	title = {Visual servoing in medical robotics: a survey. {Part} {II}: tomographic imaging modalities – techniques and applications},
	volume = {11},
	copyright = {Copyright © 2014 John Wiley \& Sons, Ltd.},
	issn = {1478-596X},
	shorttitle = {Visual servoing in medical robotics},
	url = {https://onlinelibrary.wiley.com/doi/abs/10.1002/rcs.1575},
	doi = {10.1002/rcs.1575},
	language = {en},
	number = {1},
	urldate = {2025-08-29},
	journal = {The International Journal of Medical Robotics and Computer Assisted Surgery},
	author = {Azizian, Mahdi and Najmaei, Nima and Khoshnam, Mahta and Patel, Rajni},
	year = {2015},
	note = {\_eprint: https://onlinelibrary.wiley.com/doi/pdf/10.1002/rcs.1575},
	pages = {67--79},
}

@article{wee_systematic_2020,
	title = {A systematic review of the true benefit of robotic surgery: {Ergonomics}},
	volume = {16},
	copyright = {© 2020 John Wiley \& Sons, Ltd},
	issn = {1478-596X},
	shorttitle = {A systematic review of the true benefit of robotic surgery},
	url = {https://onlinelibrary.wiley.com/doi/abs/10.1002/rcs.2113},
	doi = {10.1002/rcs.2113},
	language = {en},
	number = {4},
	urldate = {2025-08-29},
	journal = {The International Journal of Medical Robotics and Computer Assisted Surgery},
	author = {Wee, Ian Jun Yan and Kuo, Li-Jen and Ngu, James Chi-Yong},
	year = {2020},
	note = {\_eprint: https://onlinelibrary.wiley.com/doi/pdf/10.1002/rcs.2113},
	pages = {e2113},
}

@article{schmidgall_will_2025,
	title = {Will your next surgeon be a robot? {Autonomy} and {AI} in robotic surgery},
	volume = {10},
	shorttitle = {Will your next surgeon be a robot?},
	url = {https://www.science.org/doi/10.1126/scirobotics.adt0187},
	doi = {10.1126/scirobotics.adt0187},
	number = {104},
	urldate = {2025-08-29},
	journal = {Science Robotics},
	publisher = {American Association for the Advancement of Science},
	author = {Schmidgall, Samuel and Opfermann, Justin D. and Kim, Ji Woong and Krieger, Axel},
	month = jul,
	year = {2025},
	pages = {eadt0187},
}

@article{rodrigue_soft_2024,
	title = {Soft actuators in surgical robotics: a state-of-the-art review},
	volume = {17},
	issn = {1861-2784},
	shorttitle = {Soft actuators in surgical robotics},
	url = {https://doi.org/10.1007/s11370-023-00506-1},
	doi = {10.1007/s11370-023-00506-1},
	language = {en},
	number = {1},
	urldate = {2025-08-29},
	journal = {Intelligent Service Robotics},
	author = {Rodrigue, Hugo and Kim, Jongwoo},
	month = jan,
	year = {2024},
	pages = {3--17},
}

@article{kabagambe_lessons_2018,
	title = {Lessons from the {Barn} to the {Operating} {Suite}: {A} {Comprehensive} {Review} of {Animal} {Models} for {Fetal} {Surgery}},
	volume = {6},
	issn = {2165-8102, 2165-8110},
	shorttitle = {Lessons from the {Barn} to the {Operating} {Suite}},
	url = {https://www.annualreviews.org/content/journals/10.1146/annurev-animal-030117-014637},
	doi = {10.1146/annurev-animal-030117-014637},
	language = {en},
	number = {Volume 6, 2018},
	urldate = {2025-08-29},
	journal = {Annual Review of Animal Biosciences},
	publisher = {Annual Reviews},
	author = {Kabagambe, Sandra K. and Lee, Chelsey J. and Goodman, Laura F. and Chen, Y. Julia and Vanover, Melissa A. and Farmer, Diana L.},
	month = feb,
	year = {2018},
	pages = {99--119},
}

@article{spruijt_twin-twin_2020,
	title = {Twin-twin transfusion syndrome in the era of fetoscopic laser surgery: antenatal management, neonatal outcome and beyond},
	volume = {13},
	issn = {1747-4086},
	shorttitle = {Twin-twin transfusion syndrome in the era of fetoscopic laser surgery},
	url = {https://doi.org/10.1080/17474086.2020.1720643},
	doi = {10.1080/17474086.2020.1720643},
	number = {3},
	urldate = {2025-08-29},
	journal = {Expert Review of Hematology},
	publisher = {Taylor \& Francis},
	author = {Spruijt, Marjolijn S. and Lopriore, Enrico and J. Steggerda, Sylke and Slaghekke, Femke and Van Klink, Jeanine M.M.},
	month = mar,
	year = {2020},
	note = {\_eprint: https://doi.org/10.1080/17474086.2020.1720643},
	pages = {259--267},
}

@article{peyron_kinematic_2018,
	title = {Kinematic {Analysis} of {Magnetic} {Continuum} {Robots} {Using} {Continuation} {Method} and {Bifurcation} {Analysis}},
	volume = {3},
	issn = {2377-3766},
	url = {https://ieeexplore.ieee.org/document/8410805},
	doi = {10.1109/LRA.2018.2855803},
	number = {4},
	urldate = {2025-08-29},
	journal = {IEEE Robotics and Automation Letters},
	author = {Peyron, Quentin and Boehler, Quentin and Rabenorosoa, Kanty and Nelson, Bradley J. and Renaud, Pierre and Andreff, Nicolas},
	month = oct,
	year = {2018},
	pages = {3646--3653},
}

@inproceedings{tunay_modeling_2004,
	title = {Modeling magnetic catheters in external fields},
	volume = {1},
	url = {https://ieeexplore.ieee.org/abstract/document/1403591},
	doi = {10.1109/IEMBS.2004.1403591},
	urldate = {2025-08-29},
	booktitle = {The 26th {Annual} {International} {Conference} of the {IEEE} {Engineering} in {Medicine} and {Biology} {Society}},
	author = {Tunay, I.},
	month = sep,
	year = {2004},
	pages = {2006--2009},
}

@article{stirnemann_fetal_2018,
	title = {Fetal brain imaging following laser surgery in twin-to-twin surgery},
	volume = {125},
	copyright = {© 2016 Royal College of Obstetricians and Gynaecologists},
	issn = {1471-0528},
	url = {https://onlinelibrary.wiley.com/doi/abs/10.1111/1471-0528.14162},
	doi = {10.1111/1471-0528.14162},
	language = {en},
	number = {9},
	urldate = {2025-08-28},
	journal = {BJOG: An International Journal of Obstetrics \& Gynaecology},
	author = {Stirnemann, J and Chalouhi, G and Essaoui, M and Bahi-Buisson, N and Sonigo, P and Millischer, A-E and Lapillonne, A and Guigue, V and Salomon, Lj and Ville, Y},
	year = {2018},
	note = {\_eprint: https://obgyn.onlinelibrary.wiley.com/doi/pdf/10.1111/1471-0528.14162},
	pages = {1186--1191},
}

@article{lurie_3d_2017,
	title = {{3D} reconstruction of cystoscopy videos for comprehensive bladder records},
	volume = {8},
	issn = {21567085},
	url = {http://dx.doi.org/10.1364/BOE.8.002106},
	doi = {10.1364/BOE.8.002106},
	number = {4},
	urldate = {2023-08-04},
	journal = {Biomedical Optics Express},
	publisher = {Optica Publishing Group},
	author = {Lurie, Kristen L and Angst, Roland and Zlatev, Dimitar V and Liao, Joseph C and Ellerbee Bowden, Audrey K and Ai, D and Yang, J and Fan, J and Zhao, Y and Song, X and Shen, J and Shao, L and Wang, Y and Das, A J and Valdez, T A and Vargas, J A and Saksupapchon, P and Rachapudi, P and Ge, Z and Estrada, J C and Raskar, R and Baek, -e and Lim, J S and Hyung, W J and Riley, G F and Potosky, A L and Lubitz, J D and Kessler, L G},
	month = apr,
	year = {2017},
	pages = {2106--2123},
}

@article{Akkermans2015,
	title = {Twenty-{Five} {Years} of {Fetoscopic} {Laser} {Coagulation} in {Twin}-{Twin} {Transfusion} {Syndrome}: {A} {Systematic} {Review}},
	volume = {38},
	issn = {1015-3837},
	url = {https://www.karger.com/Article/FullText/437053},
	doi = {10.1159/000437053},
	number = {4},
	urldate = {2020-06-04},
	journal = {Fetal Diagnosis and Therapy},
	publisher = {S. Karger AG},
	author = {Akkermans, Joost and Peeters, Suzanne H.P. and Klumper, Frans J. and Lopriore, Enrico and Middeldorp, Johanna M. and Oepkes, Dick},
	month = aug,
	year = {2015},
	pages = {241--253},
}

@article{fagogenis_autonomous_2019,
	title = {Autonomous robotic intracardiac catheter navigation using haptic vision},
	volume = {4},
	url = {https://www.science.org/doi/full/10.1126/scirobotics.aaw1977},
	doi = {10.1126/scirobotics.aaw1977},
	number = {29},
	urldate = {2025-08-27},
	journal = {Science Robotics},
	publisher = {American Association for the Advancement of Science},
	author = {Fagogenis, G. and Mencattelli, M. and Machaidze, Z. and Rosa, B. and Price, K. and Wu, F. and Weixler, V. and Saeed, M. and Mayer, J. E. and Dupont, P. E.},
	month = apr,
	year = {2019},
	pages = {eaaw1977},
}

@article{han_systematic_2022,
	title = {A systematic review of robotic surgery: {From} supervised paradigms to fully autonomous robotic approaches},
	volume = {18},
	copyright = {© 2021 John Wiley \& Sons Ltd.},
	issn = {1478-596X},
	shorttitle = {A systematic review of robotic surgery},
	url = {https://onlinelibrary.wiley.com/doi/abs/10.1002/rcs.2358},
	doi = {10.1002/rcs.2358},
	language = {en},
	number = {2},
	urldate = {2025-08-27},
	journal = {The International Journal of Medical Robotics and Computer Assisted Surgery},
	author = {Han, Jinpei and Davids, Joseph and Ashrafian, Hutan and Darzi, Ara and Elson, Daniel S. and Sodergren, Mikael},
	year = {2022},
	note = {\_eprint: https://onlinelibrary.wiley.com/doi/pdf/10.1002/rcs.2358},
	pages = {e2358},
}

@article{lee_levels_2024,
	title = {Levels of autonomy in {FDA}-cleared surgical robots: a systematic review},
	volume = {7},
	copyright = {2024 The Author(s)},
	issn = {2398-6352},
	shorttitle = {Levels of autonomy in {FDA}-cleared surgical robots},
	url = {https://www.nature.com/articles/s41746-024-01102-y},
	doi = {10.1038/s41746-024-01102-y},
	language = {en},
	number = {1},
	urldate = {2025-08-27},
	journal = {npj Digital Medicine},
	publisher = {Nature Publishing Group},
	author = {Lee, Audrey and Baker, Turner S. and Bederson, Joshua B. and Rapoport, Benjamin I.},
	month = apr,
	year = {2024},
	pages = {103},
}

@article{dupont_grand_2025,
	title = {The grand challenges of learning medical robot autonomy},
	volume = {10},
	url = {https://www.science.org/doi/full/10.1126/scirobotics.adz8279},
	doi = {10.1126/scirobotics.adz8279},
	number = {104},
	urldate = {2025-08-27},
	journal = {Science Robotics},
	publisher = {American Association for the Advancement of Science},
	author = {Dupont, Pierre E. and Degirmenci, Alperen},
	month = jul,
	year = {2025},
	pages = {eadz8279},
}

@article{amberg_why_2021,
	title = {Why {Do} the {Fetal} {Membranes} {Rupture} {Early} after {Fetoscopy}? {A} {Review}},
	volume = {48},
	issn = {1015-3837},
	shorttitle = {Why {Do} the {Fetal} {Membranes} {Rupture} {Early} after {Fetoscopy}?},
	url = {https://doi.org/10.1159/000517151},
	doi = {10.1159/000517151},
	number = {7},
	urldate = {2025-08-26},
	journal = {Fetal Diagnosis and Therapy},
	author = {Amberg, Benjamin J. and Hodges, Ryan J. and Rodgers, Karyn A. and Crossley, Kelly J. and Hooper, Stuart B. and DeKoninck, Philip L.J.},
	month = aug,
	year = {2021},
	pages = {493--503},
}

@article{lussier_sheep_2025,
	title = {Sheep {Models} in {Translational} {Surgery}},
	volume = {66},
	issn = {0014-312X},
	url = {https://doi.org/10.1159/000546157},
	doi = {10.1159/000546157},
	number = {1},
	urldate = {2025-08-24},
	journal = {European Surgical Research},
	author = {Lussier, Bertrand and Behr, Luc and Borenstein, Nicolas and Brants, Irena and Garabedian, Charles and Ghesquiere, Louise and Le Duc, Kevin and Sharma, Dyuti and Storme, Laurent and Touzot-Jourde, Gwenola and White, Jeff and Hubert, Thomas},
	month = apr,
	year = {2025},
	pages = {39--45},
}

@article{bamberg_twin--twin_2022,
	series = {Controversies in {Twin}/{Multiple} {Pregnancy}},
	title = {Twin-to-twin transfusion syndrome: {Controversies} in the diagnosis and management},
	volume = {84},
	issn = {1521-6934},
	shorttitle = {Twin-to-twin transfusion syndrome},
	url = {https://www.sciencedirect.com/science/article/pii/S1521693422000529},
	doi = {10.1016/j.bpobgyn.2022.03.013},
	urldate = {2025-07-23},
	journal = {Best Practice \& Research Clinical Obstetrics \& Gynaecology},
	author = {Bamberg, Christian and Hecher, Kurt},
	month = nov,
	year = {2022},
	pages = {143--154},
}

@article{Akkermans2017,
	title = {Impact of {Laser} {Power} and {Firing} {Angle} on {Coagulation} {Efficiency} in {Laser} {Treatment} for {Twin}-{Twin} {Transfusion} {Syndrome}: {An} ex vivo {Placenta} {Study}},
	volume = {42},
	issn = {1015-3837},
	url = {https://www.karger.com/Article/FullText/464323},
	doi = {10.1159/000464323},
	number = {3},
	urldate = {2020-06-04},
	journal = {Fetal Diagnosis and Therapy},
	publisher = {S. Karger AG},
	author = {Akkermans, Joost and van der Donk, Loes and Peeters, Suzanne H.P. and van Tuijl, Sjoerd and Middeldorp, Johanna M. and Lopriore, Enrico and Oepkes, Dick},
	month = oct,
	year = {2017},
	pages = {204--209},
}

@article{wang_apriltag_2016,
	title = {{AprilTag} 2: {Efficient} and robust fiducial detection},
	issn = {21530866},
	url = {https://doi.org/10.1109/IROS.2016.7759617},
	doi = {10.1109/IROS.2016.7759617},
	urldate = {2023-08-23},
	journal = {2016 IEEE/RSJ International Conference on Intelligent Robots and Systems (IROS)},
	publisher = {Institute of Electrical and Electronics Engineers Inc.},
	author = {Wang, John and Olson, Edwin},
	month = nov,
	year = {2016},
	note = {ISBN: 9781509037629},
	pages = {4193--4198},
}

@article{olson_apriltag_2011,
	title = {{AprilTag}: {A} robust and flexible visual fiducial system},
	issn = {10504729},
	url = {http://dx.doi.org/10.1109/ICRA.2011.5979561},
	doi = {10.1109/ICRA.2011.5979561},
	urldate = {2023-08-23},
	journal = {2011 IEEE International Conference on Robotics and Automation},
	author = {Olson, Edwin},
	year = {2011},
	note = {ISBN: 9781612843865},
	pages = {3400--3407},
}

@article{dreyfus_dexterous_2024,
	title = {Dexterous helical magnetic robot for improved endovascular access},
	volume = {9},
	url = {https://www.science.org/doi/10.1126/scirobotics.adh0298},
	doi = {10.1126/scirobotics.adh0298},
	number = {87},
	urldate = {2024-09-24},
	journal = {Science Robotics},
	publisher = {American Association for the Advancement of Science},
	author = {Dreyfus, R. and Boehler, Q. and Lyttle, S. and Gruber, P. and Lussi, J. and Chautems, C. and Gervasoni, S. and Berberat, J. and Seibold, D. and Ochsenbein-Kölble, N. and Reinehr, M. and Weisskopf, M. and Remonda, L. and Nelson, B. J.},
	month = feb,
	year = {2024},
	pages = {eadh0298},
}

@article{Pandya2020a,
	title = {Current {Practice} and {Protocols}: {Endoscopic} {Laser} {Therapy} for {Twin}-{Twin} {Transfusion} {Syndrome}},
	volume = {2},
	issn = {26415895},
	url = {http://dx.doi.org/10.1097/FM9.0000000000000035},
	doi = {10.1097/fm9.0000000000000035},
	number = {1},
	urldate = {2020-06-04},
	journal = {Maternal-Fetal Medicine},
	publisher = {Ovid Technologies (Wolters Kluwer Health)},
	author = {Pandya, Viral M. and Stirnemann, Julien and Colmant, Claire and Ville, Yves},
	month = jan,
	year = {2020},
	pages = {34--47},
}

@article{mattille_autonomous_2024,
	title = {Autonomous {Magnetic} {Navigation} in {Endoscopic} {Image} {Mosaics}},
	volume = {11},
	issn = {2198-3844},
	url = {https://onlinelibrary.wiley.com/doi/10.1002/advs.202400980},
	doi = {10.1002/advs.202400980},
	number = {19},
	urldate = {2024-09-20},
	journal = {Advanced Science},
	publisher = {John Wiley \& Sons, Ltd},
	author = {Mattille, Michelle and Boehler, Quentin and Lussi, Jonas and Ochsenbein, Nicole and Moehrlen, Ueli and Nelson, Bradley J.},
	month = may,
	year = {2024},
	pages = {2400980},
}

@article{li_robust_2023,
	title = {Robust endoscopic image mosaicking via fusion of multimodal estimation},
	volume = {84},
	issn = {13618415},
	url = {http://dx.doi.org/10.1016/j.media.2022.102709},
	doi = {10.1016/J.MEDIA.2022.102709},
	urldate = {2023-08-19},
	journal = {Medical Image Analysis},
	publisher = {Elsevier},
	author = {Li, Liang and Mazomenos, Evangelos and Chandler, James H. and Obstein, Keith L. and Valdastri, Pietro and Stoyanov, Danail and Vasconcelos, Francisco},
	month = feb,
	year = {2023},
	pages = {102709},
}

@article{petruska_model-based_2017,
	title = {Model-{Based} {Calibration} for {Magnetic} {Manipulation}},
	volume = {53},
	issn = {00189464},
	url = {http://dx.doi.org/10.1109/TMAG.2017.2653080},
	doi = {10.1109/TMAG.2017.2653080},
	number = {7},
	urldate = {2023-08-04},
	journal = {IEEE Transactions on Magnetics},
	publisher = {Institute of Electrical and Electronics Engineers Inc.},
	author = {Petruska, Andrew J. and Edelmann, Janis and Nelson, Bradley J.},
	month = jul,
	year = {2017},
	pages = {1--6},
}

@inproceedings{wu_model-free_2015,
	title = {Model-free image guidance for intelligent tubular robots with pre-clinical feasibility study: {Towards} minimally invasive trans-orifice surgery},
	shorttitle = {Model-free image guidance for intelligent tubular robots with pre-clinical feasibility study},
	url = {https://ieeexplore.ieee.org/document/7279384},
	doi = {10.1109/ICInfA.2015.7279384},
	urldate = {2024-09-24},
	booktitle = {2015 {IEEE} {International} {Conference} on {Information} and {Automation}},
	author = {Wu, Keyu and Wu, Liao and Lim, Chwee Ming and Ren, Hongliang},
	month = aug,
	year = {2015},
	pages = {749--754},
}

@article{cruz-martinez_flexible_2023,
	title = {Flexible {Video} {Fetoscopy}: {Feasibility} and {Outcomes} of a {Novel} {Modality} for {Laser} {Therapy} in {Twin}-to-{Twin} {Transfusion} {Syndrome} {Presenting} with {Inaccessible} {Anterior} {Placenta}},
	volume = {50},
	copyright = {https://karger.com/pages/terms-and-conditions},
	issn = {1015-3837, 1421-9964},
	shorttitle = {Flexible {Video} {Fetoscopy}},
	url = {https://karger.com/FDT/article/doi/10.1159/000528815},
	doi = {10.1159/000528815},
	language = {en},
	number = {2},
	urldate = {2024-09-25},
	journal = {Fetal Diagnosis and Therapy},
	author = {Cruz-Martínez, Rogelio and Gil Pugliese, Savino and Villalobos-Gómez, Rosa and Martinez-Rodriguez, Miguel and Gámez-Varela, Alma and López-Briones, Hugo and Chávez-González, Eréndira and Diaz-Primera, Ramiro},
	year = {2023},
	pages = {106--114},
}

@article{lussi_magnetically_2022,
	title = {Magnetically {Guided} {Laser} {Surgery} for the {Treatment} of {Twin}-to-{Twin} {Transfusion} {Syndrome}},
	volume = {4},
	copyright = {© 2022 The Authors. Advanced Intelligent Systems published by Wiley-VCH GmbH},
	issn = {2640-4567},
	url = {https://onlinelibrary.wiley.com/doi/abs/10.1002/aisy.202200182},
	doi = {10.1002/aisy.202200182},
	language = {en},
	number = {11},
	urldate = {2024-09-20},
	journal = {Advanced Intelligent Systems},
	author = {Lussi, Jonas and Gervasoni, Simone and Mattille, Michelle and Dreyfus, Roland and Boehler, Quentin and Reinehr, Michael and Ochsenbein, Nicole and Nelson, Bradley J and Moehrlen, Ueli},
	year = {2022},
	note = {\_eprint: https://onlinelibrary.wiley.com/doi/pdf/10.1002/aisy.202200182},
	pages = {2200182},
}

@article{ahmad_development_2023,
	title = {Development and validation of a flexible fetoscope for fetoscopic laser coagulation},
	volume = {18},
	issn = {1861-6429},
	url = {https://doi.org/10.1007/s11548-023-02905-2},
	doi = {10.1007/s11548-023-02905-2},
	language = {en},
	number = {9},
	urldate = {2024-09-20},
	journal = {International Journal of Computer Assisted Radiology and Surgery},
	author = {Ahmad, Mirza Awais and Ourak, Mouloud and Wenmakers, Dirk and Valenzuela, Ignacio and Basurto, David and Ourselin, Sebastien and Vercauteren, Tom and Deprest, Jan and Poorten, Emmanuel Vander},
	month = sep,
	year = {2023},
	pages = {1603--1611},
}

@article{gervasoni_human-scale_2024,
	title = {A {Human}-{Scale} {Clinically} {Ready} {Electromagnetic} {Navigation} {System} for {Magnetically} {Responsive} {Biomaterials} and {Medical} {Devices}},
	volume = {36},
	copyright = {© 2024 The Author(s). Advanced Materials published by Wiley-VCH GmbH},
	issn = {1521-4095},
	url = {https://onlinelibrary.wiley.com/doi/abs/10.1002/adma.202310701},
	doi = {10.1002/adma.202310701},
	language = {en},
	number = {31},
	urldate = {2024-09-23},
	journal = {Advanced Materials},
	author = {Gervasoni, Simone and Pedrini, Norman and Rifai, Tarik and Fischer, Cedric and Landers, Fabian C. and Mattmann, Michael and Dreyfus, Roland and Viviani, Silvia and Veciana, Andrea and Masina, Enea and Aktas, Buse and Puigmartí-Luis, Josep and Chautems, Christophe and Pané, Salvador and Boehler, Quentin and Gruber, Philipp and Nelson, Bradley J.},
	year = {2024},
	note = {\_eprint: https://onlinelibrary.wiley.com/doi/pdf/10.1002/adma.202310701},
	pages = {2310701},
}
\bibliographystyle{sciencemag}

%%%%%%%%%%%%%%%% ACKNOWLEDGEMENTS %%%%%%%%%%%%%%%

\section*{Acknowledgments}
\begin{figure}[p]
	\centering
	\includegraphics[width=1\textwidth]{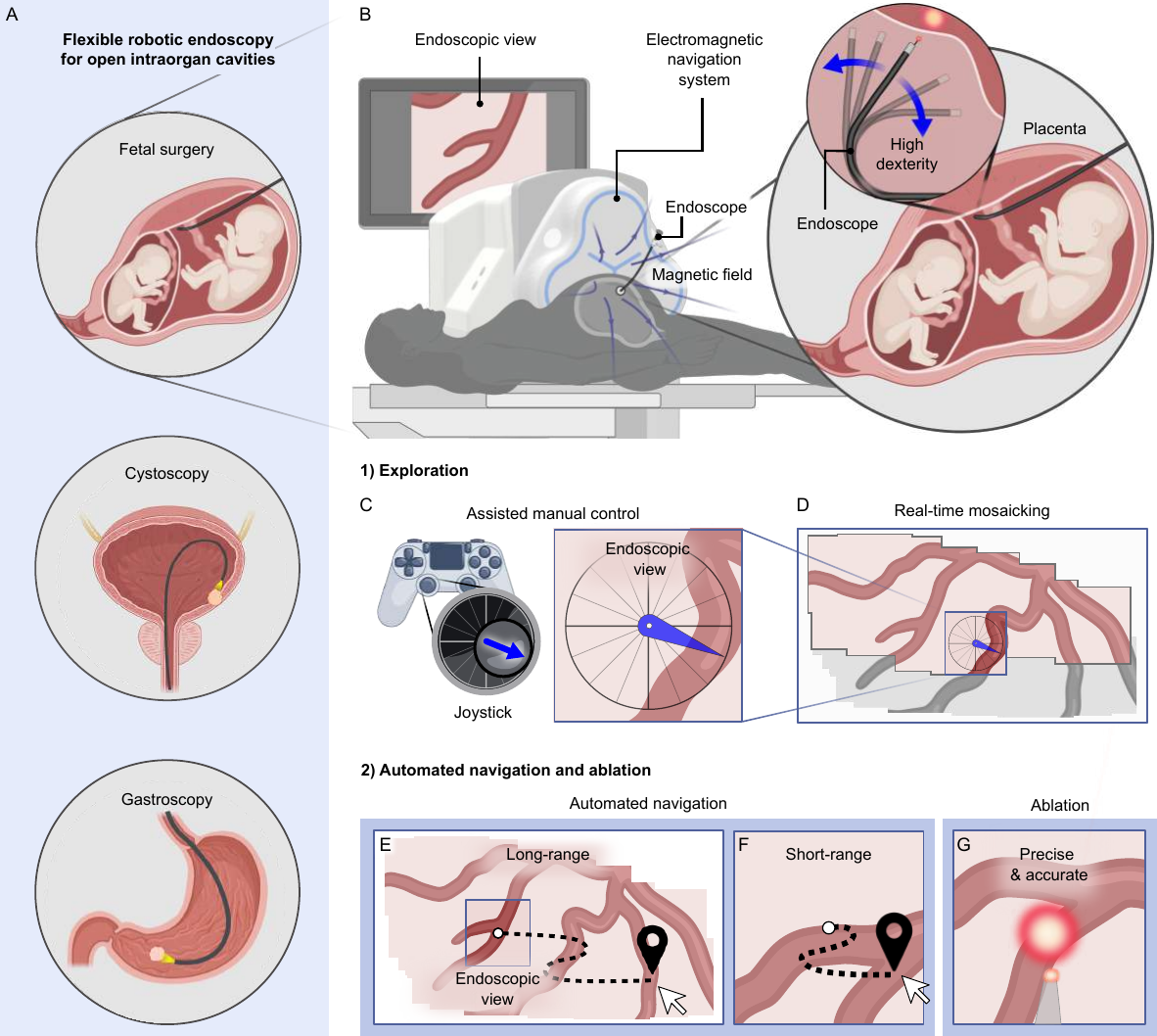}
	\caption{\textbf{Overview of our robotic flexible endoscopy platform.}
    (\textbf{A}) Relevant endoscopic surgeries in open intraorgan cavities. (\textbf{B}) Platform configuration for the treatment of twin to twin transfusion syndrome: the electromagnetic navigation system is laterally positioned with respect to the patient, and generates a magnetic field to steer the tip of a flexible robotic fetoscope. (\textbf{C}) During the exploration phase of the procedure the fetoscope can be controlled manually using a joystick (\textbf{D}) while generating image mosaics in real time. (\textbf{E}) It can navigate automatically to targets on the image mosaic or (\textbf{F}) on the endoscopic image for (\textbf{G}) high precision vessel ablations.}
	\label{fig:overview}
\end{figure}

\begin{figure}[p]
	\centering
	\includegraphics[width=1\textwidth]{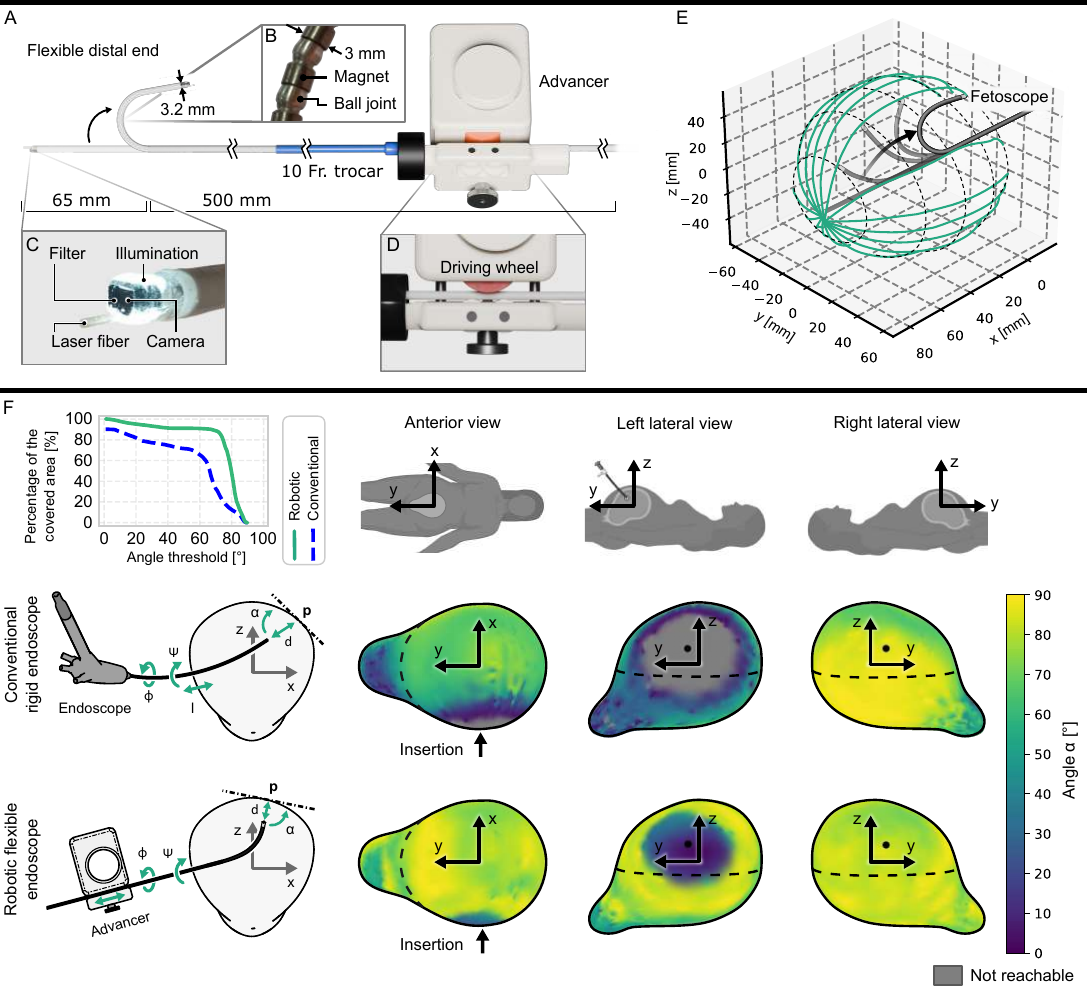} 
	\caption{\textbf{Design and characterization of the robotic fetoscope.}
    (\textbf{A}) Robotic fetoscope with a flexible tip, which fits through a 10~Fr. trocar and is inserted over an advancer. (\textbf{B}) Flexible distal section consisting of an array of magnets and ball joints. The tip (\textbf{C}), and the advancer (\textbf{D}) containing an active driving wheel, and two passive wheels. (\textbf{E}) Stable positions of the fetoscope's tip for a sweep in magnetic field orientations. (\textbf{F}) Best angles~$\alpha$ for a conventional rigid fetoscope (11508AAK, Karl Storz) and the robotic fetoscope on a uterine model for a lateral insertion point during fetoscopic laser coagulation. The plot depicts the percentage of the anterior surface area (above the dotted line in the uterus model on the bottom right) that is equal or above the angle thresholds for both devices.}
	\label{fig:workspace}
\end{figure}

We thank Lilian and Joni Rom for their contribution to the ex vivo ablation experiment. We also thank Sean Lyttle for designing and building the advancer unit, Joaquim Llacer for his help in the fabrication of the fetoscope, and Elizabeth Zuurmond for proofreading the manuscript. Figure~\ref{fig:overview} and Figure~\ref{fig:workspace} contain elements created in BioRender. Mattille, M. (2026) https://BioRender.com/3oi66am and https://BioRender.com/3axc39q.
\paragraph*{Funding:}
This work was supported by the Swiss National Science Foundation grant 200020\_212885 (M.M. and Q.B.) ; and ITC-InnoHK grant 16312 (A.M., M.M., and Q.B.).
\paragraph*{Author contributions:}
Conceptualization: M.M., and Q.B. Methodology: M.M., A.M., and Q.B. Software: M.M. Hardware: A.M. and M.M. Validation: M.M., A.M., Q.B., M.W., N.O., and U.M. Formal analysis: M.M. Investigation: M.M., M.W., N.O., and U.M. Resources: M.W., N.O., U.M., and B.J.N. Data Curation: M.M. Writing—original draft: M.M., and Q.B. Writing—review and editing: M.M., Q.B., B.J.N., A.M., U.M, N.O. and M.W. Visualization: M.M, and A.M. Supervision: B.J.N., and Q.B. Project administration: B.J.N. Funding acquisition: B.J.N. 
\paragraph*{Competing interests:}
B.J.N. is a co-founder of Nanoflex Robotics AG and MagnebotiX AG. B.J.N. is on the editorial board of Science Robotics. A.M and B.J.N. are authors on patent applications CN119998001A and EP4598616A1 “Steerable device for use inside of a mammalian body” describing the magnetic articulating tip technology used in the robotic fetoscope design. The other authors declare that they have no competing interests.
\paragraph*{Data, code and materials availability:}
All (other) data needed to evaluate the conclusions in the paper are present in the paper or the Supplementary Materials. All reagents and materials associated with this study are listed in the Methods section and/or are available commercially. The data for this study have been deposited in the database DOI: 10.5061/dryad.zpc866tpb.

\clearpage % flush any earlier floats

%%%%%%%%%%%%%%%% MAIN TEXT FIGURES %%%%%%%%%%%%%%%

%Fig 1

%Fig 2

%Fig 3
\begin{figure}[p]
	\centering
	\includegraphics[width=1\textwidth]{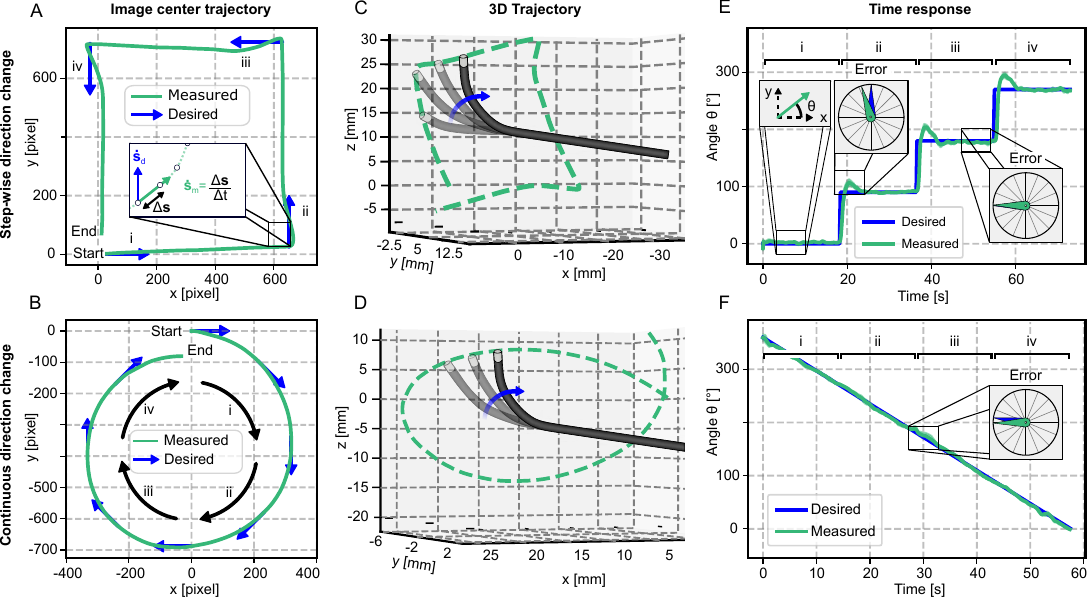} 
	\caption{\textbf{In vitro evaluation of the image-based control method.}
    Median signal of 5 repetitions for a stepwise input direction change (first row), and for a continuous direction change (second row). (\textbf{A},\textbf{B}) Integrated velocity of the image center. (\textbf{C},\textbf{D}) Robotic fetoscope's tip trajectories in 3D space. (\textbf{E},\textbf{F}) Desired and measured angle over time.}
	\label{fig:pov_eval}
\end{figure}

%Fig 4

%Fig 5
\begin{figure}[p]
	\centering
	\includegraphics[width=1\textwidth]{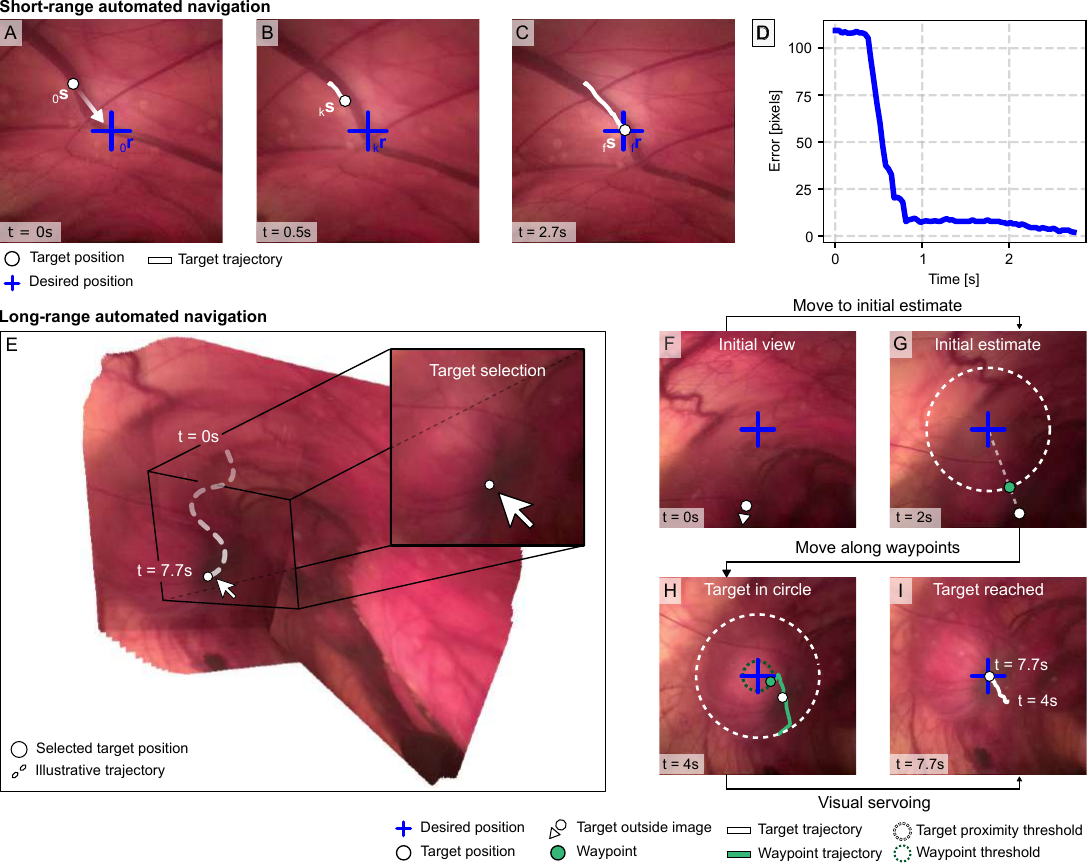} 
	\caption{\textbf{In vivo automated navigation.}
     (\textbf{A-C}) Endoscopic images of the short-range automated navigation to a selected target in the endoscopic image in vivo in an ovine model with (\textbf{D}) the error over time. (\textbf{E}) Endoscopic image mosaic used as user interface for the long-range automated navigation in vivo with an illustration of the trajectory to the selected target location. (\textbf{F}) The target is not visible on the initial endoscopic view. (\textbf{G}) The robotic fetoscope then moves to the initial estimate based on the magnetic field, where the target is at the border of the image and a waypoint is used for navigating closer to the target until (\textbf{H}) the waypoint is reached within the waypoint threshold and the selected target lies within the target proximity threshold. (\textbf{I}) Short-range automated navigation to reach the selected target.
        }
	\label{fig:in_vivo_nav_automated}
\end{figure}

%Fig 6

%Fig 7
\begin{figure}[p]
	\centering
	\includegraphics[width=1\textwidth]{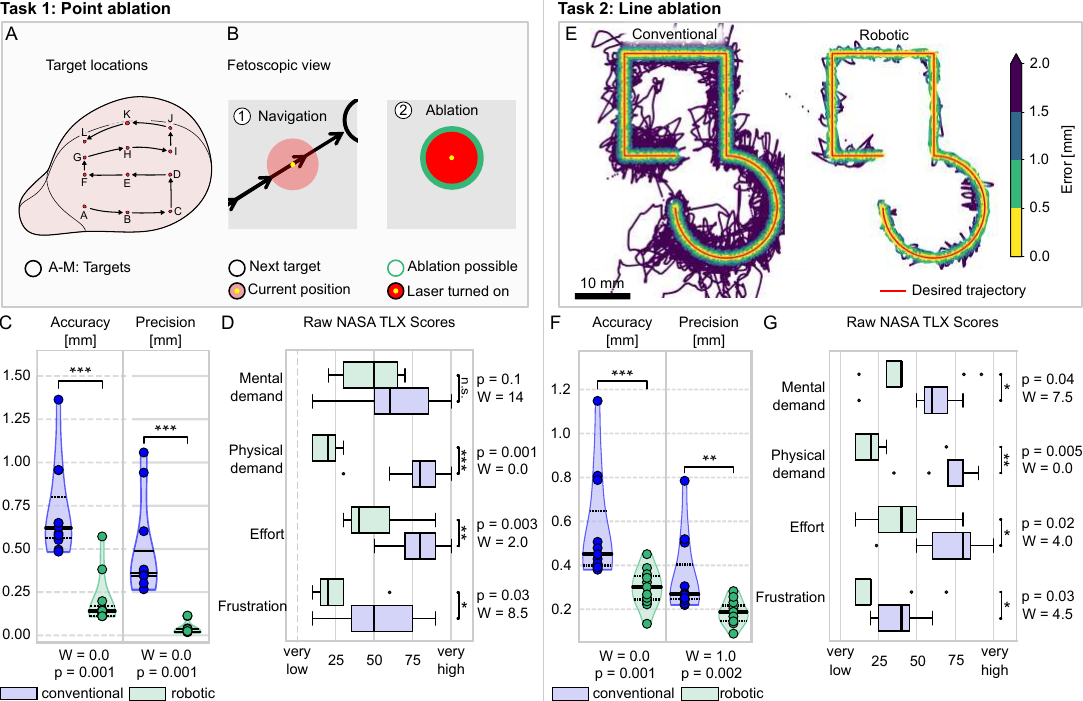} 
	\caption{\textbf{Results of the usability study.}
    (\textbf{A}) Illustration of the point ablation task (Task 1) with the target locations on a uterus model and (\textbf{B}) user interface during navigation and ablation. (\textbf{C}) Median point ablation accuracy and precision over all targets, and (\textbf{D}) raw NASA TLX scores of 11 participants (N = 11). (\textbf{E-G}) Results of the line ablation task (Task 2), including (\textbf{E}) all trajectories of all participants for both fetoscopes, (\textbf{F}) accuracy and precision of the ablation and (\textbf{G}) raw NASA TLX scores. In the violin plots (\textbf{C} and \textbf{F}), the center lines denote the median and the dotted lines indicate the IQR. In the box plots (\textbf{D} and \textbf{G}), the boxes illustrate the IQR, the center lines indicate the median, and the whiskers show the range within 1.5$\times$IQR. Statistical significance was assessed over a paired Wilcoxon signed-rank test: *$p<0.05$, **$p<0.01$ and ***$p<0.001$.} 
	\label{fig:user_study}
\end{figure}

%Fig 8
\begin{figure}[p]
	\centering
	\includegraphics[width=1\textwidth]{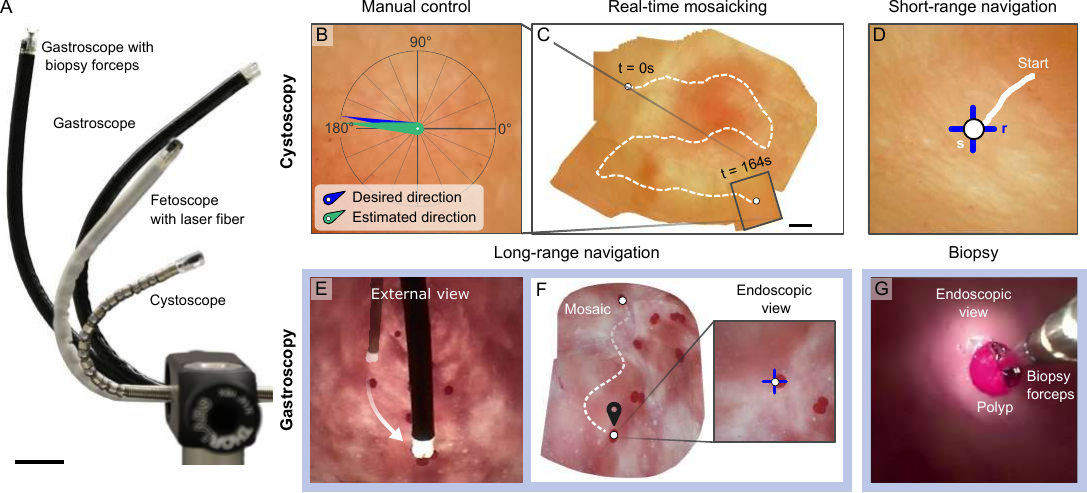} 
	\caption{\textbf{Other applications.}
    (\textbf{A}) Robotic endoscopes for different applications submitted to a constant magnetic field of~25 mT pointing toward the right. %, at 200 mm from the electromagnetic navigation system. 
    The scale bar corresponds to 10 mm. 
    Illustration of the navigation methods ex vivo for (\textbf{B}-\textbf{D}) cystoscopy and (\textbf{E}-\textbf{G}) gastroscopy. (\textbf{B}) Endoscopic image of the manual navigation in an ex vivo bladder. (\textbf{C}) Mosaic of the ex vivo bladder with a scale bar corresponding to 100~pixels. (\textbf{D}) Short-range navigation with the cystoscope. (\textbf{E}-\textbf{F})} Long-range navigation of the gastroscope, shown in (\textbf{E}) an external view and (\textbf{F}) the mosaic with the endoscopic view when the target was reached. (\textbf{G}) Endoscopic view while performing a biopsy.
	\label{fig:other_app}
\end{figure}

%TC:ignore

\clearpage % flush the 8 main-figure floats before starting the supplement

%%%%%%%%%%%%%%%% END OF MAIN TEXT %%%%%%%%%%%%%%%

\newpage

%%%%%%%%%%%%%%%% START OF SUPPLEMENT %%%%%%%%%%%%%%%

% Figures, tables, equations and pages in the supplement are numbered S1, S2 etc.
\renewcommand{\thefigure}{S\arabic{figure}}
\renewcommand{\thetable}{S\arabic{table}}
\renewcommand{\theequation}{S\arabic{equation}}
\renewcommand{\thepage}{S\arabic{page}}
\setcounter{figure}{0}
\setcounter{table}{0}
\setcounter{equation}{0}
\setcounter{page}{1} % not 0 as \newpage already started a supplementary page
% References continue the numbering from the main text.

\subsection*{Supplementary Methods}

\subsubsection*{Robotic fetoscope design}

The robotic fetoscope, shown in Figure~\ref{sfig:fetoscope}, addresses the limitations of our previous design reported in \cite{lussi_magnetically_2022}. It enables monitoring vessel ablations in the endoscopic view, improves stability in the magnetic field, and achieves larger bending angles in the magnetic field. Replacing LED's with optical fibers in combination with a transparent tip enabled a more homogeneous and stronger illumination. The fetoscope exhibits the functionalities of regular fetoscopes and was tested with two commercial laser fibers, which have been evaluated for FLC ex vivo (OD of 500~$\mu$m, KLS Martin) and in human (OD of 300~$\mu$m, Giga Laser, China)~\cite{lussi_magnetically_2022,cruz-martinez_flexible_2023}.
Our robotic fetoscope contains a small CMOS camera (OVM6946, OmniVision), optical fibers (optical grade plastic, OD of 250 $\mu$m, Edmund Optics) for illumination and a channel for a laser fiber (polyimide, OD of 0.69 mm, ID of 0.64 mm, ZEUS Inc., USA). A filter is placed over the camera to avoid saturation during lasering due to reflections on the tissue (2.1~x~1.6~mm size, Colored Glass Heat Absorbing Shortpass Filter KG5, SCHOTT, Germany). The tip of the fetoscope is 3D-printed and transparent (PX 521HT Polyurethane).

The distal section comprises a series of stacked magnets and precision-engineered steel ball joints (see Figure~\ref{fig:workspace}B), designed using the open-source software toolbox introduced in~\cite{mesot_parametric_2025}, to allow large bending angles and precise magnetic navigation. An array of 9 NdFeB magnets is placed behind the tip (OD of 3 mm, segments with ID of 1.9 and 1.7 lengths of 2 and 4 mm, X-Magnet, China) followed by a flexible section consisting of an array of soft-magnetic 440 stainless steel ball joints with magnets in between (OD of 3 mm, ID of 1.7 mm, full joint length of 4.2 mm). Magnetic ball joints provide flexibility and stability in magnetic fields of 10-23 mT~\cite{mesot_parametric_2025}. For stability, the proximal ball joint is welded to a nonmagnetic 316 stainless steel hypodermic tube (OD of 2.8 mm, ID of 1.7 mm, McMaster Carr, USA). A flexible endoscopic insertion tube section consisting of a spring with a USP class VI biocompatible polyurethane jacket (EDC, USA) connects the stainless steel tube to the casing of the camera board, acting as a stress relief connector at the mechanical junction. The casing is waterproof and contains connectors for illumination and the camera and contains an insertion point for the channel of the laser fiber. It is mounted on a FISSO surgical arm (see Figure~\ref{sfig:in_vivo_setup}A). The outer layer between the fetoscope tip and the casing consists of a soft Pebax jacket (Durometer = 35 D, Nordson Medical, USA).

\subsubsection*{Calibration process for the image Jacobian matrix}

\begin{figure}[ht]
	\centering
	\includegraphics[width=1\textwidth]{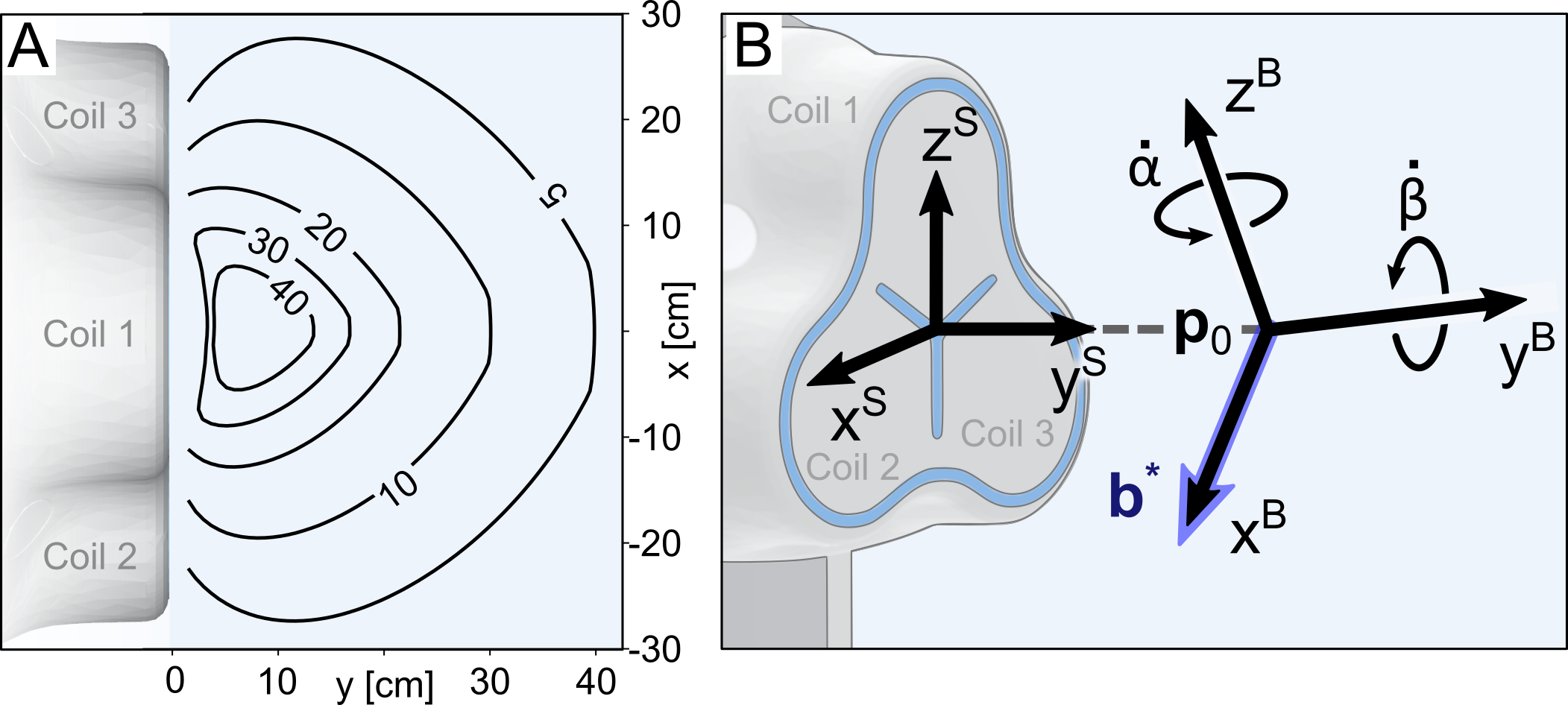} 
	\caption{\textbf{Workspace and reference frames of the electromagnetic navigation system (eMNS).}
    (\textbf{A}) Isolines of achievable field magnitude in space in mT for maximal currents of 45~A in the electromagnets. (\textbf{B}) eMNS with the reference frame "S" attached to the system and the rotating magnetic field "B", which is attached to the desired magnetic field $\mathbf{b}^*$ generated by the system such that ${}_B \mathbf{b}^* = \begin{bmatrix}
        b & 0& 0
    \end{bmatrix}$.
    }
	\label{sfig:navion}
\end{figure}

To calibrate the image Jacobian matrix~$\mathbf{J}$ upon starting navigation, two orthogonal calibration motions $\dot{\mathbf{q}}_1$, $\dot{\mathbf{q}}_2$ corresponding to magnetic field rotations are performed while the velocity of the image center is estimated, resulting in $\dot {\mathbf{s}}_1$ and $\dot{\mathbf{s}}_2$. The image Jacobian matrix~$\mathbf{J}$ is subsequently estimated from the mean values of the observed metrics.

\begin{equation}
    \hat{\mathbf{J}} = \begin{bmatrix} \bar{\dot{ \mathbf{s}}}_1 & \bar{\dot{ \mathbf{s}}}_2 \end{bmatrix} \begin{bmatrix} \bar{\dot{ \mathbf{q}}}_1 & \bar{\dot{ \mathbf{q}}}_2 \end{bmatrix}^{-1}
\label{eq:jacobian_calibration}
\end{equation}

\noindent where $\hat{\mathbf{J}}$ is the estimated image Jacobian, $\bar{\dot{ \mathbf{s}}}_1$ and $\bar{\dot{ \mathbf{s}}}_2$ are the mean image center velocities, and $\bar{\dot{ \mathbf{q}}}_1$ and $\bar{\dot{ \mathbf{q}}}_2$ are the mean intrinsic rotation angles of the B-frame (see Figure~\ref{sfig:navion}B).

To obtain a more robust velocity estimate than in our previous works~\cite{lussi_magnetically_2022, mattille_autonomous_2024}, we estimate the image velocity using the method described in Section~``Manual control".

\subsubsection*{Autonomous exploration motions}

\begin{figure}[ht]
	\centering
	\includegraphics[width=1\textwidth]{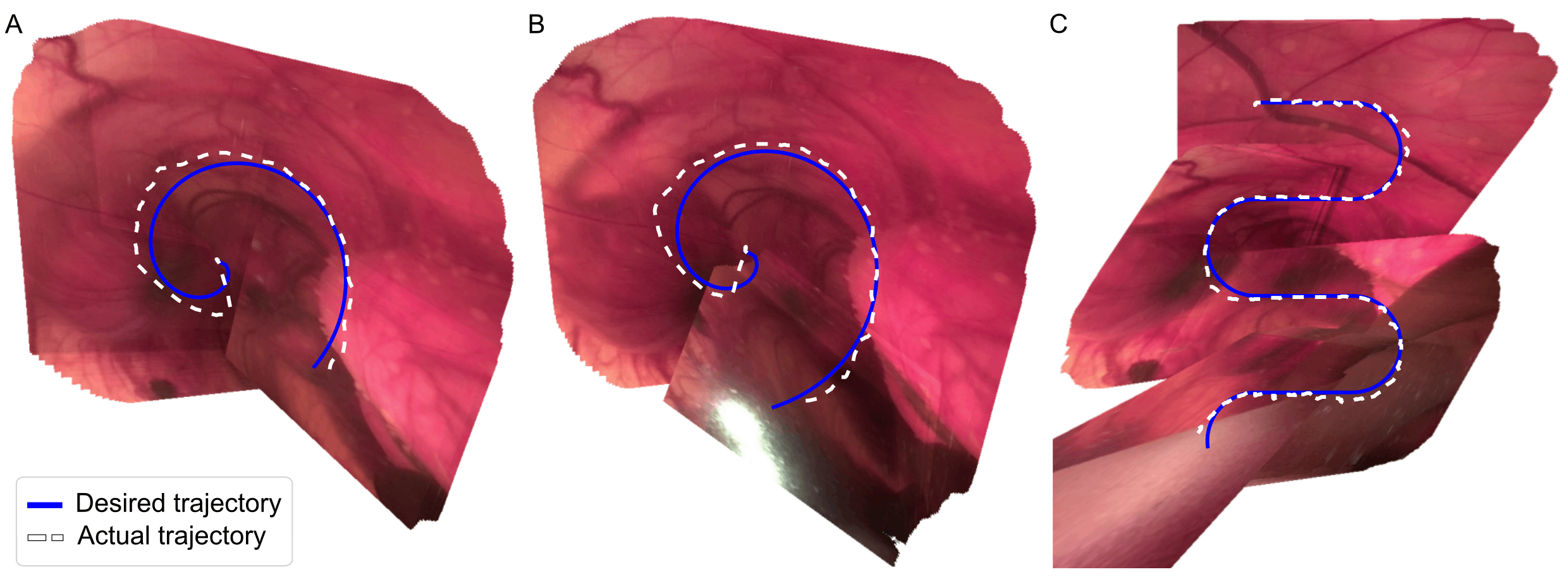} 
	\caption{\textbf{In vivo automated exploration.}
    Endoscopic image mosaics generated during the in vivo automated exploration motions. The desired trajectory and the actual trajectories of the image center points are shown for (\textbf{A-B}) spiral and (\textbf{C}) rounded raster exploration patterns.
        }
	\label{sfig:in_vivo_automated_exploration}
\end{figure}

The endoscopic image mosaics generated during the in vivo experiment were obtained while the robotic fetoscope autonomously explored its environment following a spiral or grid trajectory in the mosaic frame (see Figure~\ref{sfig:in_vivo_automated_exploration}). The method proposed here improves on the one introduced in~\cite{mattille_autonomous_2024}, which was only applicable to explore relatively small regions around the posterior orientation of the robotic fetoscope. The present work addresses this limitation by updating the image Jacobian matrix~$\mathbf{J}$ for a more accurate mapping over larger areas, as well as increased robustness in various endoscopic orientations (see Figure~\ref{sfig:detailed_nav_eval}).

\subsubsection*{Long-range automated control}

First, the magnetic field~$\hat{\mathbf{b}}$ at the target location is estimated by interpolating previous field orientations, assuming all points of the trajectory of the endoscopic image centers in the mosaic do not lie on the same line. After the robotic fetoscope navigated to the magnetic field~$\hat{\mathbf{b}}$, the target is projected from the image~$I_t$, which is the endoscopic image from which the pixel of the target on the mosaic originates, to the frame “k” of the current endoscopic image~$I_k$. If the norm of the error~$\mathbf{e}$ between the reference position and the target position is larger than a threshold $t_w=125$~pixels, a waypoint~$\mathbf{p}_w$ is introduced. It is defined as the projection of the target on a circle of radius~$t_w$ centered in
$\mathbf{r}$ (see Figure~\ref{fig:in_vivo_nav_automated}G)

\begin{equation}
    \mathbf{p}_w = \mathbf{r} + t_w \frac{\mathbf{e}}{\left\|\mathbf{e} \right\|}
\end{equation}

The robotic fetoscope is then guided with short-range navigation towards $\mathbf{p}_w$. When the error~$\left\| \mathbf{r}-\mathbf{p_w} \right\|$ is below the target proximity threshold of~$40$~pixels the target is projected again from the mosaic image onto the current image and the fetoscope is guided with visual servoing towards another waypoint or the final target (see Figure~\ref{fig:in_vivo_nav_automated}H-I).

We extended the visual servoing part from~\cite{mattille_autonomous_2024} to make it more robust to in vivo conditions. Instead of assuming that the target is visible in the endoscopic image~$I_k$ at the magnetic field~$\hat{\mathbf{b}}$, we now only assume that $I_k$ contains enough common features with the previous image $I_t$, from which the selected target on the mosaic originates, so it is possible to determine the homography between the two images. This makes the algorithm robust to small patient movements. In the event of large patient motions, when $I_t$ and $I_k$ no longer overlap, the homography cannot be estimated, the user is informed and the mosaic must be recreated.

Since the first phase solely relies on the interpolation of the magnetic field from previously visited positions, it can guide the fetoscope to any target that is visible in the mosaic (including inside the current endoscopic view), independent of the initial position. In the second phase, the target is reached using closed-loop visual servoing, which corrects any interpolation errors from the first phase.

Compared to the method presented in our previous work~\cite{mattille_autonomous_2024}, the image Jacobian matrix~$\mathbf{J}$ is updated to support various endoscope orientations. Prior to calculating the projection from the mosaic onto $I_k$, the images were converted to grayscale and masked to mitigate reflections or insufficient illumination. This new method also leads to a higher accuracy in vivo compared to the ex vivo experiment in our prior work~\cite{mattille_autonomous_2024}.

\subsubsection*{Setups for vessel ablations on ex vivo human placenta}\label{s:ex_vivo_setups}

\begin{figure}[ht]
	\centering
    \includegraphics[width=1\textwidth]{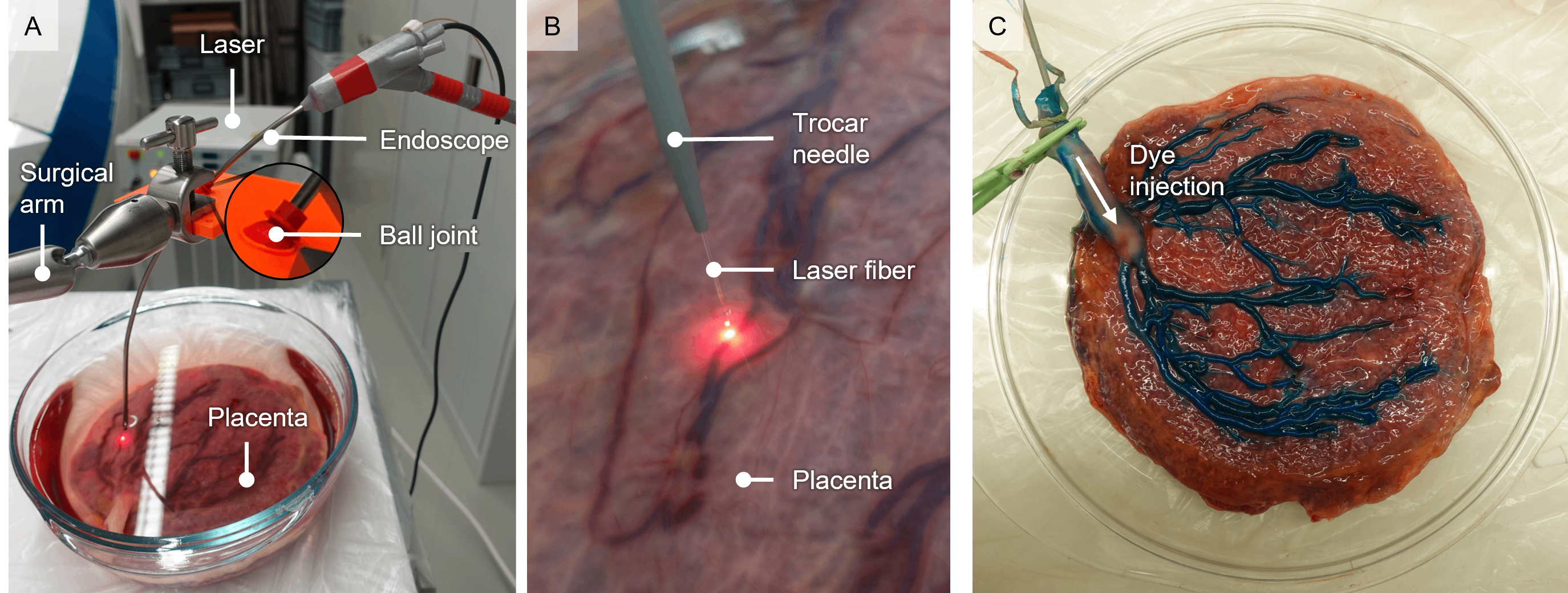} 
	\caption{\textbf{Setup of the ex vivo vessel ablation experiment.}
    (\textbf{A}) Replica of a conventional fetoscope (11508AAK, Karl Storz)
 inserted through a ball joint simulating the insertion point to ablate vessels on an ex vivo human placenta in saline solution. (\textbf{B}) Laser fiber fixed in space with a trocar needle simulating vessel ablation without hand tremors. (\textbf{C}) Dye injection through the umbilical cord for analysis of the ablations. 
        }
	\label{sfig:ex_vivo_placenta_setups}
\end{figure}

\begin{figure}[ht]
	\centering
    \includegraphics[width=1\textwidth]{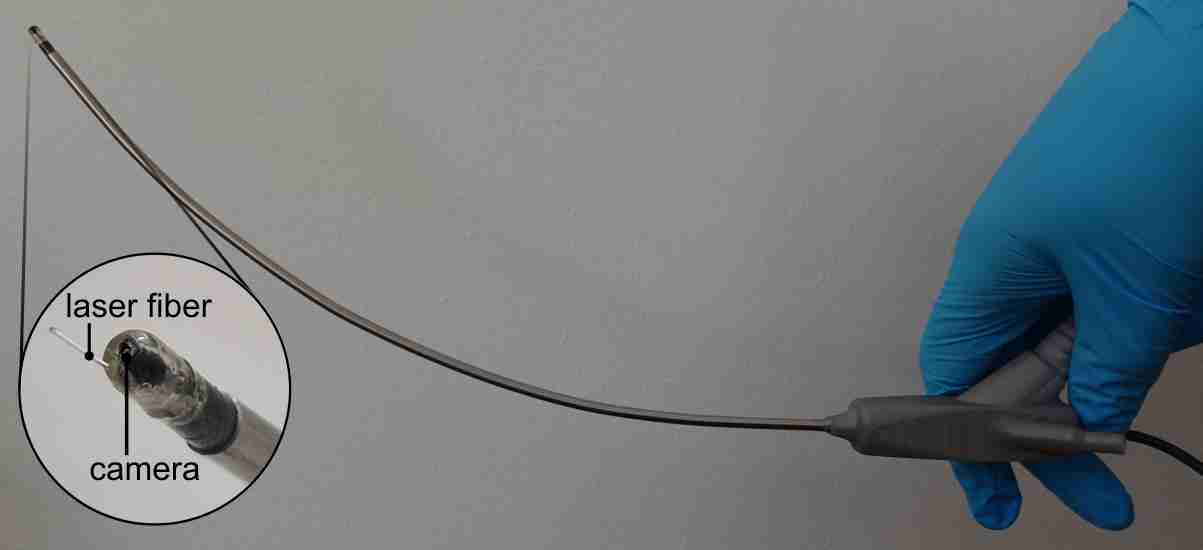} 
	\caption{
    \textbf{Replica of a rigid, precurved fetoscope.} Replica of a commercial fetoscope (model 11508AAK, Karl Storz) with integrated camera and laser fiber channel. Geometrical dimensions match those of the original fetoscope.
        }
	\label{sfig:replica}
\end{figure}

Two setups depicted in Figure~\ref{sfig:ex_vivo_placenta_setups} were used to assess vessel ablations on ex vivo human placenta:

(a) Ablation with a hand-held fetoscope: a replica of a precurved, rigid fetoscope (11508AAK, Karl Storz; see Figure~\ref{sfig:replica}) was operated manually and the ablation was affected by the hand tremors of the operator. The replica was inserted through the inner socket of a ball joint, whose outer socket is fixed in space by a surgical arm (FISSO articulated arm, Baitella AG, Switzerland). This apparatus simulated the insertion point in the maternal abdomen and the stability provided by the body wall (see Figure~\ref{sfig:ex_vivo_placenta_setups}A). Before each ablation, a placeholder ensured a distance of 2~mm between the vessel and the laser fiber, and the operator attempted to maintain this distance during the ablations. 

(b) Ablation with a fixed fiber: this setup is designed to simulate a robotic ablation where the laser fiber is positioned at a fixed location in space with a constant distance to the target vessel. During ablations, the laser fiber was held in place by the trocar needle, which was fixed in space using the surgical arm (see Figure~\ref{sfig:ex_vivo_placenta_setups}B). The laser fiber, inserted through the trocar needle, was positioned 2~mm above the target vessel using the placeholder.

\subsubsection*{Analysis of vessel ablations on ex vivo human placenta}

The vessel ablations were assessed using the protocol of Lopriore et al.~\cite{lopriore_accurate_2011}. The umbilical cord was cut to a length of a few centimeters and the amnion was peeled off for better visualization. The clotted blood from the vessels was squeezed out of the vessels through the umbilical cord and the placental vessels by gentle massage of the vessels. Saline solution was then injected through a trocar needle inserted into the umbilical vein (10~Fr, Cook Medical) and arteries (8~Fr, Cook Medical) to flush out remaining blood clots. After the saline solution was removed by massaging the vessels again, colored dye was injected through the umbilical vessels to mimic blood flow and the vessels were massaged to disperse the dye. Through visual inspection it could then be easily determined if a vessel was fully ablated or if the dye penetrated further along the vessel after the ablation point.

\subsubsection*{In vivo experiment}

The in vivo experiment was conducted in a pregnant ewe (black mountain sheep, 50~kg, 90-100~days of gestation, singleton pregnancy). Before the experiment, the endoscopic camera was calibrated in saline solution (Ri – Ac/Mal, Ringerfundin B. Braun) to mimic the properties of amniotic fluid. A median laparotomy with uterine exteriorization was performed. A trocar (Cook Medical, US) was then placed anteriorly in the amniotic sac of the fetus at a distance of 135~mm from the eMNS as the insertion point for the robotic fetoscope. 

The robotic fetoscope was pushed in the amniotic sac through an anteriorly placed trocar (Cook Medical, US). Before the experiment, the endoscopic camera was calibrated in saline solution (Ri – Ac/Mal, Ringerfundin B. Braun) to mimic the properties of amniotic fluid. 

For visualization of the fetoscope and the surgical environment a cannula (VersaOne, Medtronic, Ireland) was placed laterally to insert a custom-built rigid endoscope with a camera (OCHSA10, OmniVision) and illumination at its tip. To increase the volume of the uterus and increase visibility in the amniotic fluid, warm saline solution (Ri – Ac/Mal, Ringerfundin B. Braun) was infused into the uterus. 

\subsubsection*{Localization of the endoscopic camera for ground truth measurements}\label{s:localization_w_apriltags}

\begin{figure}[ht]
	\centering
	\includegraphics[width=0.8\textwidth]{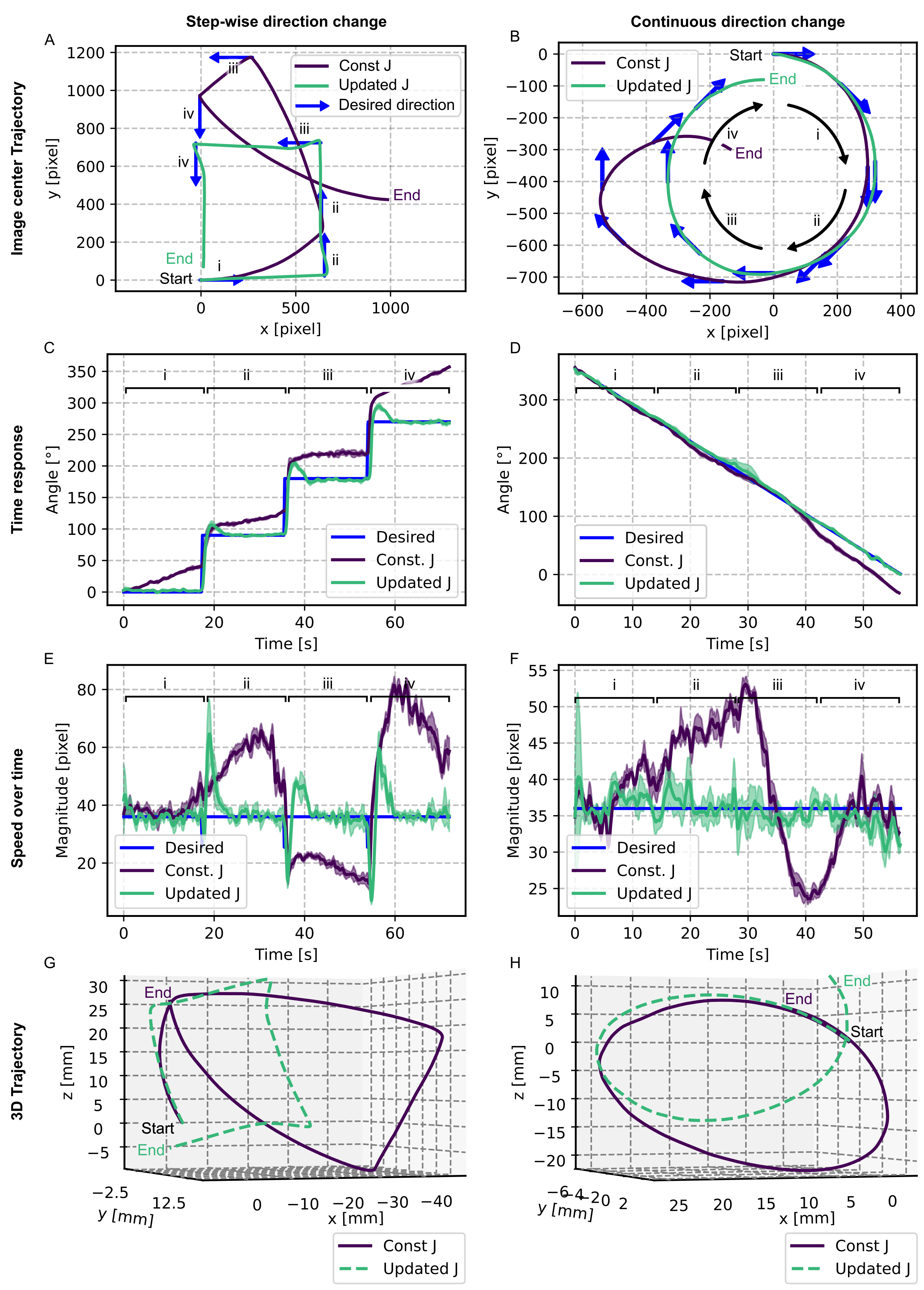} 
	\caption{\textbf{In vitro evaluation of the image-based field control methods.} Influence of updating the image Jacobian matrix during stepwise direction changes (first column) and continuous direction changes (second column). (\textbf{A-B}) Integrated velocity and desired directions. (\textbf{C-D}) Angle~$\alpha$ over time and (\textbf{E-F}) corresponding speed over time. (\textbf{G-H}) 3D trajectories of the robotic fetoscope's tip position.
        }
	\label{sfig:detailed_nav_eval}
\end{figure}

\begin{figure}[ht]
	\centering
	\includegraphics[width=1\textwidth]{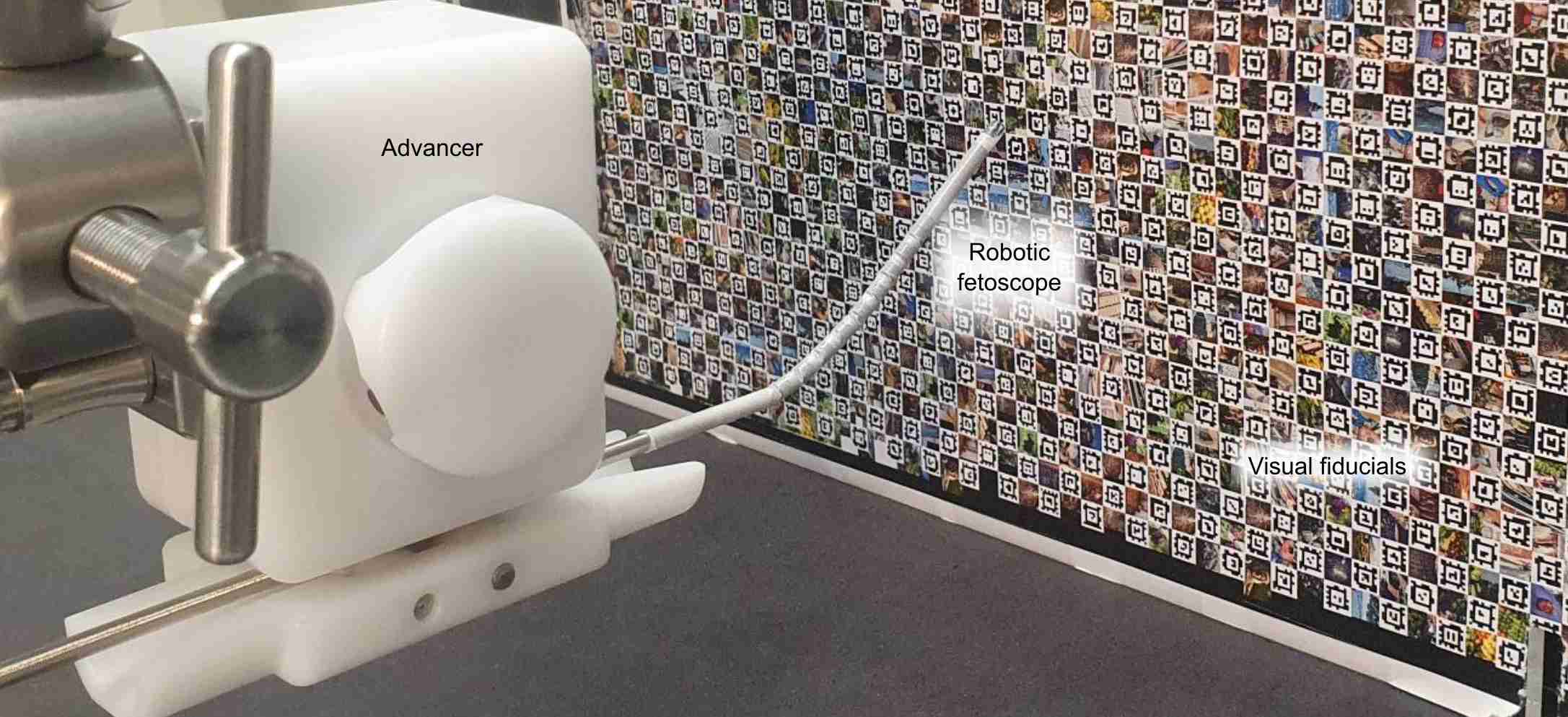} 
	\caption{\textbf{In vitro evaluation setup.}
    The robotic fetoscope is facing a plate with visual fiducials at a distance of 145~mm in front of the electromagnetic navigation system, which generates a magnetic field with 15~mT magnitude. 
        }
	\label{sfig:in_vitro_setup}
\end{figure}

To provide a ground-truth localization of the fetoscope's camera with respect to its environment, a plate with visual fiducials (AprilTags, tag family 41h12, size of 4.5~mm~\cite{olson_apriltag_2011,wang_apriltag_2016}) was used to obtain the cameras extrinsics with respect to the plate origin (see Figure~\ref{sfig:in_vitro_setup} and Figure~\ref{sfig:study_setup}B). This setup was used both for the in vitro characterization of the image-based control and for the usability study (see Figure~\ref{fig:pov_eval}, Figure~\ref{fig:user_study}, Figure~\ref{sfig:detailed_nav_eval}, and Figure~\ref{sfig:user_study_additional_metrices}). The fiducials were arranged in a checkerboard pattern, where each black square corresponded to a visual fiducial. The white squares were filled with small images to mitigate the challenges posed by repetitive patterns for the steering algorithms.

\begin{figure}[ht]
	\centering
	\includegraphics[width=1\textwidth]{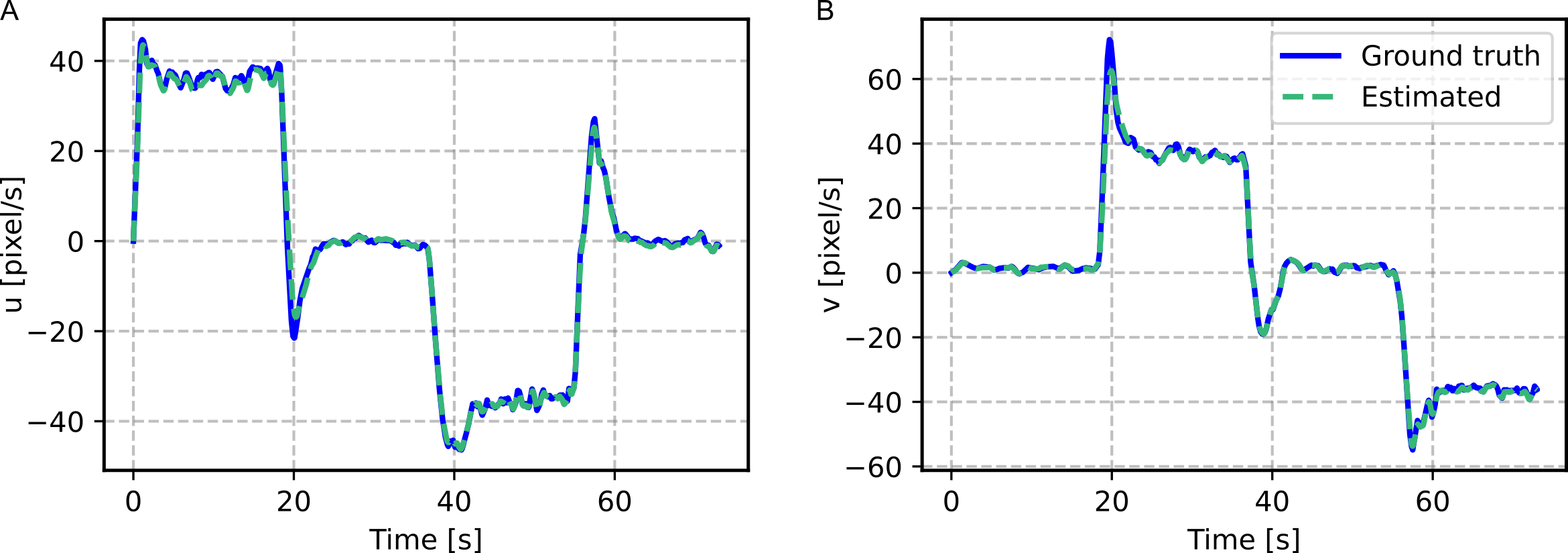} 
	\caption{\textbf{In vitro evaluation of the estimated image velocity.} Estimated and ground truth motion of the image center in the (\textbf{A}) x-direction and (\textbf{B}) in the y-direction. Both signals show the mean values over time over 10 runs.} 
	\label{sfig:detailed_nav_eval_est_vs_truth}
\end{figure}

The ground truth signal of the velocity of the center of the image shown in Figure~\ref{sfig:detailed_nav_eval_est_vs_truth} was calculated based on the camera poses obtained over the localization of the plate with the visual fiducials. The velocity is calculated over~$\mathbf{v} = \Delta \mathbf{x}/ \Delta T$, where $\mathbf{x} = {}_{k}\mathbf{s}_k -{}_{k-1}\mathbf{s}_k$ is the difference between the current image center and the projection of the previous image center on the current image frame and $\Delta T = 0.04~s$ is the difference in time between the frames. The projection between the images $I_{k-1}$ and $I_k$ is calculated based on the camera poses.

\subsubsection*{Usability study setup}

\begin{figure}[ht]
	\centering
	\includegraphics[width=1\textwidth]{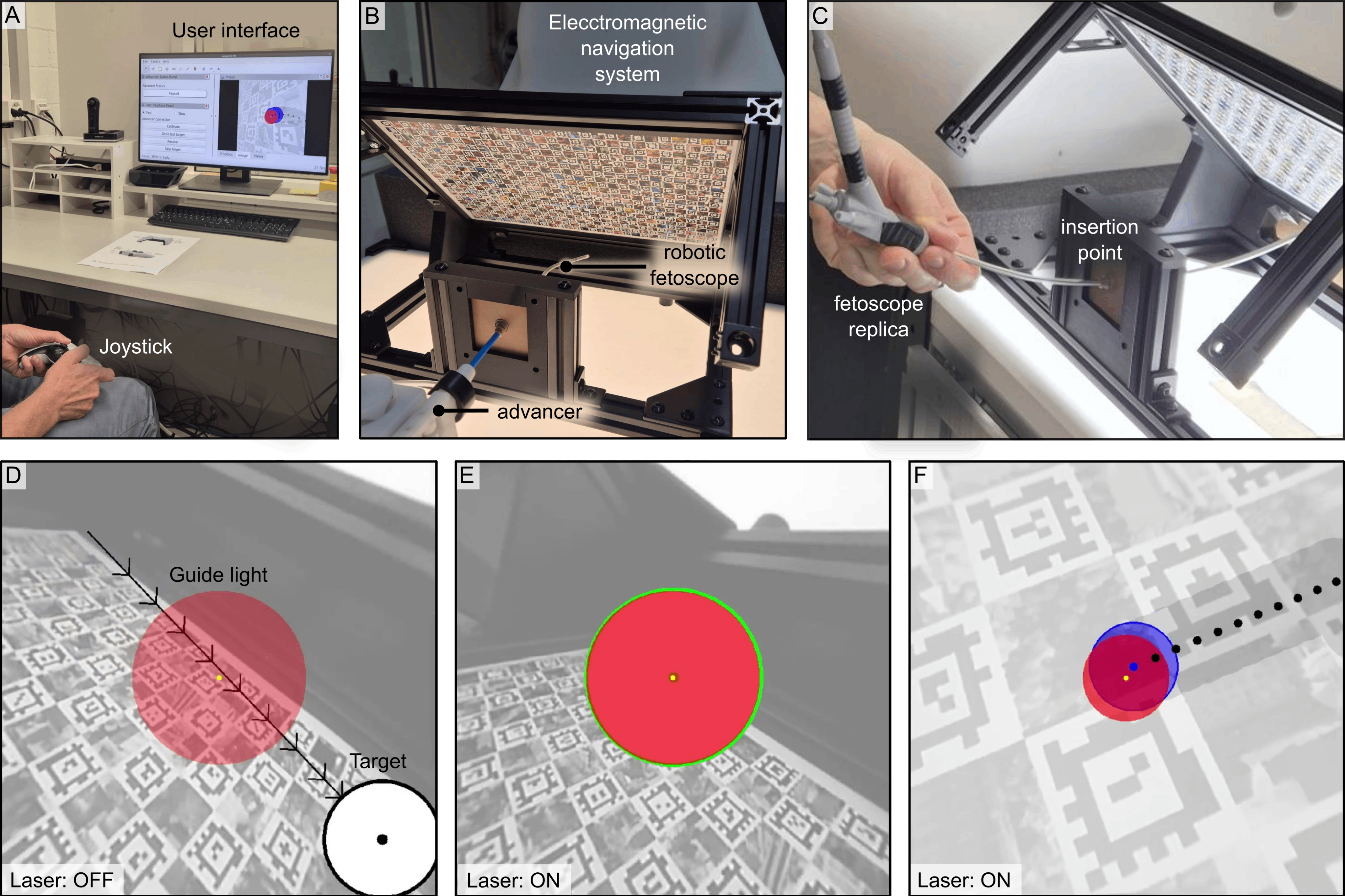} 
	\caption{\textbf{Usability study setup.} The setup of the usability study is shown (\textbf{A}-\textbf{B}) for the robotic fetoscope steered over a joystick and (C) for the hand-held replica of the commercial fetoscope (11508AAK, Karl Storz). An augmented reality setup with an insertion point for the fetoscopes and a tilted plate with visual fiducials was used for camera localization and a foot pedal for simulated ablations. The aim of the study was to (\textbf{D}-\textbf{E}) follow arrows in the image to reach targets and ablate them or (\textbf{F}) ablate continuously a line consisting of many targets.
    }
	\label{sfig:study_setup}
\end{figure}

The usability study was performed in an augmented reality environment using a tilted plate with visual fiducials (see Section~``Localization of the endoscopic camera for ground truth measurements"), and a fixed ball joint in space served as the insertion point for the two devices (see Figure~\ref{sfig:study_setup}). Participants steered the fetoscopes based on visual feedback consisting of the fetoscopic camera images, which were augmented with task-specific overlays including a simulation of the laser guide, and ablation targets. 

The coordinates~${}_w \mathbf{p}_t \in \mathbb{R}^{3}$ of each ablation target during the usability study were defined in a world reference frame (left-subscript ``$w$") attached to the fiducial plate (see Figure~\ref{sfig:study_setup}) and based on a virtual uterus model (see Figure~\ref{fig:user_study}A). Given the pose of the camera estimated through the visual fiducials (see Section~``Localization of the endoscopic camera for ground truth measurements"), the targets were projected onto the endoscopic images to provide an augmented reality view to the user, using perspective projection with a pinhole camera model

\begin{equation}
    \lambda \begin{bmatrix}{}_i \mathbf{p}_t \\ 1 \end{bmatrix} = \mathbf{K} \begin{bmatrix} \mathbf{R} & \mathbf{t} \end{bmatrix} \begin{bmatrix}{}_w \mathbf{p}_t \\ 1 \end{bmatrix}
    \label{eq:pinhole_projection}
\end{equation}
\noindent where $\lambda$ is a scale factor, $\mathbf{K}$ denotes the projection matrix, which projects world points to the rectified camera images and was determined through calibration, the pose of the camera was defined as the rotation $\mathbf{R}$ and translation $\mathbf{t}$ and ${}_i \mathbf{p}_t \in \mathbb{R}^{2}$ are the coordinates of target projected in the image reference frame (left-subscript ``$i$").

The target was overlaid on the endoscopic image as a circle centered at~${}_i \mathbf{p}_t$. Its radius~$r$ was scaled according to the distance between the camera and the target, so that it matched the radius of the laser guide when the desired distance for ablation was reached (see Movie~S7 and Movie~S8).

Pressing a foot pedal simulated laser ablation, which was shown as an intensity change of the laser light in the camera images (see Figure~\ref{sfig:study_setup}E). For the robotic platform, both the robotic fetoscope and the advancer were controlled via a joystick (PlayStation 5, Sony, Tokyo, Japan). Users were able to switch between two predefined speed modes of the fetoscope and advancer designed for fast and precise navigation. The replica was manually steered. A change in target color indicated whether the fetoscope was close enough to ablate a target. To motivate participants to improve accuracy, another color change was performed for very precise targets (in-plane error below $160~\mu m$ and depth error below 1~mm). A target was considered to be reached if it was ablated during 10~s for the first task and 1~s for the second task. Users had the option of skipping a target during point ablations if they were not able to reach it.

\subsubsection*{Usability study data collection, processing and evaluation}

\begin{figure}[ht]
	\centering
	\includegraphics[width=0.5\textwidth]{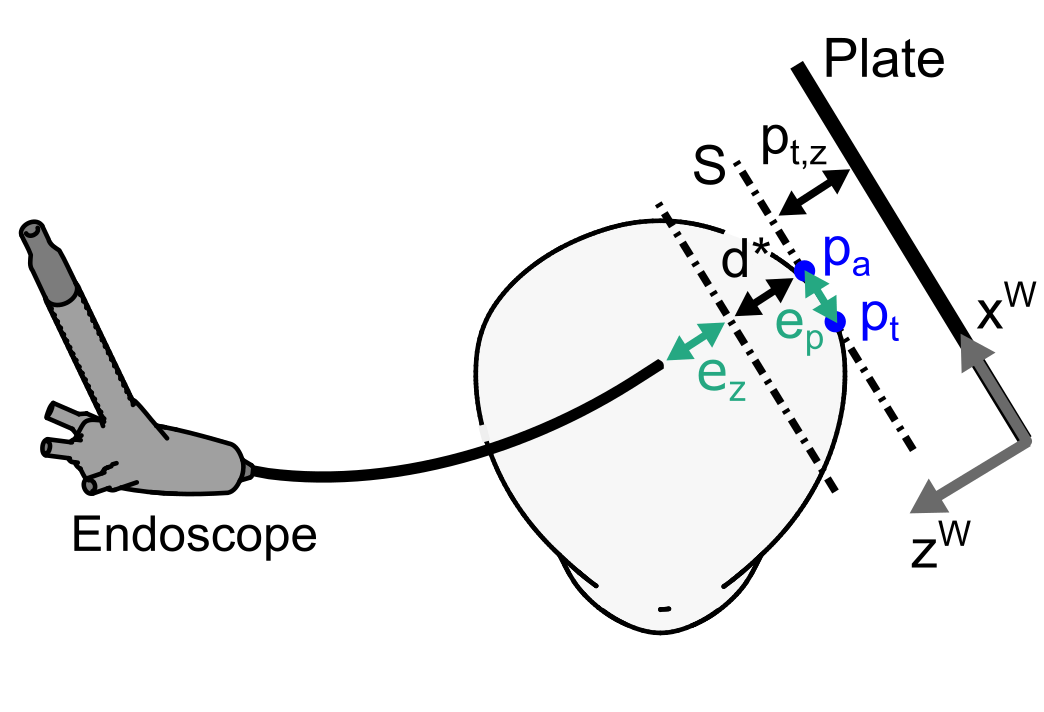} 
	\caption{\textbf{Errors definition in usability study.} Illustration of the depth error~$e_z$ and the in-plane error~$e_p$.}
\label{sfig:study_errors}
\end{figure}

For each ablation point, a depth error~$e_z$ and an in-plane error~$e_p$ were evaluated (see Figure~\ref{sfig:study_errors}). The depth error was defined in the z-direction of the world frame as the difference between the desired distance~$d^*$ and actual distance~$d_{cam}$ between the target and the camera

\begin{equation}
    e_z = \lvert d^* - d_{cam} \lvert
\end{equation}

This metric was only used to assess whether the fetoscope was in a depth range that allowed the user to ablate, which was indicated to the user in the graphical user interface. The in-plane error was defined as the distance between the target~$\mathbf{p}_t$ and the virtual ablation point~$\mathbf{p}_a$ which coordinates are known in the world reference frame

\begin{equation}
    e_p = \|\mathbf{p}_t -\mathbf{p}_a \|_2
    \label{eq:e_p}
\end{equation}

The point~$\mathbf{p}_a$ was defined as the intersection between the laser ray and the plane~$S$, which is the plane passing through the target~$ \mathbf{p}_{t}$ and is parallel to the xy-plane of the world reference frame (see Figure~\ref{sfig:study_errors}). 

For the first task, the accuracy, precision, navigation phase duration and ablation phase duration were reported for each participant~$s$ as shown in Figure~\ref{fig:user_study}C and Figure~\ref{sfig:user_study_additional_metrices}C. 
For each target~$t$ in run~$j$ of participant~$s$, the accuracy and precision were calculated as the mean and standard deviation of the in-plane error~$e_p(s,t,j,n)$ across every sample~$n$. For each metric~$m(s,t,j)$, the median across runs was first calculated per target, giving $m(s,t)$. The median of~$m(s,t)$ across all targets then provided a participant-specific value~$m(s)$ for each metric as shown in Figure~\ref{fig:user_study}. Targets were discarded if they were not reached, skipped or if a user forgot to switch of the laser in between targets leading to unwanted higher errors during the ablation and faster ablation times. In addition, if the pedal was pressed and no target was visible in the camera image the ablation was discarded for the evaluation as it was an error by the user independent of the used device.

For the second task, the accuracy, precision and task duration were reported for each participant~$s$ as shown in Figure~\ref{fig:user_study}F and Figure~\ref{sfig:user_study_additional_metrices}D. The coordinates of the virtual ablation points~${}_w \mathbf{p}_a(n)$ were calculated for each sample~$n$ when the user had the foot pedal pressed for ablation using the inverse perspective projection. The other trajectory segments were discarded. Each of those ablated trajectory segments was then resampled over time to ensure constant time intervals between the samples. For each participant~$s$ and run~$j$, the accuracy was defined as a weighted median of $e_p(s,j,n)$ across every sample~$n$, where~$\mathbf{p}_t(n)$ was defined as the closest position to~$\mathbf{p}_a(n)$ along the optimal trajectory. For each pair of consecutive samples~$n-1$ and~$n$, a mean error~$\bar{e}_p(s, j, n)$ and a weight~$w(s, j, n)$ were first calculated as
\begin{equation}
\bar{e}_p(s, j, n) = \frac{e_p(s, j, n-1) + e_p(s, j, n)}{2}
\end{equation}

and

\begin{equation}
w(s, j, n) = \lVert e_p(s, j, n) - e_p(s, j, n-1) \rVert_2,
\end{equation}

\noindent and the weighted median~$\bar{e}_{p_d}$ was then found as the value of~$\bar{e}_p$ at sample~$n$ for which the cumulative weight~$w(s, j, n)$ of all samples with $\bar{e}_p \leq \bar{e}_{p_d}$ was at least half of the total weight:

\begin{equation}
\sum_{\bar{e}_p(s, j, n) \leq \bar{e}_{p_d}} w(s, j, n) \geq \frac{1}{2} \sum_{n} w(s, j, n).
\end{equation}

Precision was calculated as the weighted median of the median absolute deviations of $e_p(s,j,n)$, using the Euclidean distances between ablated trajectory points as weights. The median over three runs per participant was then computed for all metrics $m(s,j)$, yielding the value of $m(s)$ shown in Figure~\ref{fig:user_study}.

%The statistical significance in the results of the user study was obtained over a paired Wilcoxon signed-rank test. The degree of statistical significance is illustrated in the figures with star annotations, where one star corresponds to $p<0.05$, two stars represent $p<0.01$ and three stars $p<0.001$.

\subsubsection*{Additional results of the usability study}

\begin{figure}[ht]
	\centering
	\includegraphics[width=1\textwidth]{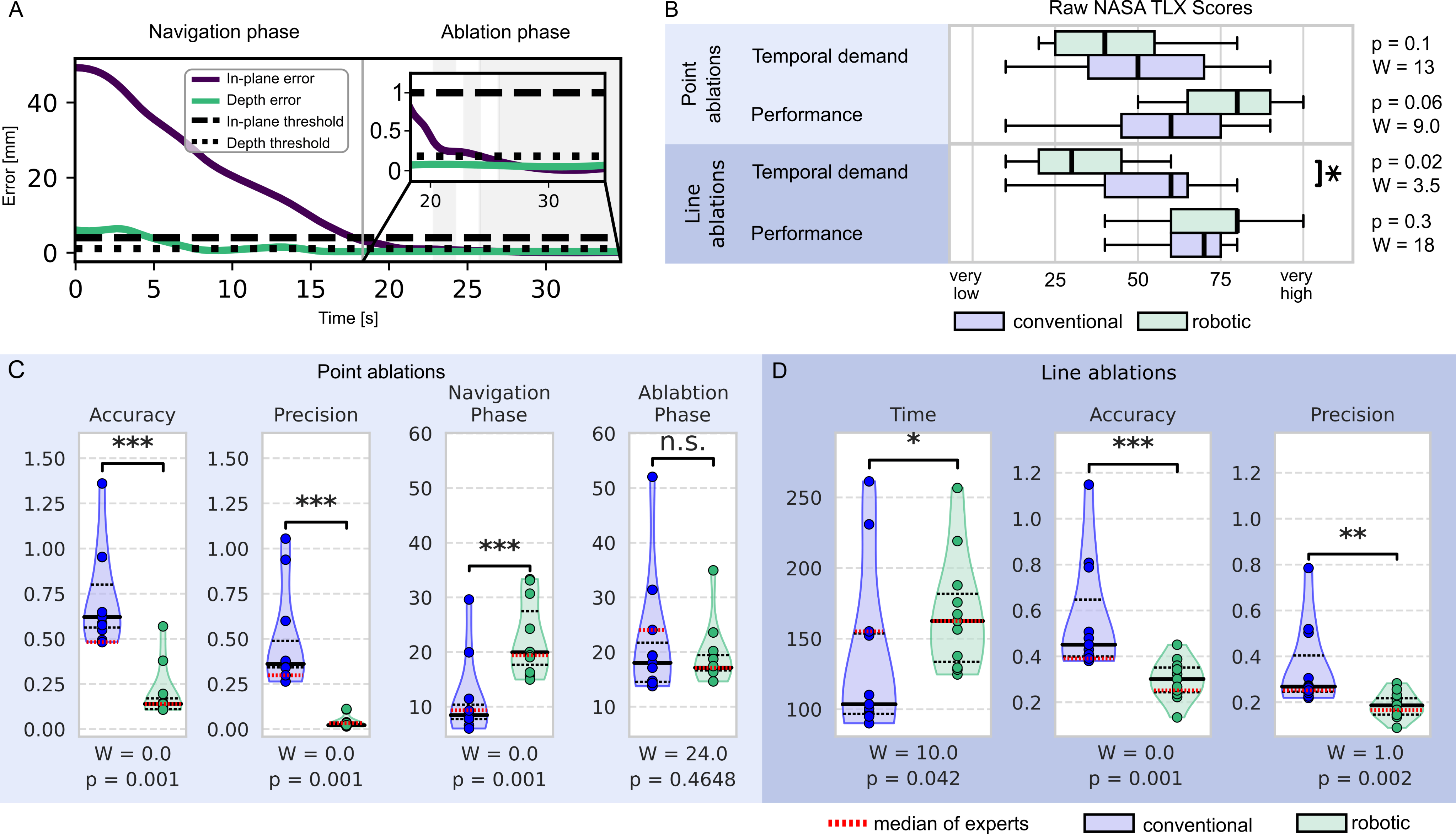} 
	\caption{\textbf{Additional results of the usability study.} (\textbf{A}) Illustration of the navigation phase and ablation phase for point ablation task. (\textbf{B}) Raw NASA TLX scores of the perceived temporal demand and performance of the two devices. The boxes illustrate the IQR, the center lines indicate the median, and the whiskers show the range within 1.5$\times$IQR. (\textbf{C-D}) Performance metrics for the two task with the median of the expert group indicated in red. The center lines denote the median and the dotted lines indicate the IQR. Statistical significance was assessed over a paired Wilcoxon signed-rank test: *$p<0.05$, **$p<0.01$ and ***$p<0.001$.
    }
	\label{sfig:user_study_additional_metrices}
\end{figure}

Figure~\ref{sfig:user_study_additional_metrices} shows the results of the navigation and ablation durations for the point ablation task and the overall duration for the line ablation tasks. The navigation duration was defined as the time it took to navigate from the previous target to the first point in time it was possible to ablate the current target regardless of when the participants actually started ablating it. This was defined as the first point in time when both errors were below their respective predefined thresholds (see Figure~\ref{sfig:user_study_additional_metrices}A). The navigation duration was significantly higher for the robotic fetoscope with 20~s compared to 8.5~s for the point ablation task and 162~s and 104~s in the line ablation task whereas no significant difference for the ablation duration was observed (see Figure~\ref{sfig:user_study_additional_metrices}). This is explained by the speed limit included for safety reasons during the navigation of the robotic fetoscope, as the navigation speed of the manual device could not be restricted. Unlike during a real surgery, participants could navigate extremely fast with the manual device as they did not have to memorize the environment. Besides, the visibility was perfect and no fetuses were present in our augmented reality setup. The navigation duration was in an acceptable range for both devices for the surgery according to the fetal surgeon who participated in the study. 

\subsubsection*{Cystoscope}
The magnetic cystoscope (Figure~\ref{fig:other_app}A and Figure~\ref{sfig:cysto_setup}) contains a CMOS camera (OVM6946, OmniVision) in its transparent tip (OD of 2.6~mm). The steerable, flexible section behind the tip comprises a chain of alternating axially magnetized NdFeB ring magnets and stainless steel ball joints (OD of 2.6~mm, ID of 1.1~mm). Its distal end consists of a 200~mm long extension spring (OD of 2.6~mm, ID of 1.5~mm, Durovis, Switzerland). 

\subsubsection*{Cystoscopy setup}

\begin{figure}[ht]
	\centering
	\includegraphics[width=0.5\textwidth]{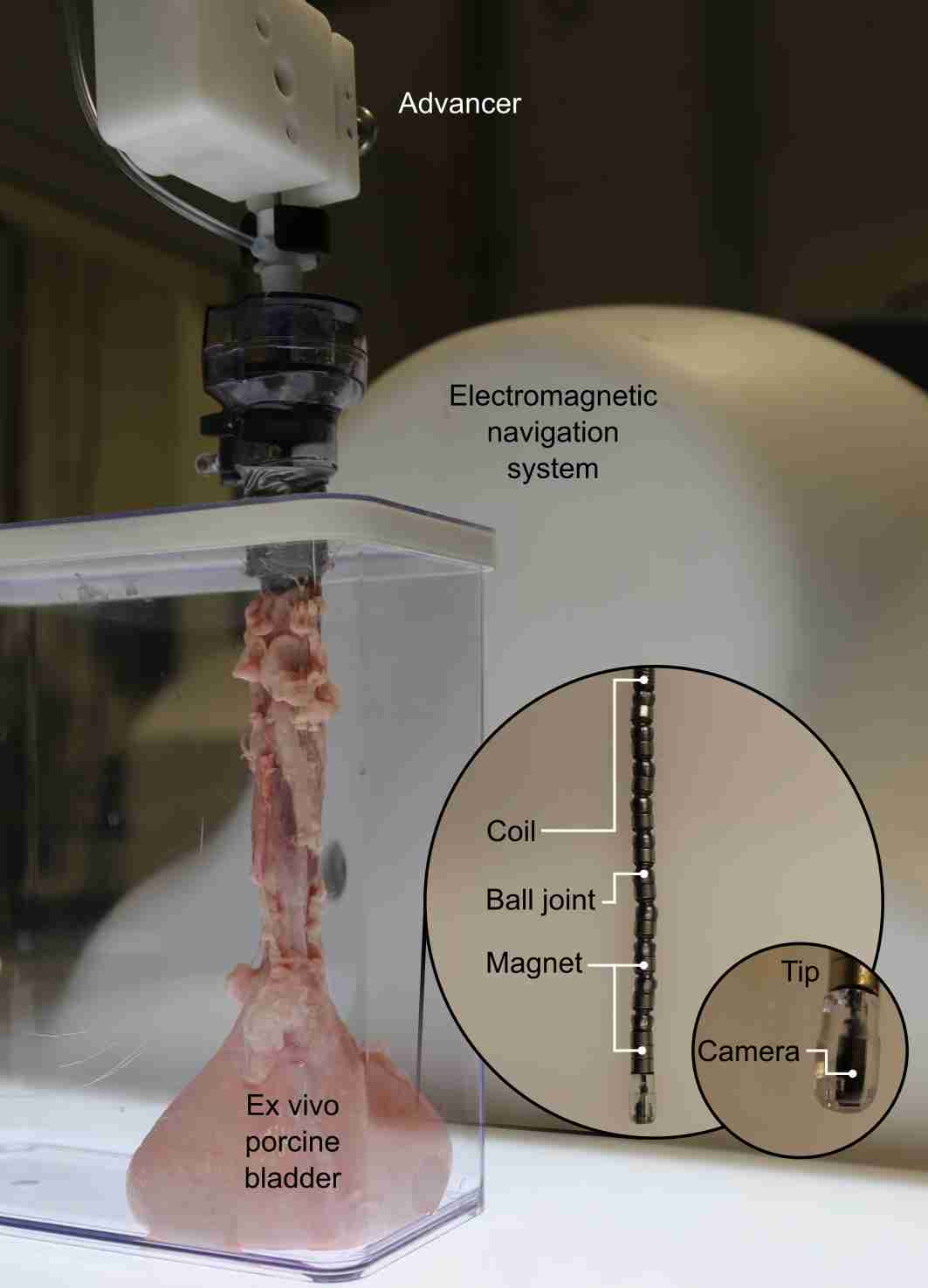} 
	\caption{\textbf{Ex vivo porcine urinary bladder model setup.}
    A magnetic cystoscope is inserted into an ex vivo porcine urinary bladder filled with urine. The flexible distal section is steered by an electromagnetic navigation system, and the insertion depth is controlled by an advancer unit.
    }
	\label{sfig:cysto_setup}
\end{figure}

An ex vivo porcine urinary bladder was placed 180~mm in front of the eMNS. The flexible, magnetic cystoscope was inserted through the urethra into the urine-filled bladder. The bending of its tip was controlled with the manual and automated navigation methods in a magnetic field of 18~mT. The insertion depth was changed by a mechanical advancer unit (see Figure~\ref{fig:workspace}A). The setup is shown in Figure~\ref{sfig:cysto_setup}.

\subsubsection*{Gastroscopy setup}
\begin{figure}[ht]
	\centering
	\includegraphics[width=1\textwidth]{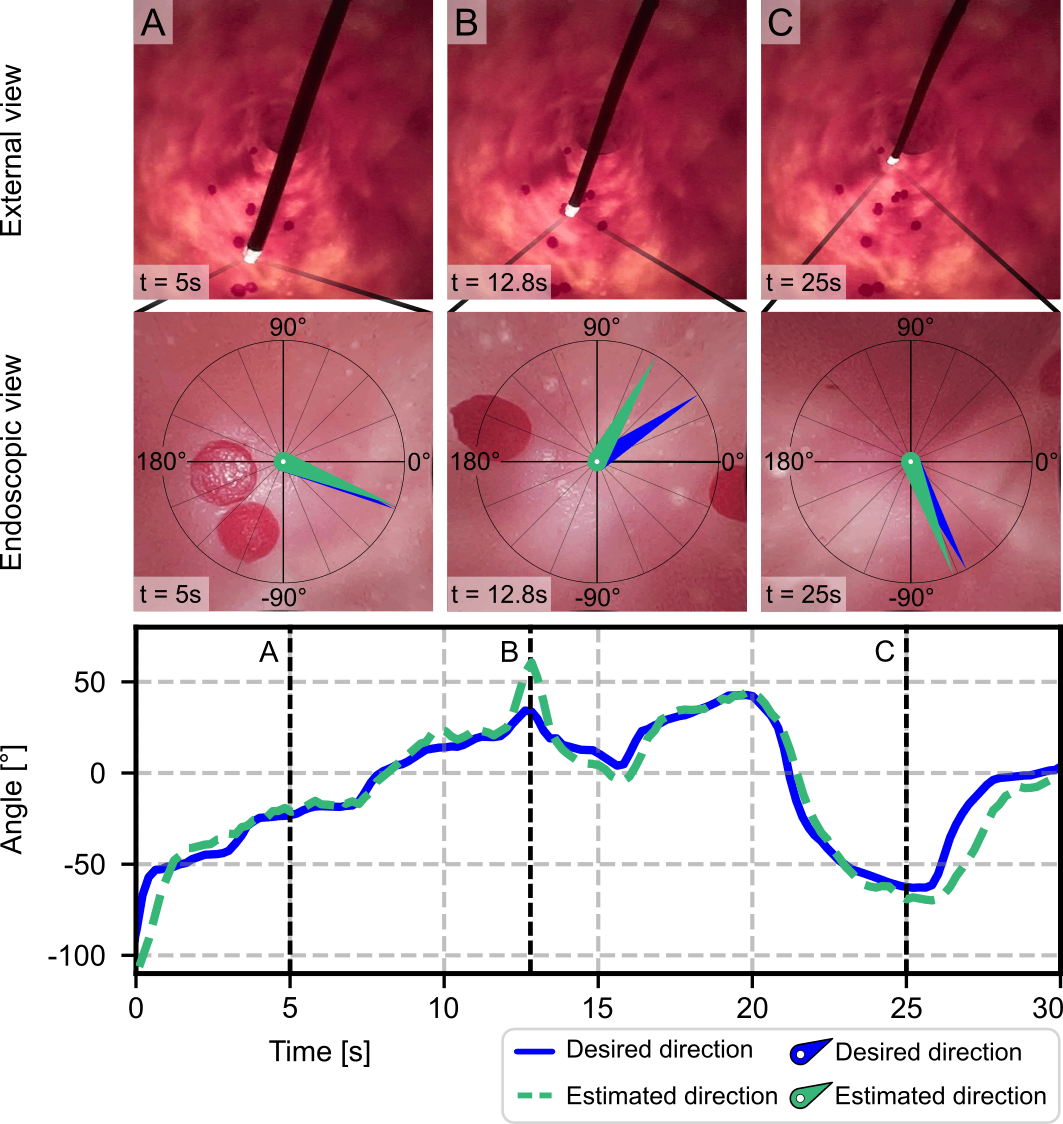} 
	\caption{\textbf{Manual navigation of a magnetic gastroscope without biopsy forceps.}
    Comparison of desired navigation directions from joystick inputs and estimated actual directions over time. (\textbf{A}--\textbf{C}) External and endoscopic views of the gastroscope in the stomach model at three points in time, overlaid with respective desired and actual navigation directions.
    }
	\label{sfig:tne_pov}
\end{figure}

\begin{figure}[ht]
	\centering
	\includegraphics[width=1\textwidth]{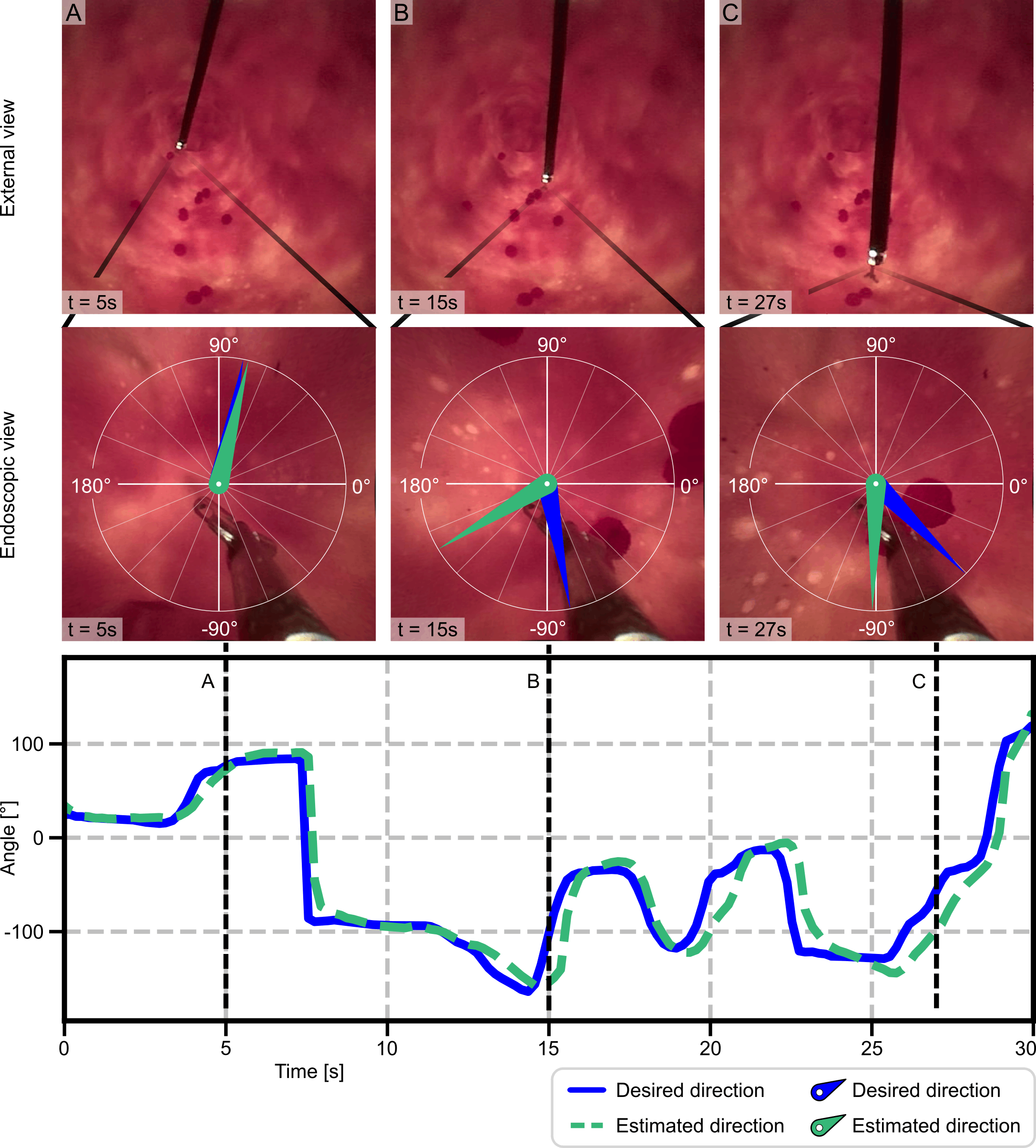} 
	\caption{\textbf{Manual gastroscope navigation with an inserted biopsy forceps.} Comparison of desired navigation directions from joystick inputs and estimated actual directions over time. (\textbf{A}--\textbf{C}) External views of the magnetic gastroscope with an inserted biopsy forceps at three points in time and corresponding endoscopic images, each overlaid with their respective desired and actual navigation directions.
    }
	\label{sfig:tne_pov_biopsy}
\end{figure}

\begin{figure}[ht]
	\centering
	\includegraphics[width=1\textwidth]{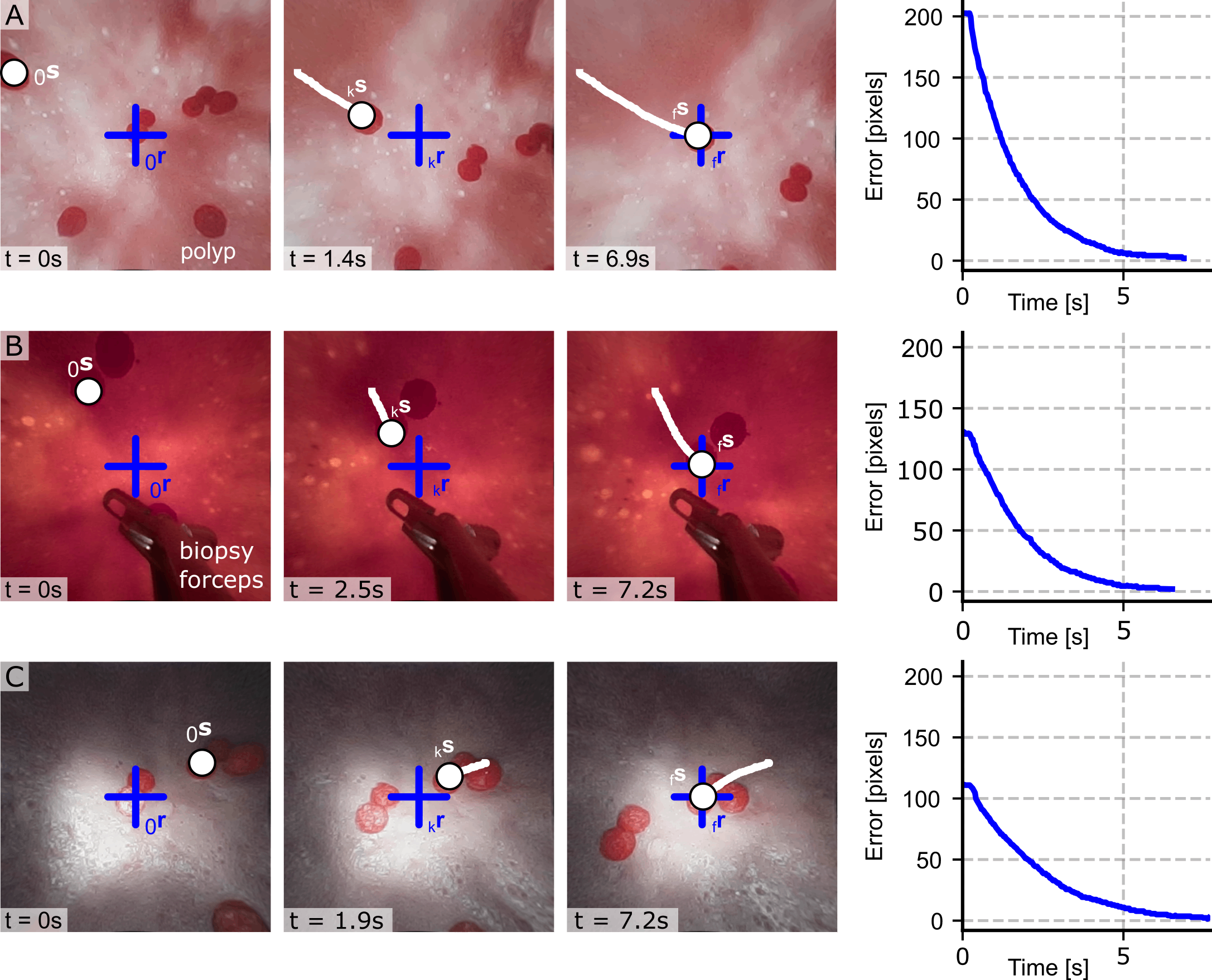} 
	\caption{\textbf{Short-range navigation of a magnetic gastroscope.} 
    Navigation to bring a target~$\mathbf{s}$ to a reference position~$\mathbf{r}$ in the endoscopic image with (\textbf{A}) external illumination, (\textbf{B}) increased device stiffness and a partially occluded image caused by the insertion of a biopsy forceps along with altered camera settings, and (\textbf{C}) solely with endoscopic illumination. The columns represent the initial endoscopic image~$I_0$, an intermediate frame~$I_k$, the final frame~$I_f$ with the reached target, and the error evolution over time. 
    }
	\label{sfig:tne_vs_in_img}
\end{figure}

\begin{figure}[ht]
	\centering
	\includegraphics[width=0.5\textwidth]{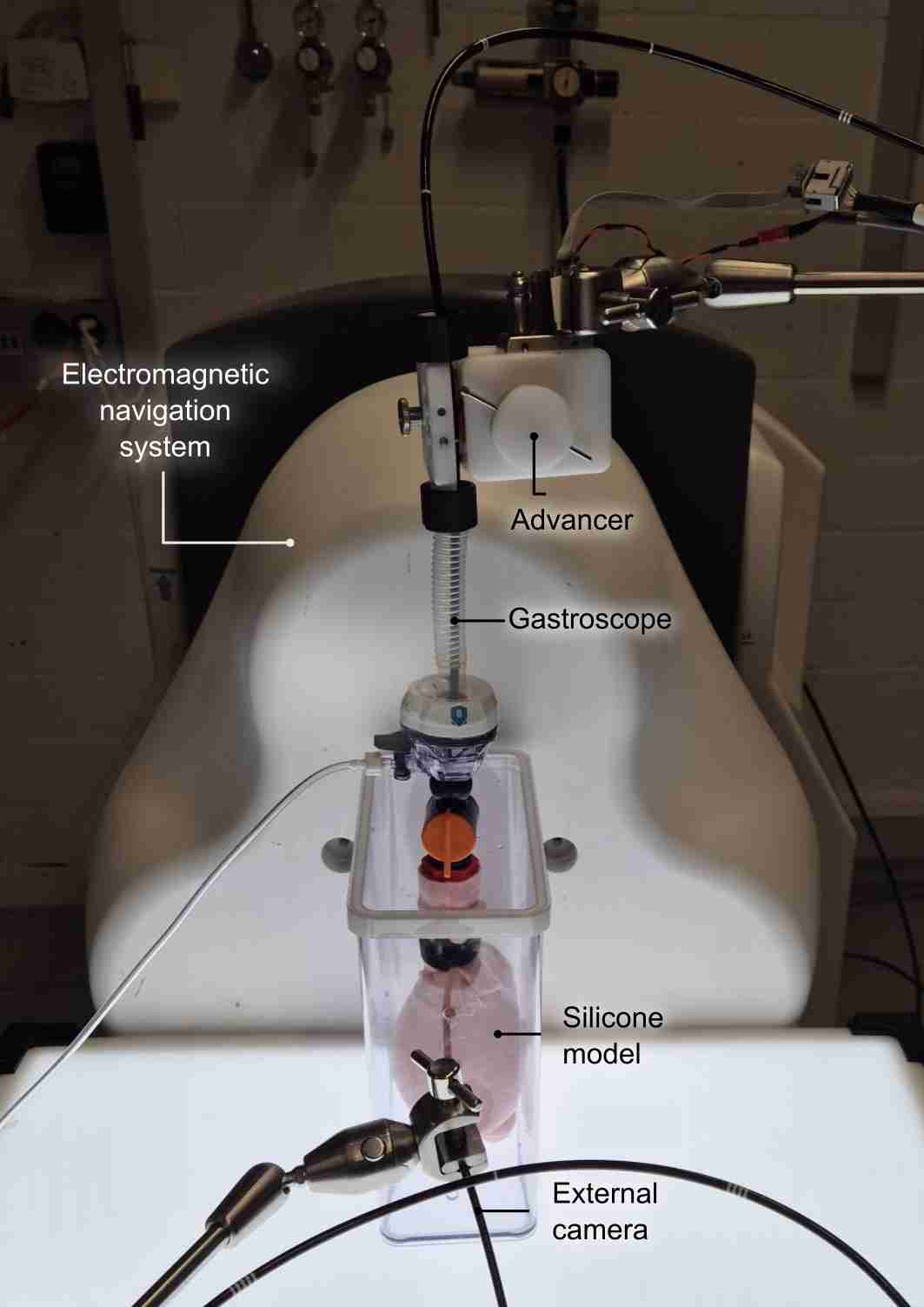} 
	\caption{\textbf{Setup of the magnetic gastroscope in stomach model.}
    The magnetic gastroscope is inserted into a stomach model. The bending of its distal section is steered over magnetic fields generated by the electromagnetic navigation system, and its insertion depth is controlled by the mechanical advancer.
    }
	\label{sfig:tne_setup}
\end{figure}

The magnetic gastroscope proposed in our previous work~\cite{mesot_teleoperated_2025} was inserted into a wet silicone human stomach model (SynDaver Labs, USA) at a distance of 200~mm in front of the eMNS and controlled with the manual and automated navigation methods with a magnetic field of 22~mT (see Figure~\ref{sfig:tne_setup}). 3D-printed polyps (Polymaker PLA, Prusa Research MK4) were sutured to the inner stomach wall to demonstrate the ability of taking biopsy samples (see Figure~\ref{fig:other_app}G and Movie~S14). The gastroscope was steered both with and without a biopsy forceps (SpyBite\texttrademark{} Max biopsy forceps, Boston Scientific, USA) inserted in its working channel and under various illumination conditions and camera settings, and both with manual and automated control (see Figures~\ref{sfig:tne_pov}-\ref{sfig:tne_vs_in_img} and Movie~S14, Movie~S15, Movie~S17 and Movie~S18). 
A second non-magnetic endoscope was inserted through the side of the stomach model to provide an external view of the gastroscope, which was only used for visualization purposes and has no influence on the experiment.

%%%%%%%%%%%%%%%% SUPPLEMENTARY FIGURES %%%%%%%%%%%%%%%

\clearpage
\subsection*{Supplementary Figures}

\begin{figure}[ht]
	\centering
	\includegraphics[width=0.46\textwidth]{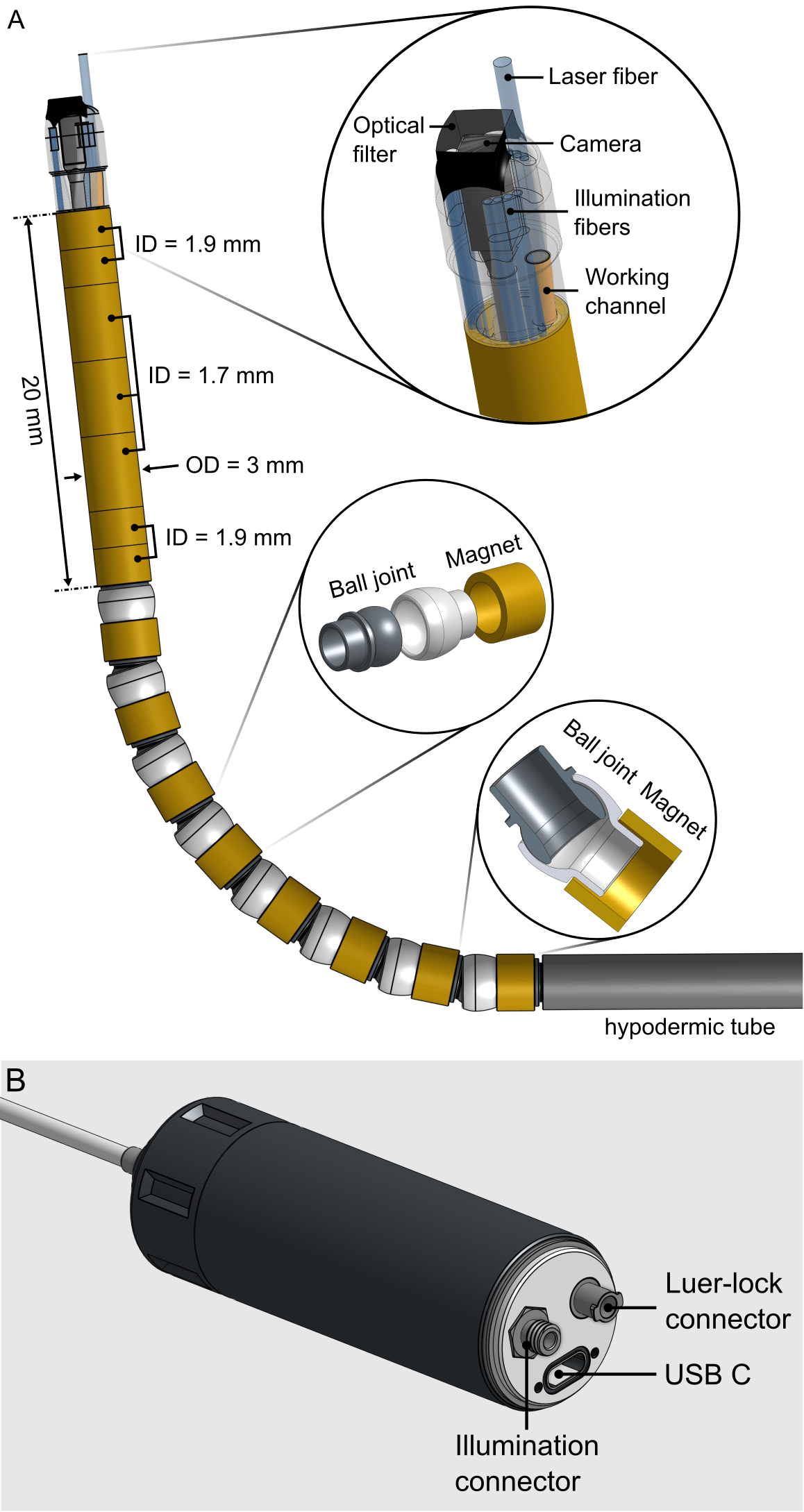} 
	\caption{\textbf{Detailed fetoscope design.}
    (\textbf{A}) Illustration of the flexible distal section without the Pebax\texttrademark{} jacket with close-ups of the tip design, an exploded view of a ball joint, and a sectional view of a ball joint. (\textbf{B}) Casing of the fetoscope's camera board with connectors for illumination, USB-C connector for the camera, and a luer-lock connector to insert the laser fiber.
    }
	\label{sfig:fetoscope}
\end{figure}

\begin{figure}[ht]
	\centering
	\includegraphics[width=1\textwidth]{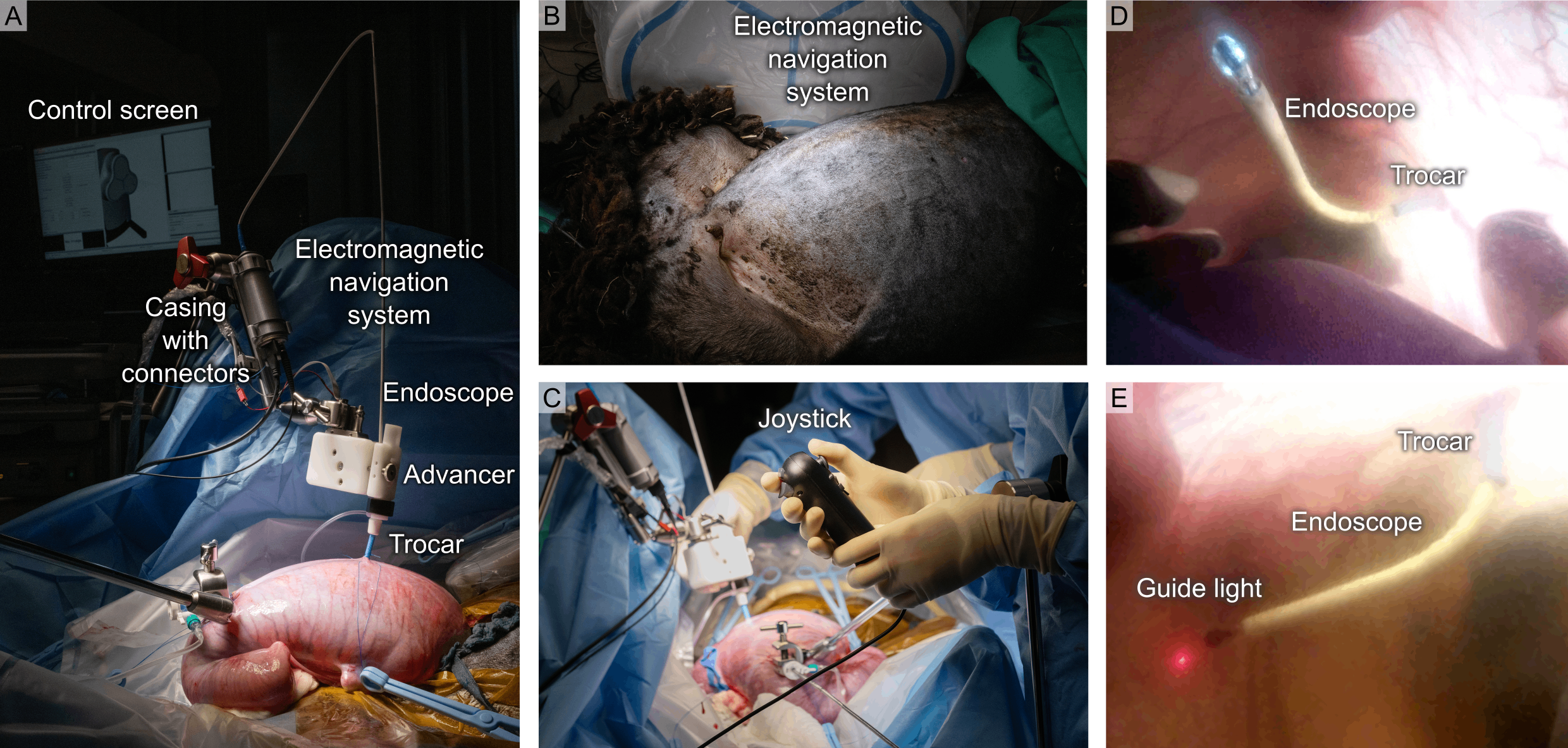} 
	\caption{\textbf{In vivo experimental setup.}
    (\textbf{A}) Exteriorized ovine uterus and placement of the robotic fetoscope. (\textbf{B}) Placement of the ovine model in front of the electromagnetic navigation system prior to the procedure. (\textbf{C}) Manual navigation of the robotic fetoscope. (\textbf{D}) In utero external view on the navigation of the robotic fetoscope and (\textbf{E}) with the guide light of the laser turned on.
        }
	\label{sfig:in_vivo_setup}
\end{figure}

\begin{figure}[ht]
	\centering
	\includegraphics[width=1\textwidth]{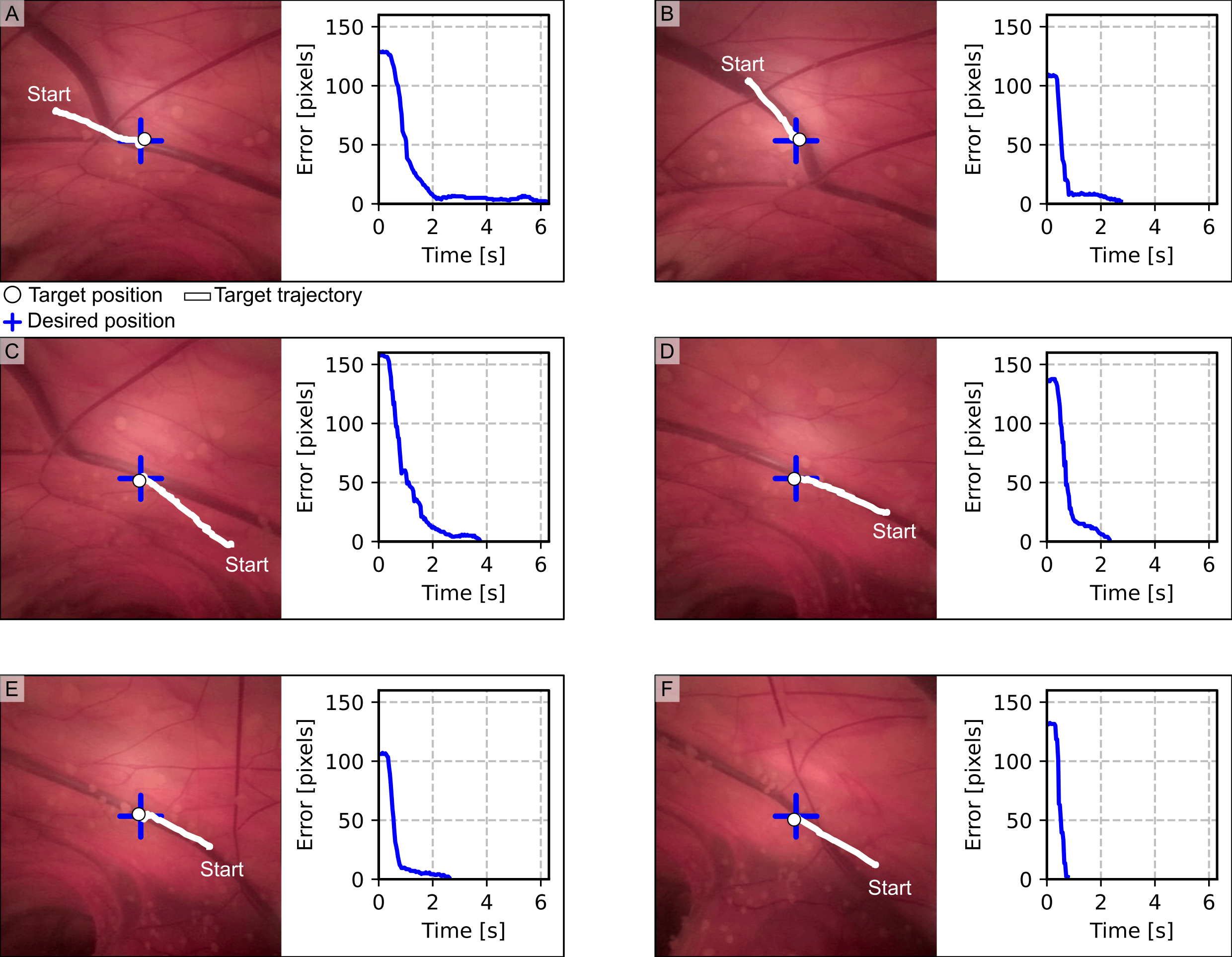} 
	\caption{\textbf{In vivo short-range navigation.} Trajectory of the tracked target position over time and plots of the pixel error over time for six different targets.
        }
	\label{sfig:in_vivo_vs_in_img}
\end{figure}

\begin{figure}[ht]
	\centering
	\includegraphics[width=0.5\textwidth]{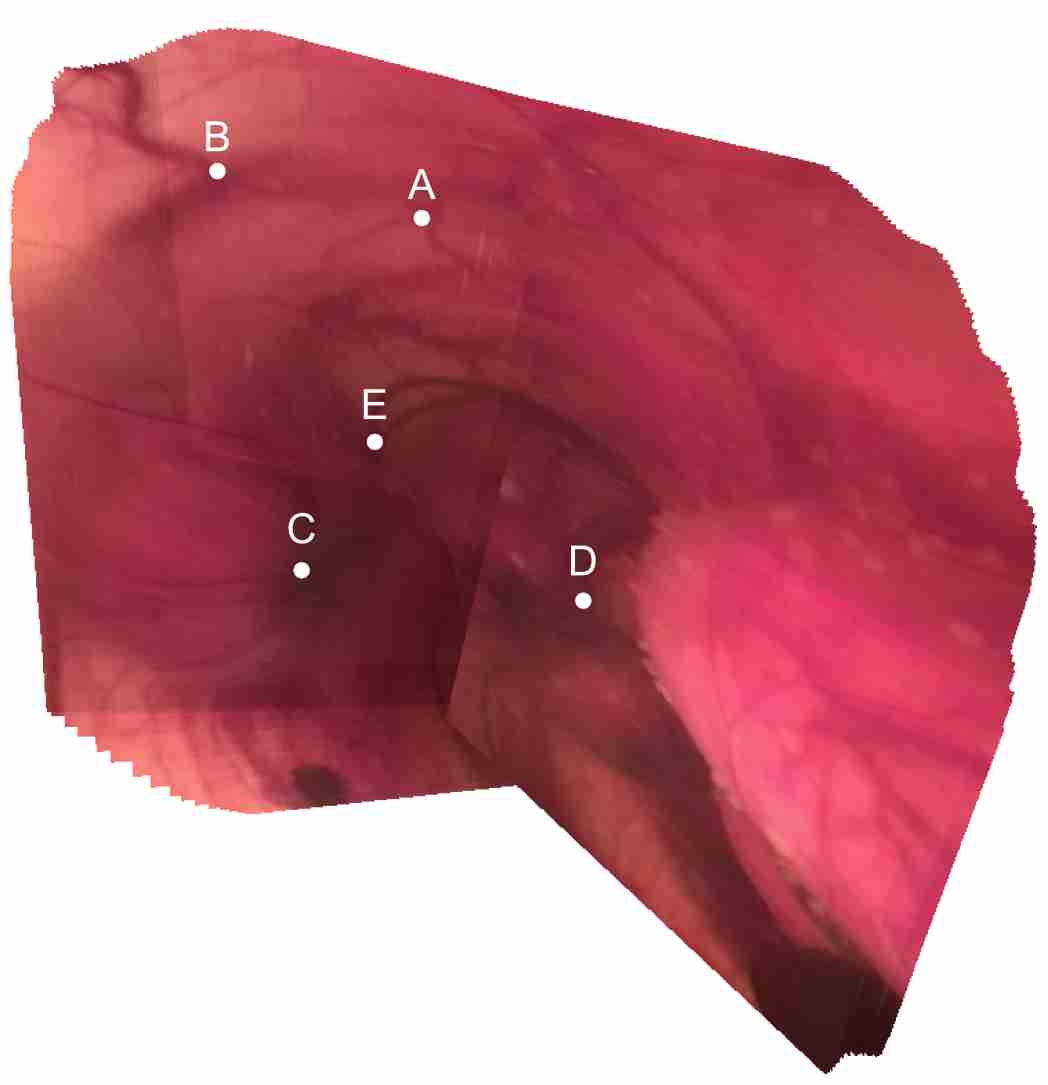} 
	\caption{\textbf{In vivo image mosaic with target locations.} Selected target locations of the long-range navigation in vivo.
    }
	\label{sfig:mosaic_w_all_targets}
\end{figure}

\begin{figure}[ht]
	\centering
	\includegraphics[width=0.7\textwidth]{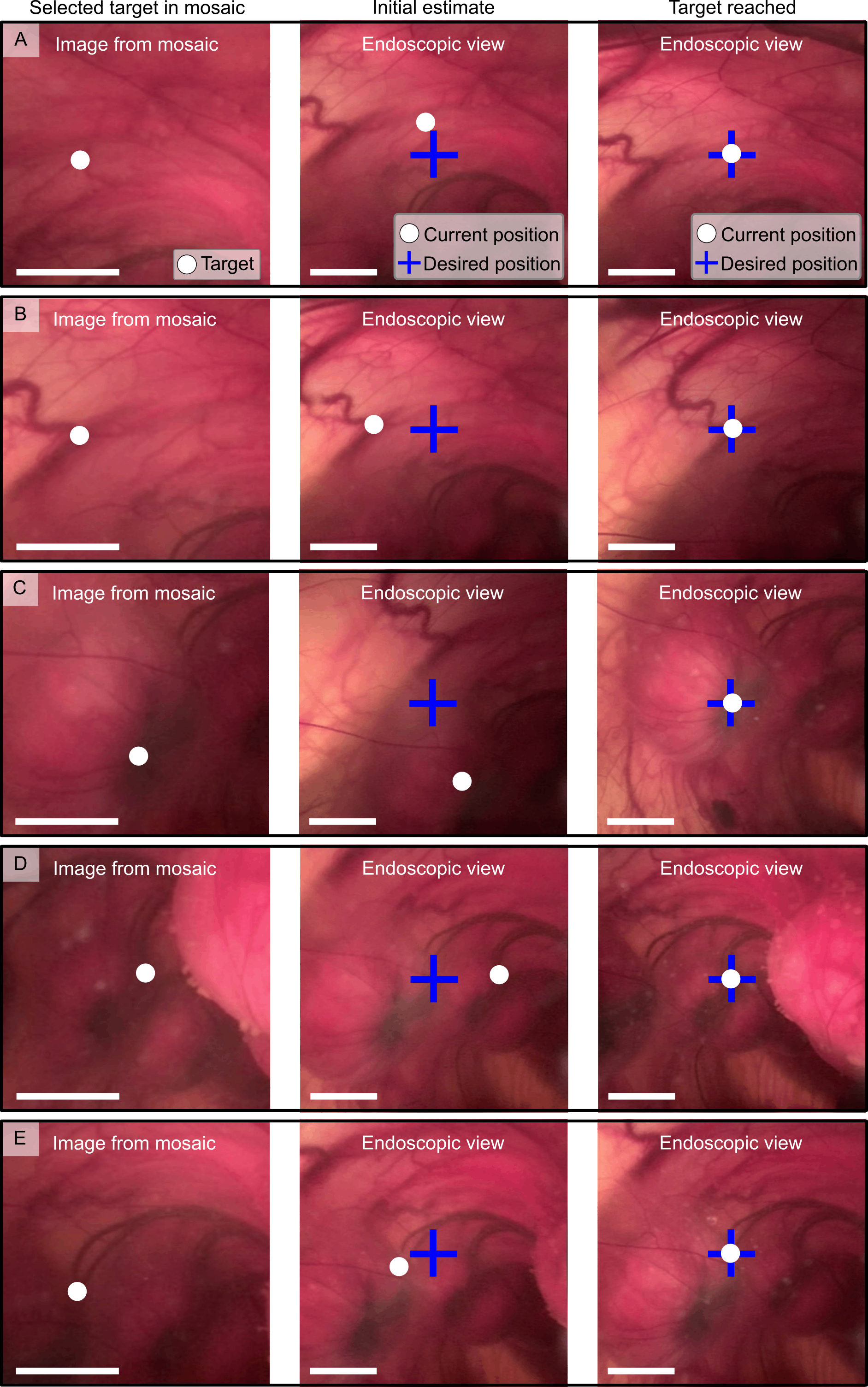} 
	\caption{\textbf{In vivo long-range navigation.}
    Keyframes of the long-range navigation to five targets. First column: selected target on the mosaic. Second column: endoscopic image of the initial open-loop estimate of the magnetic field. Third column: final image after the target was reached with closed-loop visual servoing. The scale bar corresponds to 100 pixels.
        }
	\label{sfig:in_vivo_jumping}
\end{figure}

\begin{figure}[ht]
	\centering
	\includegraphics[width=0.6\textwidth]{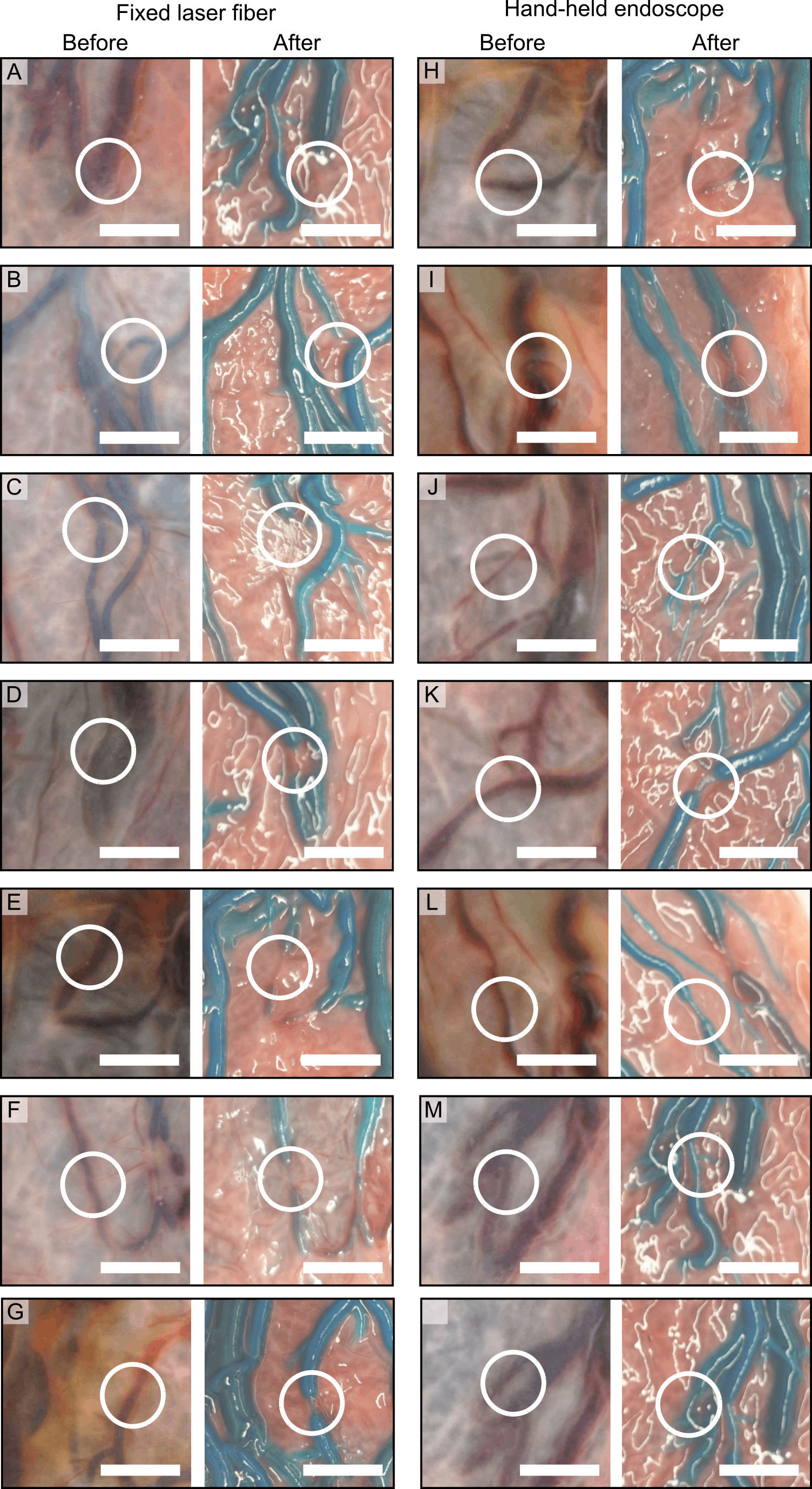} 
	\caption{\textbf{Ablation sites for ex vivo experiments.}
    On the left are all the ablation sites, where the laser fiber was fixated and on the right the ones where a replica of a hand-held, rigid fetoscope (11508AAK, Karl Storz)
 was used and the ablation was influenced by a hand tremor. (\textbf{A-C}) show complete ablations, (\textbf{D-E}) are semi-closed and (\textbf{F-N}) show open vessels. The scale bar corresponds to 10~mm.
        }
	\label{sfig:laser_eval_supp_results_1}
\end{figure}

\begin{figure}[ht]
	\centering
	\includegraphics[width=1\textwidth]{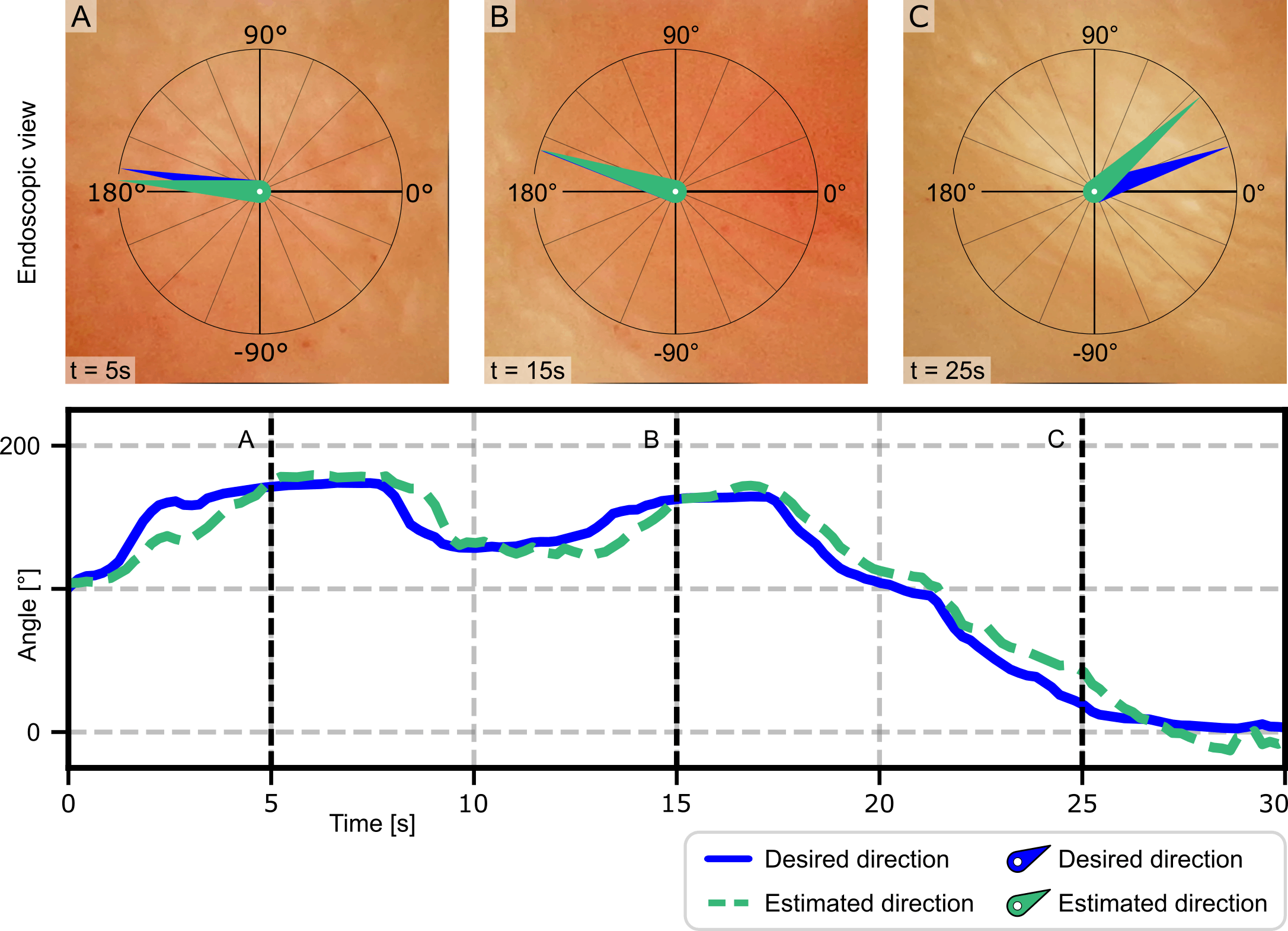} 
	\caption{\textbf{Ex vivo manual navigation in a porcine urinary bladder.}
    Comparison of desired direction angles from joystick inputs and estimated directions during image-based manual navigation. (\textbf{A}--\textbf{C}) Endoscopic images at three points in time, overlaid with the corresponding desired and actual angles for visualization.
    }
	\label{sfig:cysto_pov}
\end{figure}

\begin{figure}[ht]
	\centering
	\includegraphics[width=\textwidth]{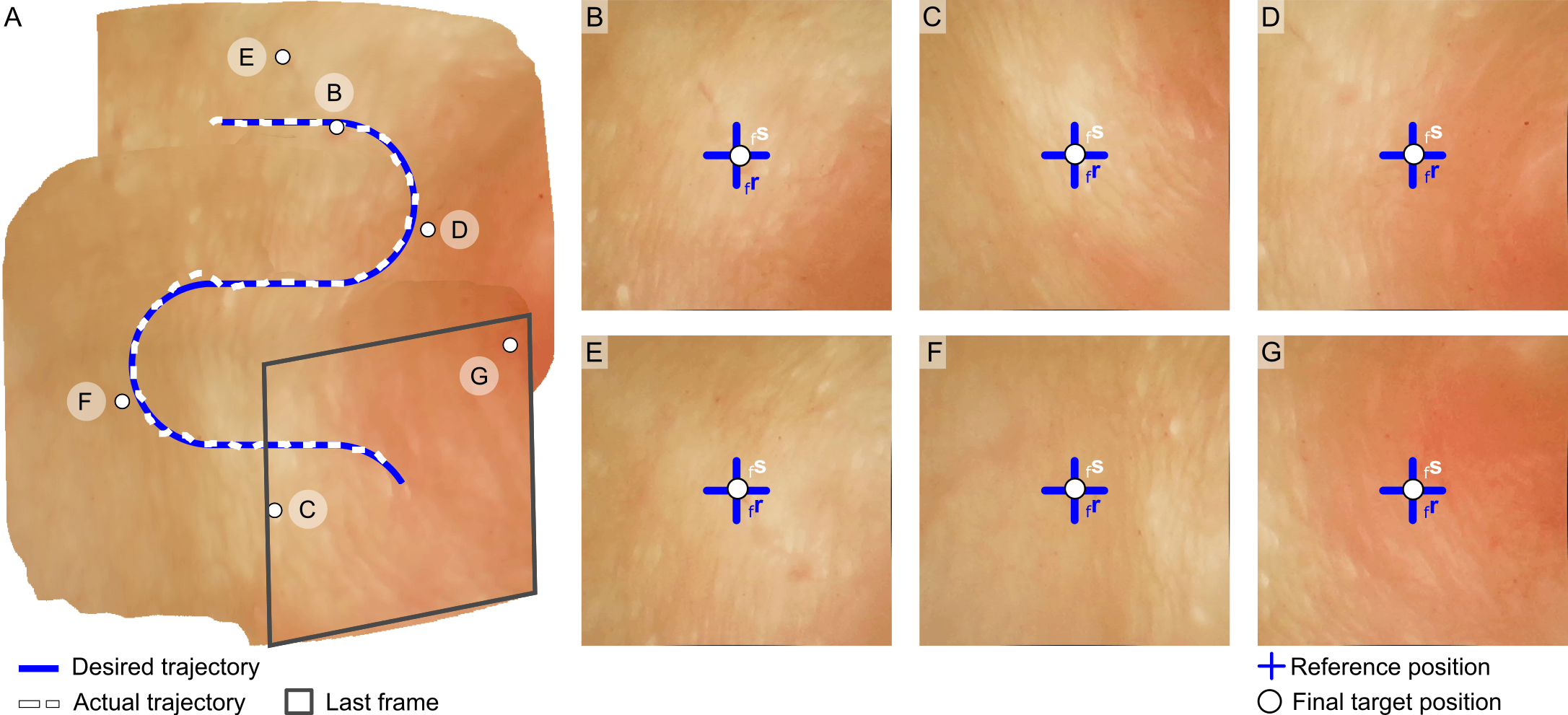} 
	\caption{\textbf{Ex vivo long-range navigation in a porcine urinary bladder.}
    (\textbf{A}) Endoscopic image mosaic serving as a user interface for the long-range automated navigation, showing the desired and actual trajectory from the generation of the mosaic. (\textbf{B--G}) Endoscopic images of the reached targets.
    }
	\label{sfig:cysto_jump}
\end{figure}

\begin{figure}[ht]
	\centering
	\includegraphics[width=\textwidth]{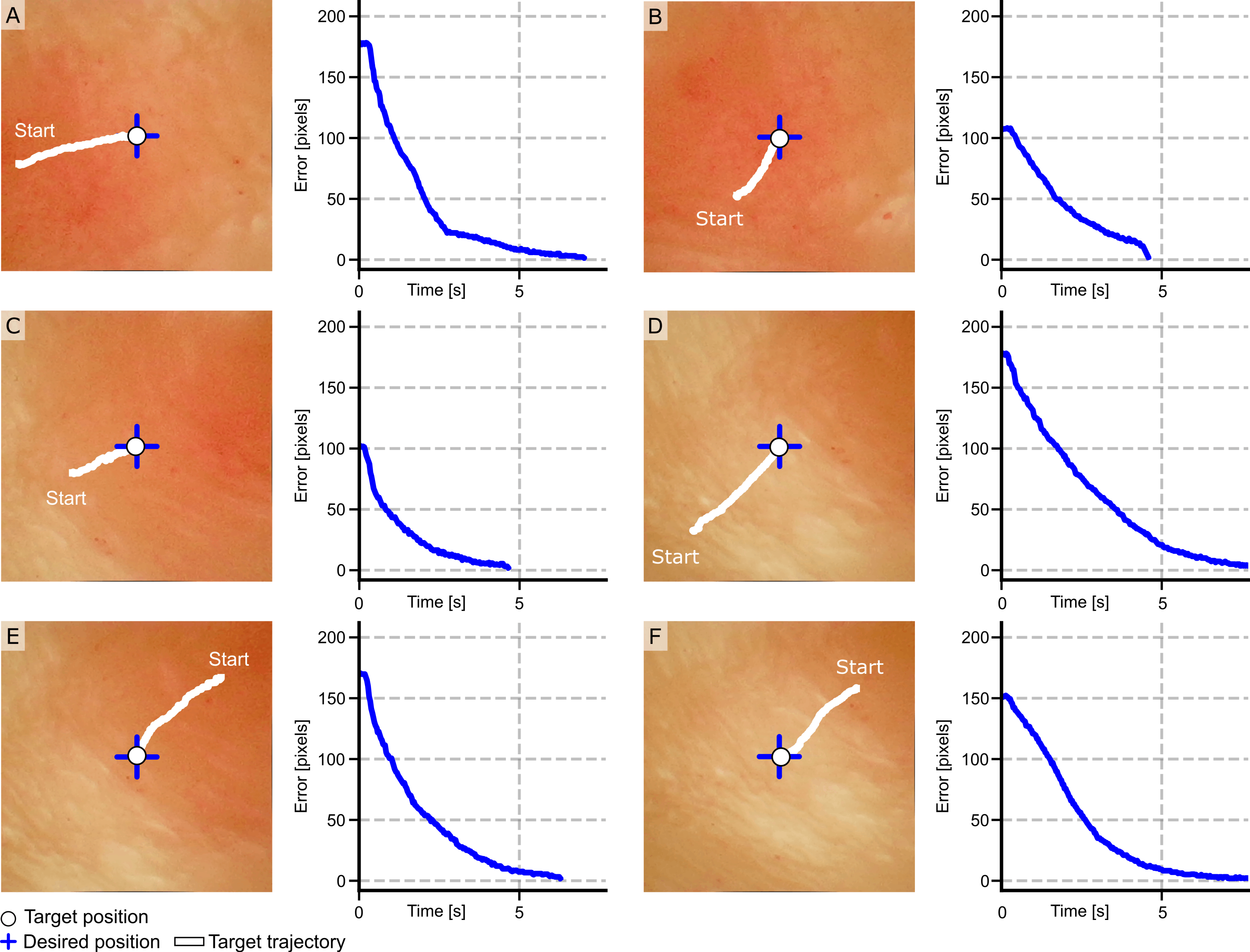} 
	\caption{\textbf{Short distance navigation in an ex vivo porcine urinary bladder.} Trajectory of the tracked target position over time and plots of the pixel error over time for five different targets.
    }
	\label{sfig:cysto_vs_in_img}
\end{figure}

\begin{figure}[ht]
	\centering
	\includegraphics[width=1\textwidth]{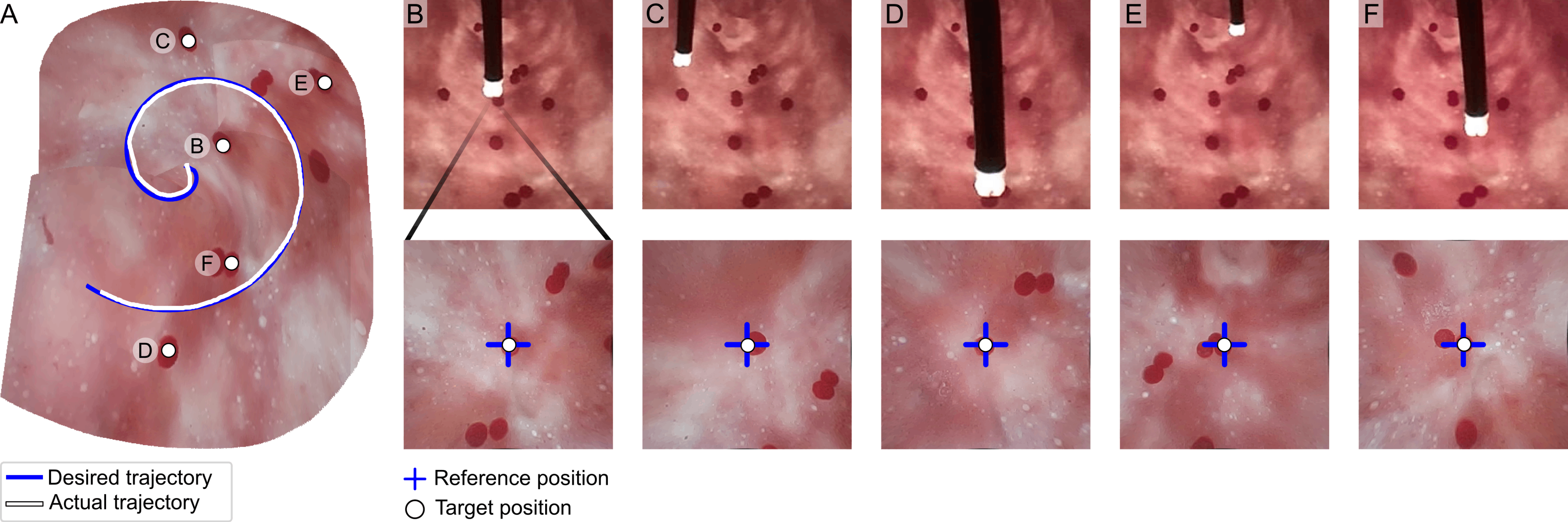} 
	\caption{\textbf{Long-range navigation of a magnetic gastroscope.} (\textbf{A}) Endoscopic image mosaic of the stomach with three manually selected target locations and the desired and actual trajectory from its generation. (\textbf{B}--\textbf{D}) External (top) and endoscopic views (bottom) of the reached target positions after using the long-range navigation method.
    }
	\label{sfig:tne_jump}
\end{figure}

\begin{figure}[ht]
	\centering
	\includegraphics[width=0.75\textwidth]{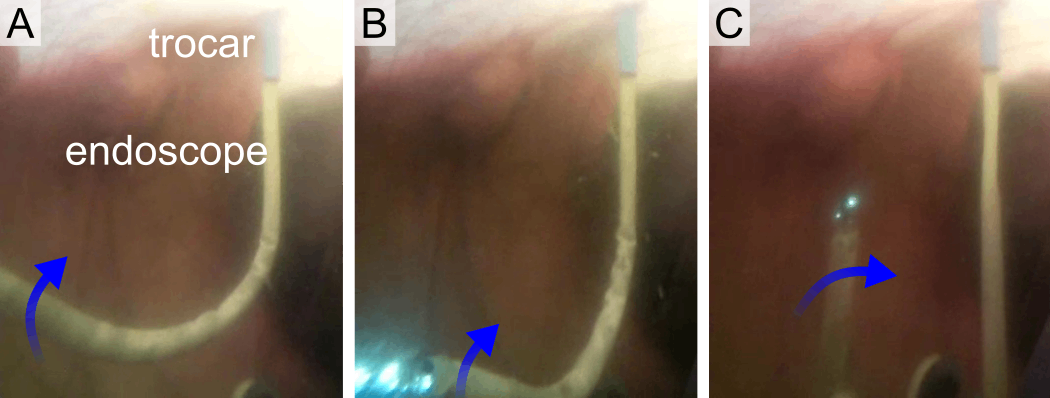} 
	\caption{\textbf{In vivo fetoscope manipulation.} (\textbf{A-C}) External view showing the dexterity of the robotic fetoscope in vivo.
        }
    \label{sfig:fetoscope_dexterity_in_vivo}
\end{figure}

\begin{figure}[ht]
	\centering
	\includegraphics[width=0.6\textwidth]{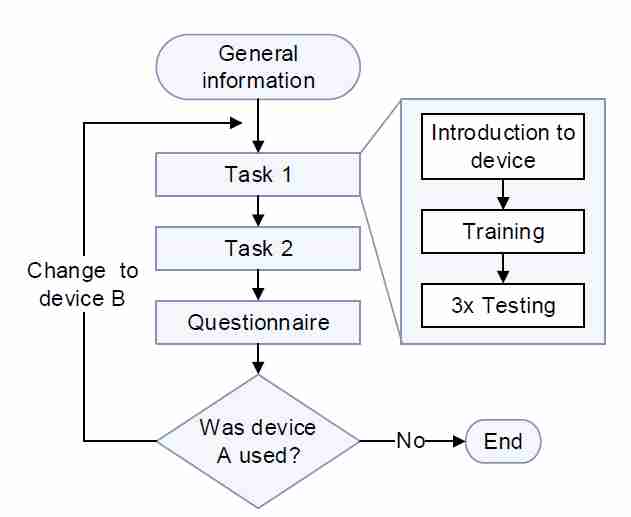} 
	\caption{\textbf{Protocol of the usability study.} Flowchart illustrating the timeline of the usability study. The device first used during the study was alternated between participants.
    }
	\label{sfig:study_protocol}
\end{figure}

\end{document}